\documentclass[7pt,twocolumn,letterpaper]{article}

\usepackage[letterpaper, margin=1.8cm]{geometry}

\usepackage{graphicx}
\usepackage{amsmath}
\usepackage{amssymb}
\usepackage{booktabs}
\usepackage{caption}
\usepackage{subcaption}
\usepackage{multirow}
\usepackage[pagebackref,breaklinks,colorlinks]{hyperref}
\usepackage{times}
\usepackage{authblk}
\usepackage{xspace}
\usepackage[accsupp]{axessibility}

\usepackage{xcolor}
\definecolor{meshred}{rgb}{0.99,0.32,0.32}
\definecolor{meshblue}{rgb}{0.26,0.62,1.0}

\usepackage[capitalize]{cleveref}
\crefname{section}{Sec.}{Secs.}
\Crefname{section}{Section}{Sections}
\Crefname{table}{Table}{Tables}
\crefname{table}{Tab.}{Tabs.}

\newcommand{\ie}{\textit{i.e.}\xspace}
\newcommand{\eg}{\textit{e.g.}\xspace}

\newcommand{\Eg}{\textit{E.g.}\xspace}

\newcommand{\cf}{\textit{cf.}\xspace}

\newcommand{\wrt}{\textit{w.r.t.}\xspace}

\newcommand{\PAR}[1]{\vskip4pt \noindent{\bf #1~}}

\begin{document}

\title{Visual Localization using Imperfect 3D Models from the Internet}
\date{}

\author[1,2]{Vojtech Panek}
\author[3]{Zuzana Kukelova}
\author[1]{Torsten Sattler}
\affil[1]{Czech Institue of Informatics, Robotics and Cybernetics, Czech Technical University in Prague}
\affil[2]{Faculty of Electrical Engineering, Czech Technical University in Prague}
\affil[3]{Visual Recognition Group, Faculty of Electrical Engineering, Czech Technical University in Prague}

\maketitle

\begin{abstract}
Visual localization is a core component in many applications, including augmented reality (AR). Localization algorithms compute the camera pose of a query image \wrt a scene representation, which is typically built from images. This often requires capturing and storing large amounts of data, followed by running Structure-from-Motion (SfM) algorithms. An interesting, and underexplored, source of data for building scene representations are 3D models that are readily available on the Internet, \eg, hand-drawn CAD models, 3D models generated from building footprints, or from aerial images. These models allow to perform visual localization right away without the time-consuming scene capturing and model building steps. Yet, it also comes with challenges as the available 3D models are often imperfect reflections of reality. \Eg, the models might only have generic or no textures at all, might only provide a simple approximation of the scene geometry, or might be stretched. This paper studies how the imperfections of these models affect localization accuracy. We create a new benchmark for this task and provide a detailed experimental evaluation based on multiple 3D models per scene. We show that 3D models from the Internet show promise as an easy-to-obtain scene representation. At the same time, there is significant room for improvement for visual localization pipelines. To foster research on this interesting and challenging task, we release our benchmark at \href{https://v-pnk.github.io/cadloc/}{v-pnk.github.io/cadloc}.
\end{abstract}

\section{Introduction}
\begin{figure}[t]
    \centering
    \includegraphics[width=\columnwidth,trim={8.0cm 14.6cm 12.0cm 15cm},clip]{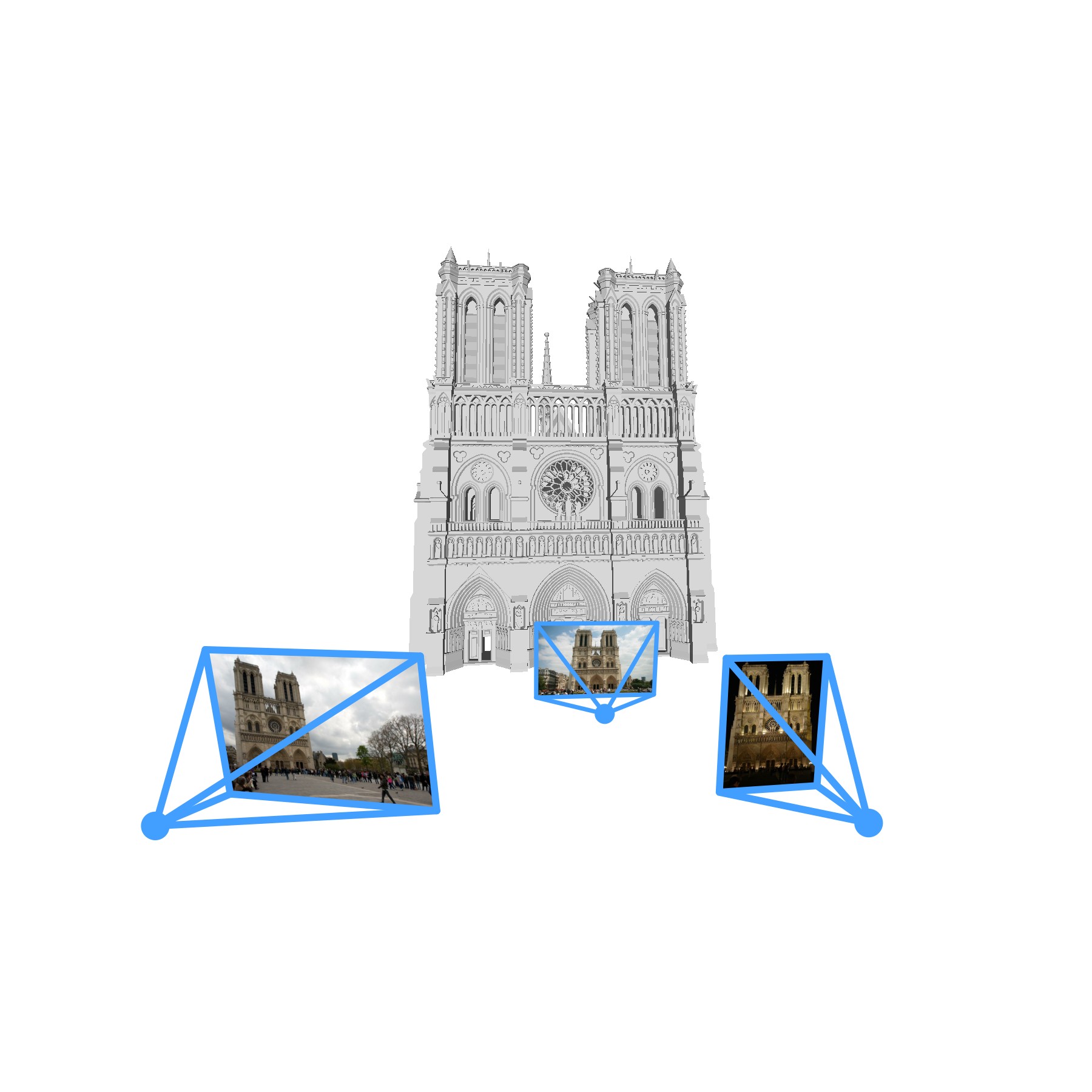}
    \caption{We evaluate the use of 3D models from the Internet for visual localization. Such models can differ significantly from the real world in terms of geometry and appearance.}
    \label{fig:teaser}
\end{figure}

Visual localization is the task of estimating the precise position and orientation, \ie, the camera pose, from which a given query image was taken. 
Localization is a core capability for many applications, including self-driving cars~\cite{Heng2019ICRA}, drones~\cite{Lim12CVPR}, and augmented reality (AR)~\cite{Lynen2020IJRR,Castle08ISWC}. 

Visual localization algorithms mainly differ in the way they represent the scene, \eg, explicitly as a 3D model~\cite{Germain2020ECCV,Li2012ECCV,Sattler2017PAMI,Svarm2017PAMI,Zeisl2015ICCV,Liu2017ICCV,Lynen2020IJRR,Sarlin2019CVPR,Sarlin2020CVPR,Sarlin2021CVPR,Germain20193DV,Schoenberger2018CVPR,Irschara09CVPR} or a set of posed images~\cite{Zheng2015ICCV,Bhayani2021ICCV,Zhang06TDPVT,Pion20203DV}, or implicitly via the weights of a neural network~\cite{Kendall2015ICCV,Kendall2017CVPR,Walch2017ICCV,Balntas2018ECCV,Moreau2021CORL,Brachmann2017CVPR,Brachmann2018CVPR,brachmann2020ARXIV,Valentin2015CVPR,Cavallari20193DV}, and in the way they estimate the camera pose, \eg, based on 2D-3D~\cite{Sattler2017PAMI,Zeisl2015ICCV,Svarm2017PAMI,Sarlin2019CVPR,Sarlin2020CVPR,Germain20193DV,Germain2020ECCV} or 2D-2D~\cite{Zheng2015ICCV,Bhayani2021ICCV} matches or as a weighted combination of a set of base poses~\cite{Kendall2015ICCV,Kendall2017CVPR,Walch2017ICCV,Moreau2021CORL,Sattler2019CVPR,Pion20203DV}. 
Existing visual localization approaches typically use RGB(-D) images to construct their scene representations. 
Yet, capturing sufficiently many images and estimating their camera poses, \eg, via Structure-from-Motion (SfM)~\cite{Snavely-IJCV08,Schoenberger2016CVPR}, are challenging tasks by themselves, especially for non-technical users unfamiliar with how to best capture images for the reconstruction task.

As shown in~\cite{Panek2022ECCV,Zhang2020IJCV,Sibbing133DV,Shan-3DV14,Aubry2016VisualGO}, modern local features such as~\cite{Dusmanu2019CVPR,DeTone2018CVPRWorkshops,Sun2021CVPR,Zhou2021CVPR} are capable of matching real images with renderings of 3D models, even though they were never explicitly trained for this task. 
This opens up a new way of obtaining the scene representations needed for visual localization: 
rather than building the representation from images captured in the scene, \eg, obtained via a service such as Google Street View, crowd-sourcing~\cite{Untzelmann13BigData}, or from photo-sharing websites~\cite{Li2010ECCV}, one can simply download a ready-made 3D model from the Internet (\cf \cref{fig:teaser}), \eg, from 3D model sharing websites such as \href{https://sketchfab.com}{Sketchfab} and  \href{https://3dwarehouse.sketchup.com/}{3D Warehouse}. 
This removes the need to run a SfM system such as COLMAP~\cite{Schoenberger2016CVPR} or \href{https://www.capturingreality.com/}{RealityCapture} on image data, which can be a very complicated step, especially for non-experts such as artists designing AR applications. 
As such, this approach has the potential to significantly simplify the deployment of visual localization-based applications.

\begin{figure*}[t]
    \centering
    \includegraphics[width=0.80\textwidth,trim={0.0cm 2.7cm 0.3cm 0.5cm},clip]{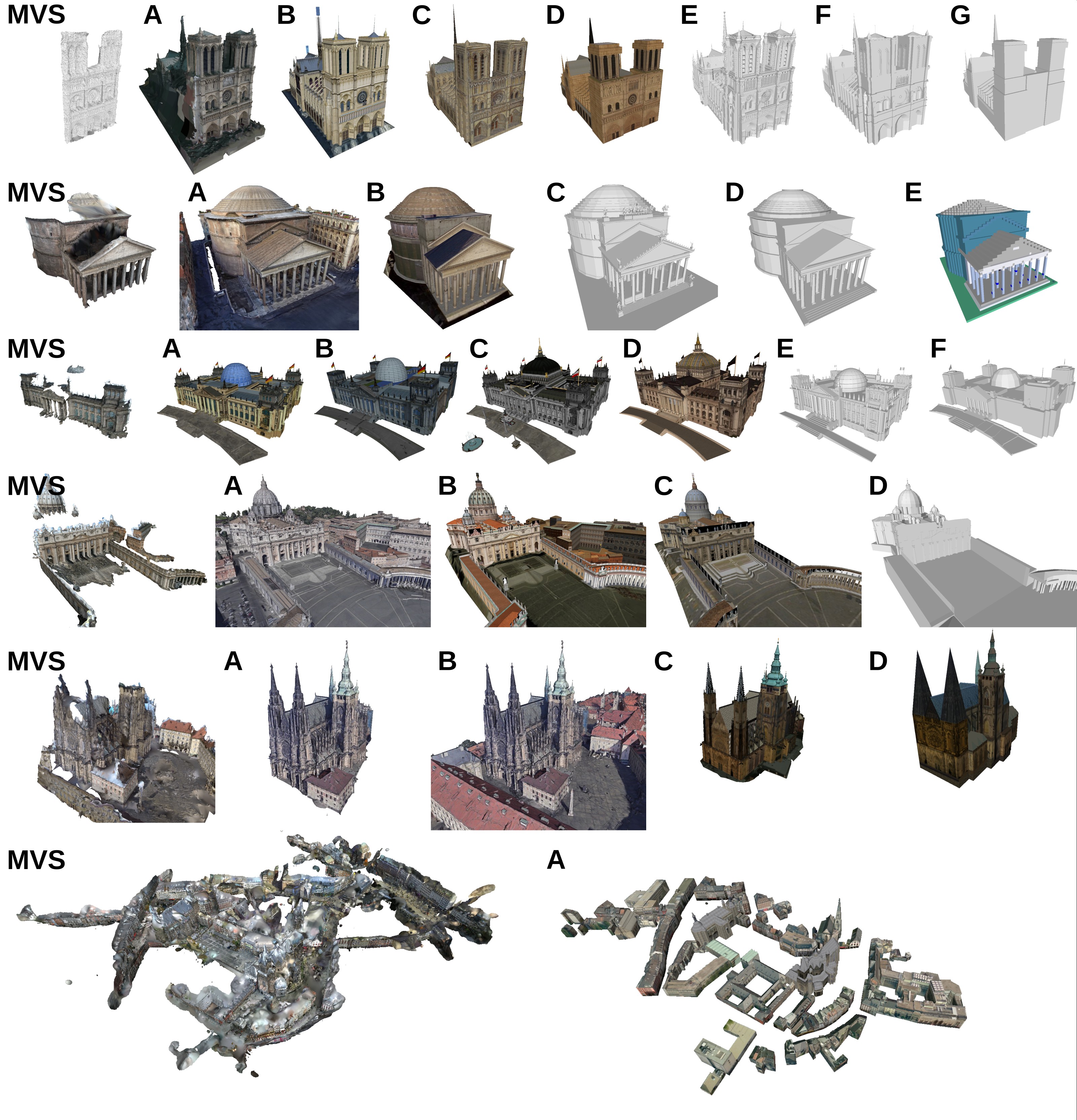} 
    \vspace{-6pt}
    \caption{Different 3D models downloaded from the Internet for multiple scenes (\cf \cref{tab:datasets} for details). Top to bottom: Notre Dame (Front Facade), Pantheon (Exterior), Reichstag, St. Peter's Square, St. Vitus Cathedral, Aachen. The first model in each row is a reference Multi-View Stereo (MVS) model reconstructed from images. The Internet models 
    vary in their fidelity of appearance, \ie, how closely a model's texture reflects reality, and the fidelity of their geometry, \ie, how accurately a model's 3D geometry matches the MVS model.}
    \label{fig:mesh_all}
\end{figure*}

However, using readily available 3D models from the Internet to define the scene representation also comes with its own set of challenges (\cf \cref{fig:mesh_all}): 
(1) \textbf{fidelity of appearance}: the 3D models might not be colored / textured, thus resulting in very abstract representations that are hard to match to real images~\cite{Panek2022ECCV}. 
Even if a model is textured, the texture might be generic and repetitive rather than based on the real appearance of the scene. 
In other cases, the texture might be based on real images, but severely distorted or stretched if these images were captured from drones or planes.  
(2) \textbf{fidelity of geometry}: some 3D models might be obtained via SfM and Multi-View Stereo (MVS), resulting in 3D models that accurately represent the underlying scene geometry. 
Yet, this does not always need to be the case. 
For example some models might be obtained by extruding building outlines, resulting in a very coarse model of the scene geometry. 
Others might be created by artists by hand, resulting in visually plausible models with overly simplified geometry, or with wrong aspect ratios, \eg, a model might be too high compared to the width of the building. 

Naturally, the imperfections listed above negatively affect localization accuracy. The goal of this work is to quantify the relation between model inaccuracy and localization accuracy. This will inform AR application designers which 3D models are likely to provide precise pose estimates. Looking at \cref{fig:mesh_all}, humans seem to be able to establish correspondences between the models, even if the models are very coarse and untextured. Similarly, humans are able to point out correspondences between coarse and untextured models and real images that can be used for pose estimation in the context of visual localization. As such, we expect that it is possible to teach the same to a machine. 
We thus hope that this paper will help researchers to develop algorithms that close the gap between human and machine performance for this challenging matching task. 

In summary, this paper makes the following contributions: (\textbf{1}) we introduce the challenging and interesting tasks of visual localization \wrt to 3D models downloaded from the Internet. 
(\textbf{2}) we provide a new benchmark for this task that includes multiple scenes and 3D models of different levels of fidelity of appearance and geometry. 
(\textbf{3}) we present detailed experiments evaluating how these different levels of fidelity affect localization performance. 
We show that 3D models from the Internet represent a promising new category of scene representations. 
Still, our results show that there is significant room for improvement, especially for less realistic models (which often are very compact to store). 
(\textbf{4}) We make our benchmark publicly available (\href{https://v-pnk.github.io/cadloc/}{v-pnk.github.io/cadloc}) to foster research on visual localization algorithms capable of handling this challenging task.

\section{Related Work}

\PAR{Localization based on images.}
Currently, state-of-the-art localization approaches that can handle strong changes in scene appearance, \eg, day-night and seasonal changes~\cite{Sattler2018CVPR,Toft2022TPAMI}, are based on local features \cite{Sattler2017PAMI,Svarm2017PAMI,Li2012ECCV,Schoenberger2018CVPR,Sarlin2019CVPR,Sarlin2020CVPR,HumenbergerX20Kapture,brejcha2020landscapear,Zeisl2015ICCV,Sattler2015ICCV,Taira2018CVPR,Taira2019ICCV, peng2021megloc, maiwald2022,Irschara09CVPR}. 
They typically represent the scene through 
Structure-from-Motion (SfM) point clouds, where each 3D point is associated with 2D image features from the database images used to reconstruct the point cloud.
2D-3D matches between features extracted in the query images and 3D points in the scene model are then used to estimate the query pose by applying a minimal solver~\cite{Persson2018ECCV, PoseLib} inside a RANSAC scheme~\cite{Fischler1981RandomSC,Lebeda2012BMVC,Chum09CVPR,Raguram2013PAMI,barath2018graph,barath2019progressive,Barth2019MAGSACMS,barath2020magsac++}. 
Hierarchical localization approaches~\cite{Irschara09CVPR,Sattler2012BMVC,Sarlin2018CORL,Taira2018CVPR,Taira2019ICCV,HumenbergerX20Kapture} use an intermediate image retrieval~\cite{Sivic03ICCV,Torii-CVPR15,Jegou-CVPR10,Arandjelovic2016CVPR,gordo2017end,revaud2019learning} step to focus on smaller parts of the scene for 2D-3D matching, thus increasing the scalability of localization techniques.  
Rather than using sparse SfM point clouds, dense meshes can also be used as a scene representation~\cite{Panek2022ECCV}.
The scene geometry can also be represented implicitly through the weights of machine learning models~\cite{Shotton2013CVPR,Brachmann2017CVPR,Brachmann2018CVPR,brachmann2020ARXIV,Valentin20163DV,Cavallari20193DV,Cavallari2017CVPR,Cavallari2019TPAMI}.

An alternative to using a 3D model is to only store a database of images with known camera poses. 
The camera pose of the query can then be either approximated based on the poses of database images retrieved via image retrieval~\cite{Sivic03ICCV,Torii-CVPR15,Jegou-CVPR10,Arandjelovic2016CVPR,gordo2017end,revaud2019learning}, or accurately estimated from 2D-2D matches between the query and the retrieved images~\cite{Bhayani2021ICCV,Zheng2015ICCV}.
It is further possible to implicitly store the database images by training neural networks that regress the camera pose of a query image \wrt the database images~\cite{Kendall2015ICCV,Kendall2017CVPR,Shavit2021ICCV,Walch2017ICCV,Moreau2021CORL,Naseer2017IROS,
Ding2019ICCV,Balntas2018ECCV,Laskar2017ICCVW,Ng2021ARXIV}.

Common to these approaches is that their scene representation is derived from images and thus accurately models the scene geometry and appearance.
We investigate how inaccurate approximations of the scene geometry and appearance, in the form of 3D models downloaded from the Internet, impact feature-based localization performance. 

\PAR{Localization using multi-modal data.}
While less popular than sparse SfM-based representations, dense meshes have been used in the visual localization literature~\cite{Panek2022ECCV, Tomesek2022WACV,Shan-3DV14,Sibbing133DV,Mueller2019PIA,Zhang2020IJCV,brejcha2020landscapear,Aubry-ACMTOG14, Aubry2016VisualGO,Ramalingam2010SKYLINE2GPSLI, Grelsson2020GPSlevelAC, Cadk2018AutomatedOD, Pltz2017AutomaticRO}. 
Prior work can roughly be divided into methods using specialized strategies to align real images to (potentially coarse) 3D models~\cite{Ramalingam2010SKYLINE2GPSLI, Grelsson2020GPSlevelAC,Pltz2017AutomaticRO}, \eg, using contours~\cite{Pltz2017AutomaticRO} or skylines~\cite{Ramalingam2010SKYLINE2GPSLI}, and works based on matching local features~\cite{Panek2022ECCV, Tomesek2022WACV,Shan-3DV14,Sibbing133DV,Mueller2019PIA,Zhang2020IJCV,brejcha2020landscapear,Aubry-ACMTOG14, Aubry2016VisualGO}. 
The later line of work has shown that modern local features can match real photos against non-photorealistic renderings of colored meshes or even against meshes without any color~\cite{Panek2022ECCV,Tomesek2022WACV,brejcha2020landscapear}. 
This observation has motivated our work on using 3D meshes downloaded from the Internet, with potentially non-realistic appearance, as scene representations for visual localization. 
Prior work in this direction typically used as-geometrically-precise 3D models as possible. 
In contrast, we investigate to what degree distortions in model geometry and appearance can be handled as we have no control over the quality of 3D models available on the Internet. 
To the best of our knowledge, we are the first who, in detail, study how the imperfections of Internet models affect localization accuracy.

CAD and other 3D models are also commonly used by %
object pose estimation approaches~\cite{Gmeli2021ROCARC,Aubry2014Seeing3C,Labbe2020CosyPoseCM, Ponimatkin_2022_CVPR,Georgakis2019LearningLR,Grabner20183DPE,Hodan2021PoseEO,Hodan2015DetectionAF,Hodan2020EPOSE6}. 
These methods learn to establish correspondences between real images and precise models of object instances or 3D models depicting the same class of objects. 
Such instance- or class-specific training is not applicable in our scenario due to a lack of scene-specific training data.

Cross-view matching estimates the camera pose based on correspondences found between the ground-level query image and a 2D bird's eye view map (such as a satellite image or a semantic landscape map)~\cite{Hu2018CVMNetCM, Workman2015WideAreaIG, Lin2015LearningDR, Hu2019ImageBasedGU, Viswanathan2014VisionBR,Shan-3DV14}. 
Similar to our problem setting, they also need to solve a challenging matching task between different modalities. 
As in our approach, these maps can often be obtained from the Internet without visiting the place. On the other hand, the maps capture the actual state of the scene, although the low resolution of such maps can be a similar challenge as the low geometric and appearance fidelity of our 3D models. 

\begin{table*}[t!]
    \setlength{\tabcolsep}{0pt}
    \centering
    \begin{minipage}{\textwidth}
    \scriptsize{
    \renewcommand{\arraystretch}{0.8}
    \begin{tabular*}{\textwidth}{@{\extracolsep{\fill}}lllllllr@{\extracolsep{\fill}}}
        \hline\noalign{\smallskip}
        Scene & ID & Model author & Model source & License & Model type & Color type & Size [MB] \\

        \hline\noalign{\smallskip}
        \multirow{7}{*}{\shortstack[l]{Notre Dame \\ (Front Facade) \\ \cite{Thomee2016YFCC100MTN,Heinly2015ReconstructingTW,Jin2020}\\ 189 queries}}
        & A & Miguel Bandera & Sketchfab (\url{https://skfb.ly/6QWu7}) & \href{https://creativecommons.org/licenses/by-nc-sa/4.0/}{CC BY-NC-SA 4.0} & MVS & texture & 22.4 \\
        & B & Chigirinsky & Sketchfab (\url{https://skfb.ly/6Rn9M}) & \href{https://creativecommons.org/licenses/by/4.0/}{CC BY 4.0} & CAD & texture & 4.8 \\
        & C & Alejandro Diaz & Sketchfab (\url{https://skfb.ly/31dba}) & \href{https://creativecommons.org/licenses/by/4.0/}{CC BY 4.0} & CAD & texture & 3.1 \\
        & D & Little-Goomba & 3D Warehouse (\url{https://bit.ly/3QWeOxY}) & \href{https://3dwarehouse.sketchup.com/tos/#license}{3DW: GML} & CAD & texture & 0.6 \\
        & E & MiniWorld3D & MyMiniWorld (\url{https://mmf.io/o/91899}) & \href{https://mmf.io/o/91899}{BY-ND-NC-EX} & CAD & raw & 30.4 \\
        & F & giotis & 3D Warehouse (\url{https://bit.ly/3QOTQ41}) & \href{https://3dwarehouse.sketchup.com/tos/#license}{3DW: GML} & CAD & raw & 0.7 \\
        & G & Jul & 3D Warehouse (\url{https://bit.ly/3ThrhOt}) & \href{https://3dwarehouse.sketchup.com/tos/#license}{3DW: GML} & CAD & raw & 0.1 \\
        
        \hline\noalign{\smallskip}
        \multirow{5}{*}{\shortstack[l]{Pantheon (Exterior) \\ \cite{Thomee2016YFCC100MTN,Heinly2015ReconstructingTW,Jin2020}\\ 141 queries}}
        & A & Fovea & Sketchfab (\url{https://skfb.ly/6RZHt}) & \href{https://creativecommons.org/licenses/by/4.0/}{CC BY 4.0} & MVS & texture & 84.2 \\
        & B & brnipon & 3D Warehouse (\url{https://bit.ly/3CCwTwP}) & \href{https://3dwarehouse.sketchup.com/tos/#license}{3DW: GML} & CAD & texture & 5.5 \\
        & C & Ultima Ratio & 3D Warehouse (\url{https://bit.ly/3AuN2BK}) & \href{https://3dwarehouse.sketchup.com/tos/#license}{3DW: GML} & CAD & raw & 61.5 \\
        & D & Adsman007 & 3D Warehouse (\url{https://bit.ly/3ASvl0b}) & \href{https://3dwarehouse.sketchup.com/tos/#license}{3DW: GML} & CAD & raw & 24.5 \\
        & E & Emanuele Viani & Sketchfab (\url{https://skfb.ly/EAKB}) & \href{https://creativecommons.org/licenses/by/4.0/}{CC BY 4.0} & CAD & raw & 36.0 \\
        
        \hline\noalign{\smallskip}
        \multirow{6}{*}{\shortstack[l]{Reichstag \\ \cite{Thomee2016YFCC100MTN,Heinly2015ReconstructingTW,Jin2020}\\ 75 queries}}
        & A & Emperor Heer 99 & 3D Warehouse (\url{https://bit.ly/3wzFhci}) & \href{https://3dwarehouse.sketchup.com/tos/#license}{3DW: GML} & CAD & texture & 13.3 \\
        & B & Emperor Heer 99 & 3D Warehouse (\url{https://bit.ly/3ATX2Wk}) & \href{https://3dwarehouse.sketchup.com/tos/#license}{3DW: GML} & CAD & texture & 7.5 \\
        & C & Emperor Heer 99 & 3D Warehouse (\url{https://bit.ly/3ASdFSr}) & \href{https://3dwarehouse.sketchup.com/tos/#license}{3DW: GML} & CAD & texture & 27.3 \\
        & D & Emperor Heer 99 & 3D Warehouse (\url{https://bit.ly/3ctJvvp}) & \href{https://3dwarehouse.sketchup.com/tos/#license}{3DW: GML} & CAD & texture & 6.0 \\
        & E & Klaus T. & 3D Warehouse (\url{https://bit.ly/3cme5qV}) & \href{https://3dwarehouse.sketchup.com/tos/#license}{3DW: GML} & CAD & raw & 5.3 \\
        & F & SH & 3D Warehouse (\url{https://bit.ly/3AP4CBM}) & \href{https://3dwarehouse.sketchup.com/tos/#license}{3DW: GML} & CAD & raw & 0.1 \\
        
        \hline\noalign{\smallskip}
        \multirow{4}{*}{\shortstack[l]{St. Peter's Square \\ \cite{Thomee2016YFCC100MTN,Heinly2015ReconstructingTW,Jin2020}\\ 126 queries}}
        & A & Brian Trepanier & Sketchfab (\url{https://skfb.ly/or8Ip}) & \href{https://creativecommons.org/licenses/by/4.0/}{CC BY 4.0} & MVS & texture & 230.1 \\
        & B & Dounia B. & 3D Warehouse (\url{https://bit.ly/3CCAYkk}) & \href{https://3dwarehouse.sketchup.com/tos/#license}{3DW: GML} & CAD & texture & 131.5 \\
        & C & mstochl & 3D Warehouse (\url{https://bit.ly/3RhqEmc}) & \href{https://3dwarehouse.sketchup.com/tos/#license}{3DW: GML} & CAD & texture & 4.2 \\
        & D & Antonino G. & 3D Warehouse (\url{https://bit.ly/3Rd3KMC}) & \href{https://3dwarehouse.sketchup.com/tos/#license}{3DW: GML} & CAD & raw & 24.5\\
        
        \hline\noalign{\smallskip}
        \multirow{4}{*}{\shortstack[l]{St. Vitus Cathedral \\ (own data) \\ 213 queries}}
        & A & Brian Trepanier & Sketchfab (\url{https://skfb.ly/o8n8D}) & \href{https://creativecommons.org/licenses/by/4.0/}{CC BY 4.0} & MVS & texture & 109.4 \\ 
        & B & Brian Trepanier & Sketchfab (\url{https://skfb.ly/o8n8D}) & \href{https://creativecommons.org/licenses/by/4.0/}{CC BY 4.0} & MVS & texture & 284.9 \\ 
        & C & Pera
        & 3D Warehouse (\url{https://bit.ly/3Tf7Bum}) & \href{https://3dwarehouse.sketchup.com/tos/#license}{3DW: GML} & CAD & texture & 3.7 \\
        & D & Hrusak & 3D Warehouse (\url{https://bit.ly/3Rja6tJ}) & \href{https://3dwarehouse.sketchup.com/tos/#license}{3DW: GML} & CAD & texture & 1.0 \\
        
        \hline\noalign{\smallskip}
        Aachen~\cite{Sattler2012BMVC,Sattler2018CVPR,Zhang2020IJCV}, 1015 queries & A & \cite{Habbecke2012EG} & - & \href{https://creativecommons.org/licenses/by-nc-sa/4.0/}{CC BY-NC-SA 4.0} & CAD & texture & 21.6 \\
        \hline
    \end{tabular*}
    }
    \vspace{-6pt}
    \caption{List of scenes and 3D models used for the evaluation. The query images for the scenes were obtained from the Image Matching Challenge (IMC) 2021~\cite{Thomee2016YFCC100MTN,Heinly2015ReconstructingTW,Jin2020}, the Aachen Day-Night v1.1 dataset~\cite{Sattler2012BMVC,Sattler2018CVPR,Zhang2020IJCV}, and our own recordings. We distinguish between models directly created from images via MVS and models created from human input (CAD).}
    \label{tab:datasets}
    \end{minipage}
\end{table*}

\section{Datasets}
\label{sec:datasets}
The goal of this work is to study how and to what degree inaccurate 3D models downloaded from the Internet, affect visual localization performance. 
We are motivated by the observation that modern local features, such as the ones used in state-of-the-art localization pipelines~\cite{Sarlin2019CVPR,Sarlin2020CVPR,Sun2021CVPR,Panek2022ECCV}, can establish correspondences between a real image and renderings of 3D models~\cite{Zhang2020IJCV,Panek2022ECCV,brejcha2020landscapear}. This allows us to apply a state-of-the-art localization pipeline~\cite{Panek2022ECCV} on renderings of different 3D models. While prior work focused on 3D models obtained from images~\cite{Zhang2020IJCV,Panek2022ECCV}, \ie, models that accurately align with reality, this work aims to answer the question to which degree deviations from reality, both in terms of geometry and appearance, can be handled. 
 
For our experimental evaluation, we collected 3D models for 6 scenes (\cf \cref{fig:mesh_all} and \cref{tab:datasets}). 
The scenes were selected based on the availability of interesting 3D models covering multiple challenges, \eg, different levels of appearance and geometric fidelity, and the availability of query images. 
For the 3D models, we distinguish between models that were automatically obtained from images via MVS and CAD models that were created manually.
The following describes the models used per scene. 
We publish our benchmark (\href{https://v-pnk.github.io/cadloc/}{v-pnk.github.io/cadloc}) to foster research on this interesting and challenging topic. 
In particular, our benchmark can be used to measure how well local features are able to handle complex matching tasks between real images and more abstract representations of the scene.

For each scene, we collected a set of query images with known poses
(in a coordinate system with an arbitrary scale).
We then aligned the models downloaded from the Internet to the coordinate system of the images. 
This allows us to measure the accuracy of the camera poses estimated by the localization algorithm in a common frame of reference and thus to compare pose accuracy between the models.
Unless mentioned otherwise, we used images from the \href{https://www.cs.ubc.ca/research/image-matching-challenge/2021/data/}{Image Matching Challenge (IMC) 2021} as queries. 
The images are part of the YFCC100M dataset~\cite{Thomee2016YFCC100MTN}, depicting popular landmarks at different points in time. 
The \href{https://www.cs.ubc.ca/research/image-matching-challenge/2021/data/}{IMC} provides the camera poses from~\cite{Heinly2015ReconstructingTW} for each image. 
Using these poses, we created a MVS mesh using \href{https://www.capturingreality.com/}{RealityCapture} and aligned the Internet models with this mesh, using ICP-based refinement~\cite{rusinkiewicz2001efficient} starting from a manual initialization. 

\noindent\textbf{Notre Dame (Front Facade).} 
We selected 7 models of Notre Dame from the Internet, representing different levels of geometric detail and fidelity of appearance. 
Model A is the only one coming from photogrammetry reconstruction, the others are CAD models created manually.
The models B, C and D are textured.
While the texture of the model B looks rather realistic, the texture of C seems to come from a painting and the texture of D is a combination of low-quality photos and generic textures.
The geometric detail is the highest for the model B and the lowest in the case of the model D, where the front facade of the building is essentially only planar.
Model C is somewhere in between.
E has a very high level of geometric detail, F is comparable to the level of detail of B and the geometry of D and G is identical.
E, F and G do not contain any color information.

\noindent\textbf{Pantheon (Exterior).} 
Here we use five models:
Model A was created by photogrammetry.
Model B contains realistically looking textures.
Models C and D originally contained generic textures, which did not correspond to reality in any way and so we decided to use just their geometry (Sec.~\ref{sec:model_issues}).
C contains a very high level of geometric detail and adds on top multiple details, \eg, statues, which the real building does not contain in its current state. 
Model D has a medium level of geometric detail.
Model E is constructed from a set of voxels and contains non-realistic coloring.

\noindent\textbf{Reichstag.} 
We have four textured CAD models of Reichstag (A, B, C, D).
Models A and B depict the current state of the building, while C and D are trying to create a historic depiction.
Models E and F do not use texture. 
Models A, C and E have a higher level of geometric detail than the others.

\noindent\textbf{St. Peter's Square.} 
We use four models:
Model A is generated by photogrammetry and also contains surrounding buildings. 
Models B and C have realistic textures.
All models have a comparably high level of geometric detail.

\noindent\textbf{St. Vitus Cathedral.} We use four models: an MVS model (B) containing a larger area, reconstructed from drone footage via photogrammetry, a version of this model that only contains the cathedral (A), and two CAD models (C and D) created using SketchUp.
Models C and D were chosen as they provide different levels of detail, with model C containing finer geometric details that are missing in model D.
Both C and D use low-resolution textures, where the same texture is used for repetitive parts of the building. 
We captured query images with cameras by walking around the cathedral, with the camera focused on the building. 
We constructed a MVS model from these images using the \href{https://www.capturingreality.com/}{RealityCapture} software and aligned the four models used for the experiments against this MVS model. 

\noindent\textbf{Aachen.} We use a CAD model that contains buildings obtained by extruding building outlines based on human modeling from aerial images~\cite{Habbecke2012EG}, as well as hand-modelled buildings. 
The model was kindly provided by the authors of~\cite{Habbecke2012EG} and was textured using aerial images. 
Viewing the model from the ground level thus leads to severe distortions due to the small viewing angle between the facades and the aerial images. 
We use the query images from the Aachen Day-Night v1.1 dataset~\cite{Sattler2012BMVC,Sattler2018CVPR,Zhang2020IJCV}. 
The CAD model does not cover the full extent of the Aachen Day-Night v1.1 model. 
Thus, we only use the queries that depict (part of) the CAD model. 
The poses of the query images are not publicly available. 
Hence, we aligned the CAD model to a MVS model, reconstructed from the database images of the Aachen dataset using Colmap~\cite{Schoenberger2016CVPR,Schoenberger2016ECCV}, which are in the same coordinate system as the queries. 
This allows us to use the evaluation service at \href{https://www.visuallocalization.net/}{visuallocalization.net} to measure pose accuracy for our experiments on Aachen. 

\begin{figure*}[t!]
    \centering
    \parbox{\textwidth}{
        \hspace{0.55cm}
        \includegraphics[width=0.5\textwidth,trim={50 220 12 52},clip]{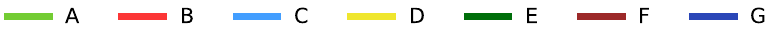}
    }
    \\
    \parbox{\textwidth}{
    \begin{subfigure}[b]{0.202\textwidth}
        \centering
        \includegraphics[width=\textwidth,trim={0 0 0 88},clip]{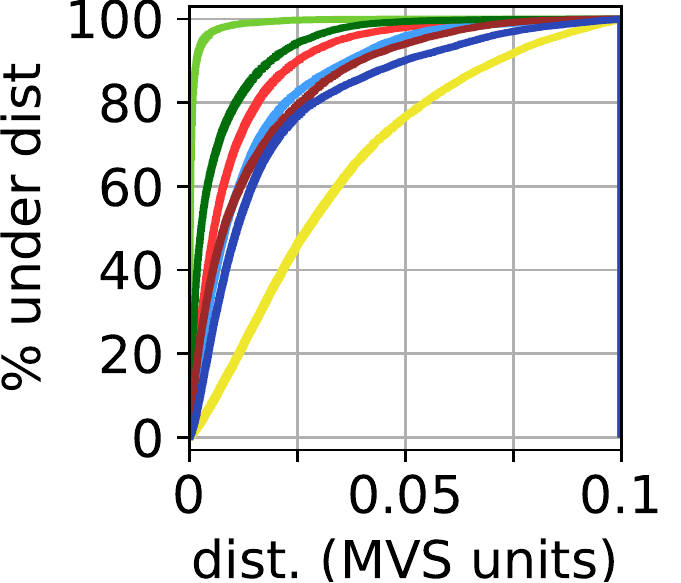}
        \caption{Notre Dame}
    \end{subfigure}
    \hfill
    \begin{subfigure}[b]{0.15\textwidth}
        \centering
        \includegraphics[width=\textwidth,trim={50 0 0 88},clip]{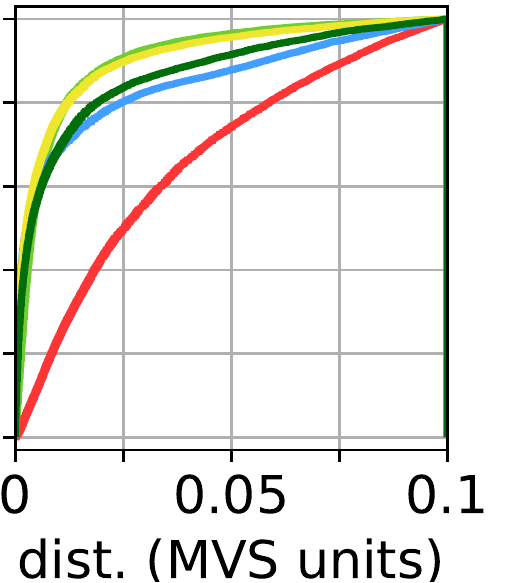}
        \caption{Pantheon}
    \end{subfigure}
    \hfill
    \begin{subfigure}[b]{0.15\textwidth}
        \centering
        \includegraphics[width=\textwidth,trim={50 0 0 88},clip]{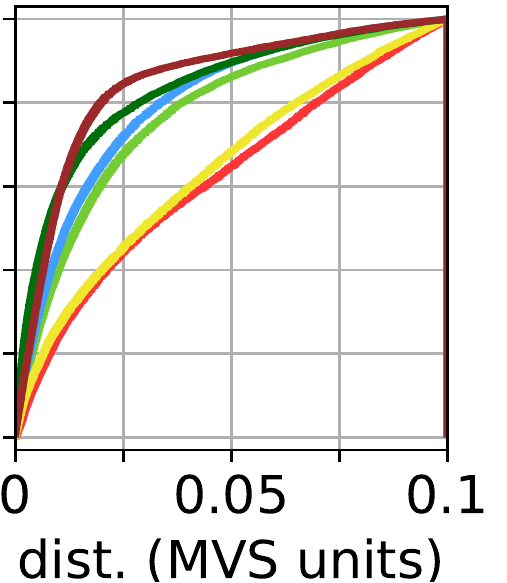}
        \caption{Reichstag}
    \end{subfigure}
    \hfill
    \begin{subfigure}[b]{0.15\textwidth}
        \centering
        \includegraphics[width=\textwidth,trim={50 0 0 88},clip]{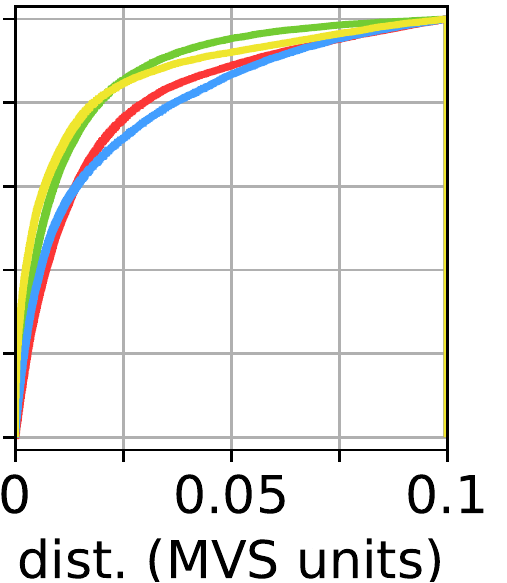}
        \caption{St. Peter's square}
    \end{subfigure}
    \hfill
    \begin{subfigure}[b]{0.15\textwidth}
        \centering
        \includegraphics[width=\textwidth,trim={50 0 0 88},clip]{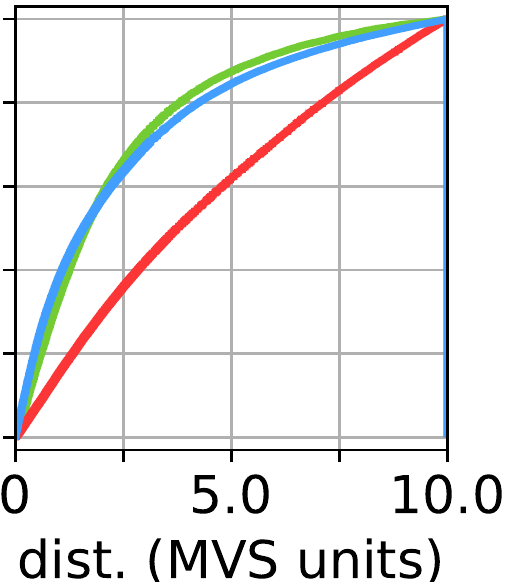}
        \caption{St. Vitus}
    \end{subfigure}
    \hfill
    \begin{subfigure}[b]{0.15\textwidth}
        \centering
        \includegraphics[width=\textwidth,trim={50 0 0 88},clip]{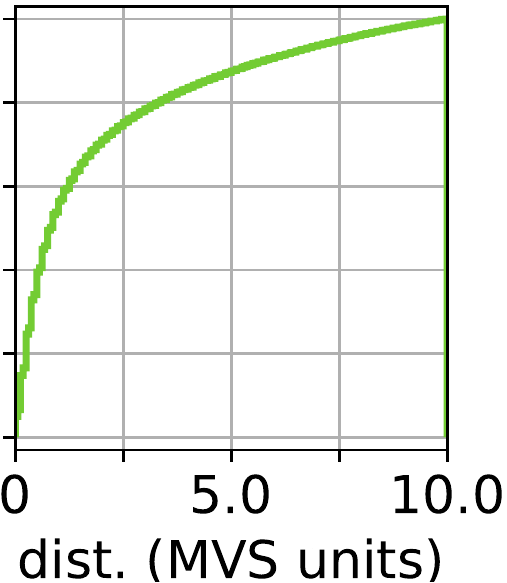}
        \caption{Aachen}
    \end{subfigure}
    \vspace{-6pt}
    \caption{Cumulative histograms of distances from query model vertices to the nearest vertex from the Internet models. Smaller distances indicate a higher level of geometric fidelity for the Internet models.}
    \label{fig:three graphs}
    }
\end{figure*}

\begin{figure*}[t!]
    \centering
    \begin{subfigure}[b]{0.2178\textwidth}
        \centering
        \includegraphics[width=\textwidth,trim={0 0 0 0},clip]{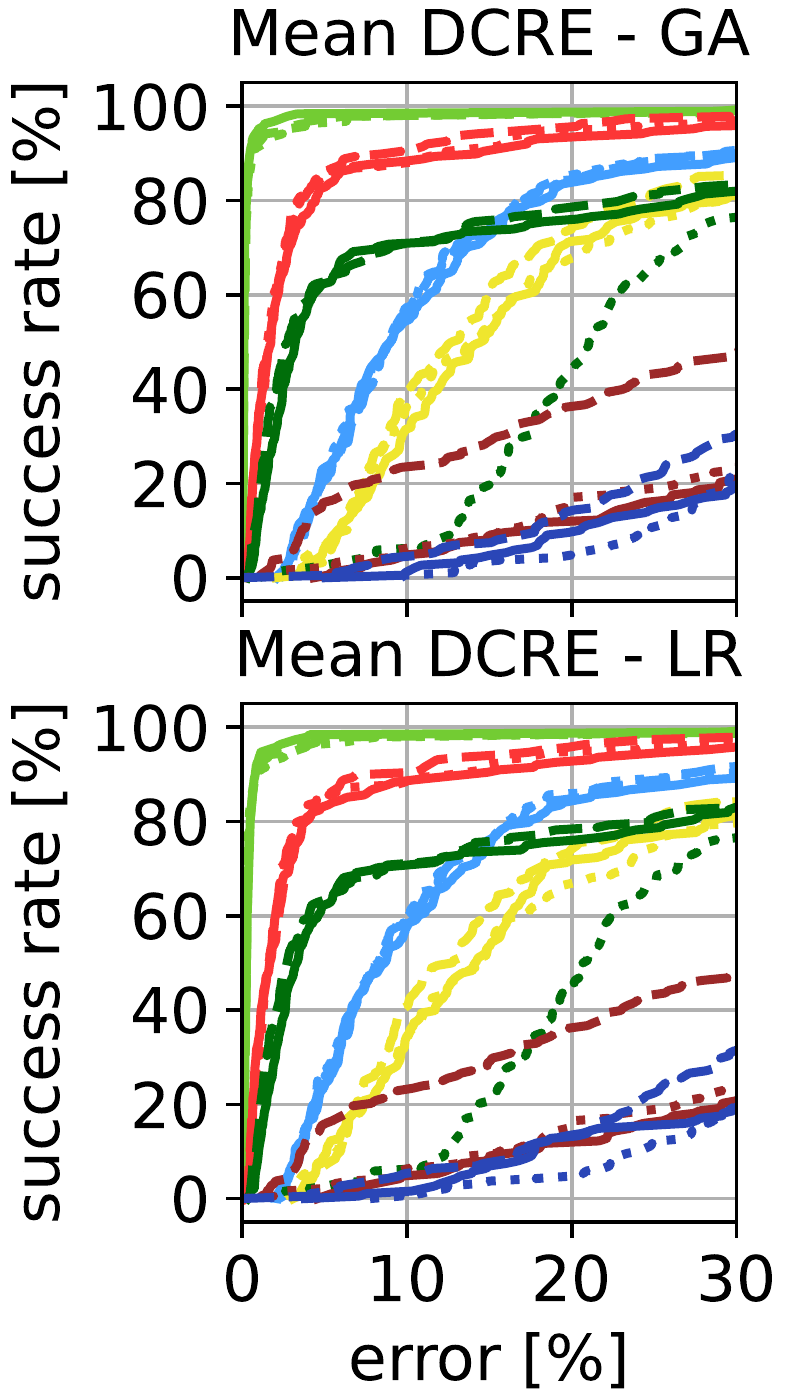}
        \caption{Notre Dame}
    \end{subfigure}
    \begin{subfigure}[b]{0.16\textwidth}
        \centering
        \includegraphics[width=\textwidth,trim={0 0 0 0},clip]{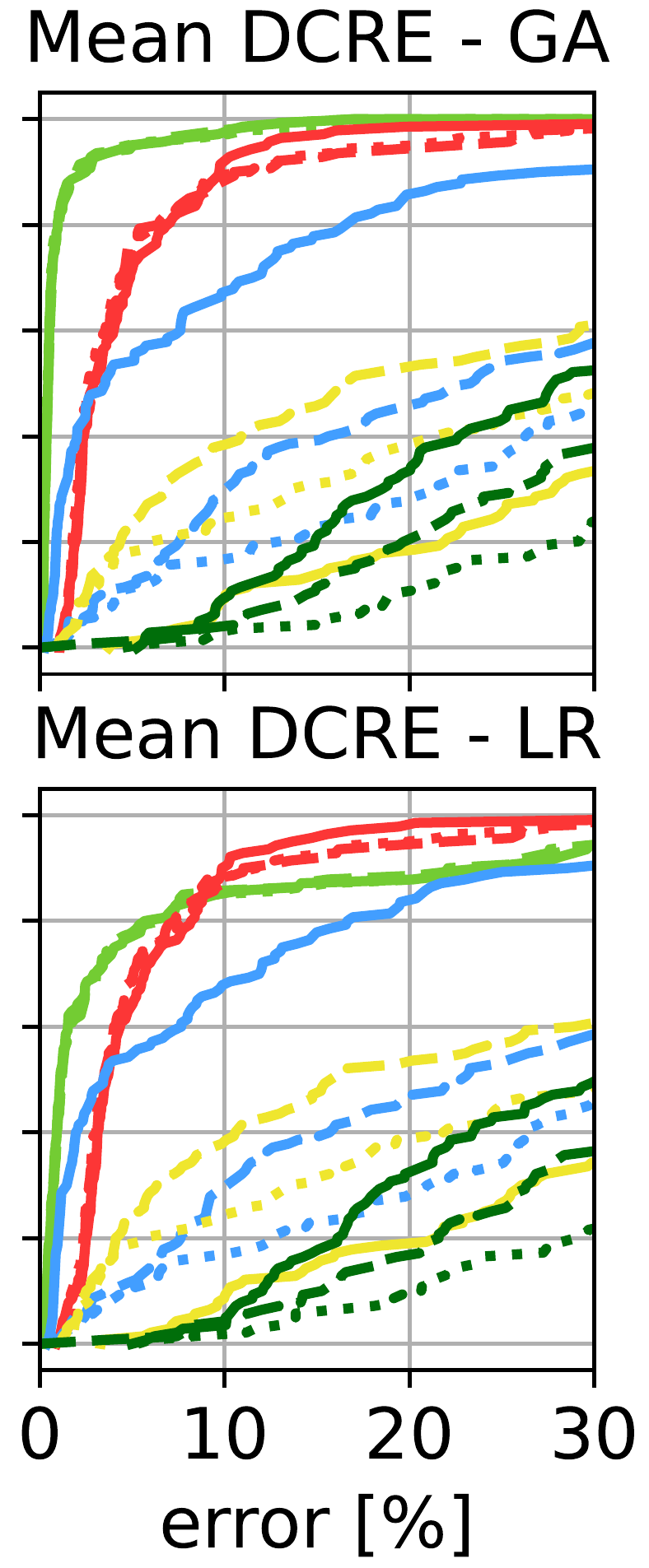}
        \caption{Pantheon}
    \end{subfigure}
    \begin{subfigure}[b]{0.16\textwidth}
        \centering
        \includegraphics[width=\textwidth,trim={0 0 0 0},clip]{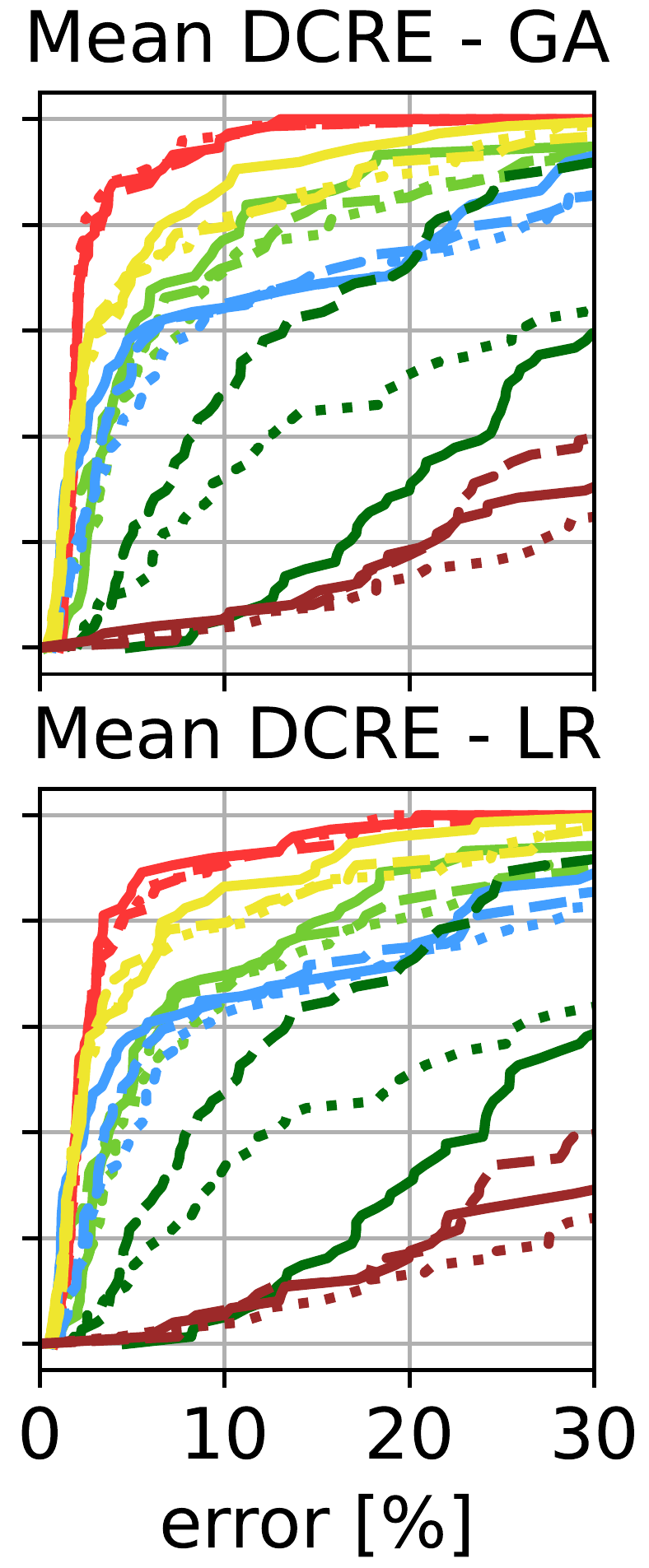}
        \caption{Reichstag}
    \end{subfigure}
    \begin{subfigure}[b]{0.16\textwidth}
        \centering
        \includegraphics[width=\textwidth,trim={0 0 0 0},clip]{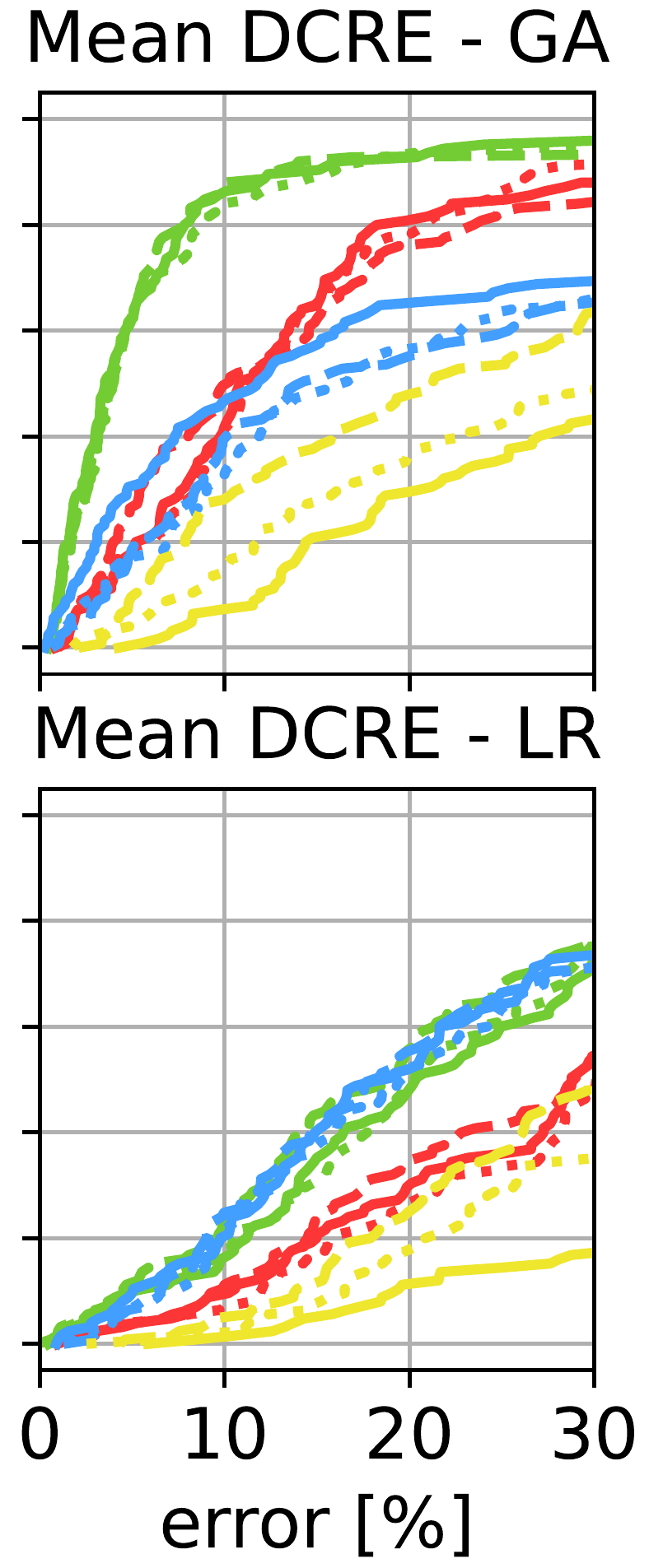}
        \caption{St. Peter's square}
    \end{subfigure}
    \begin{subfigure}[b]{0.16\textwidth}
        \centering
        \includegraphics[width=\textwidth,trim={0 0 0 0},clip]{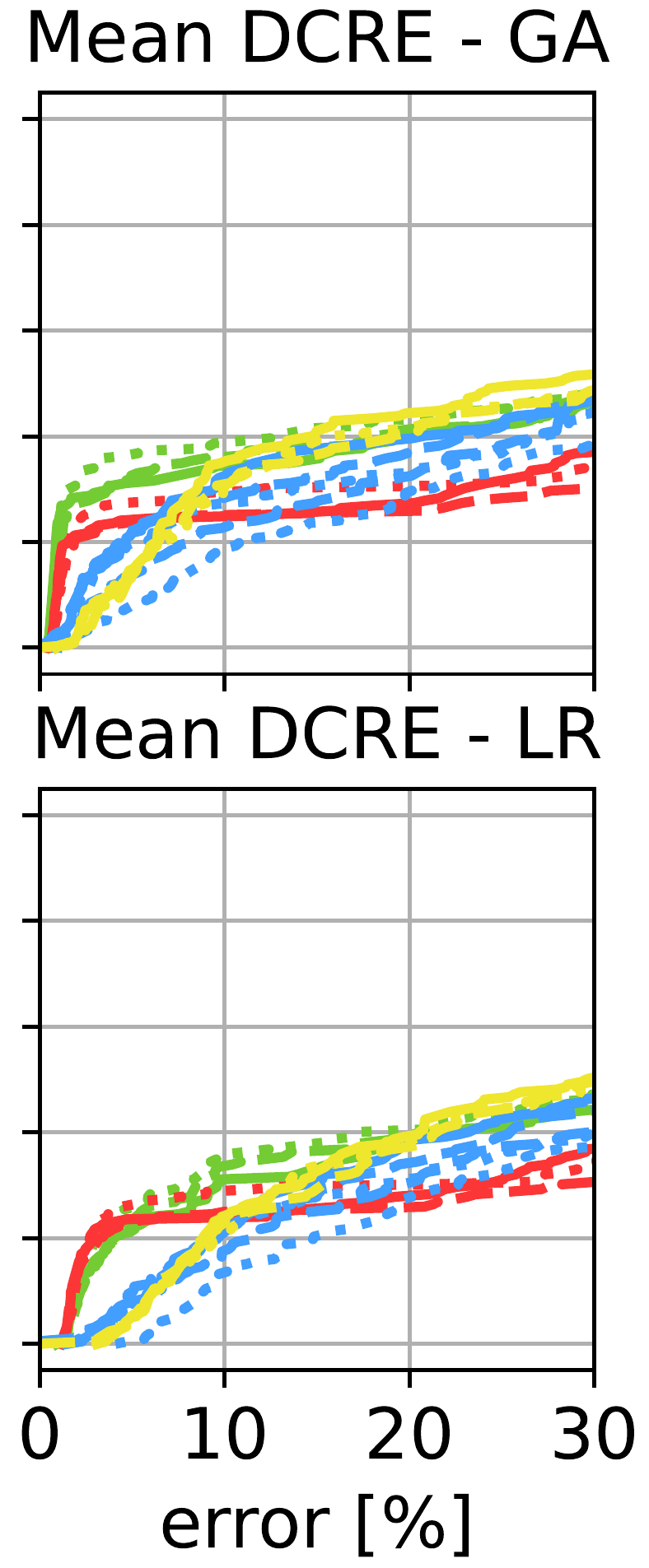}
        \caption{St. Vitus}
    \end{subfigure}
    \begin{subfigure}[b]{0.11\textwidth}
        \centering
        \includegraphics[width=\textwidth,trim={0cm -12.9cm 0cm 0cm},clip]{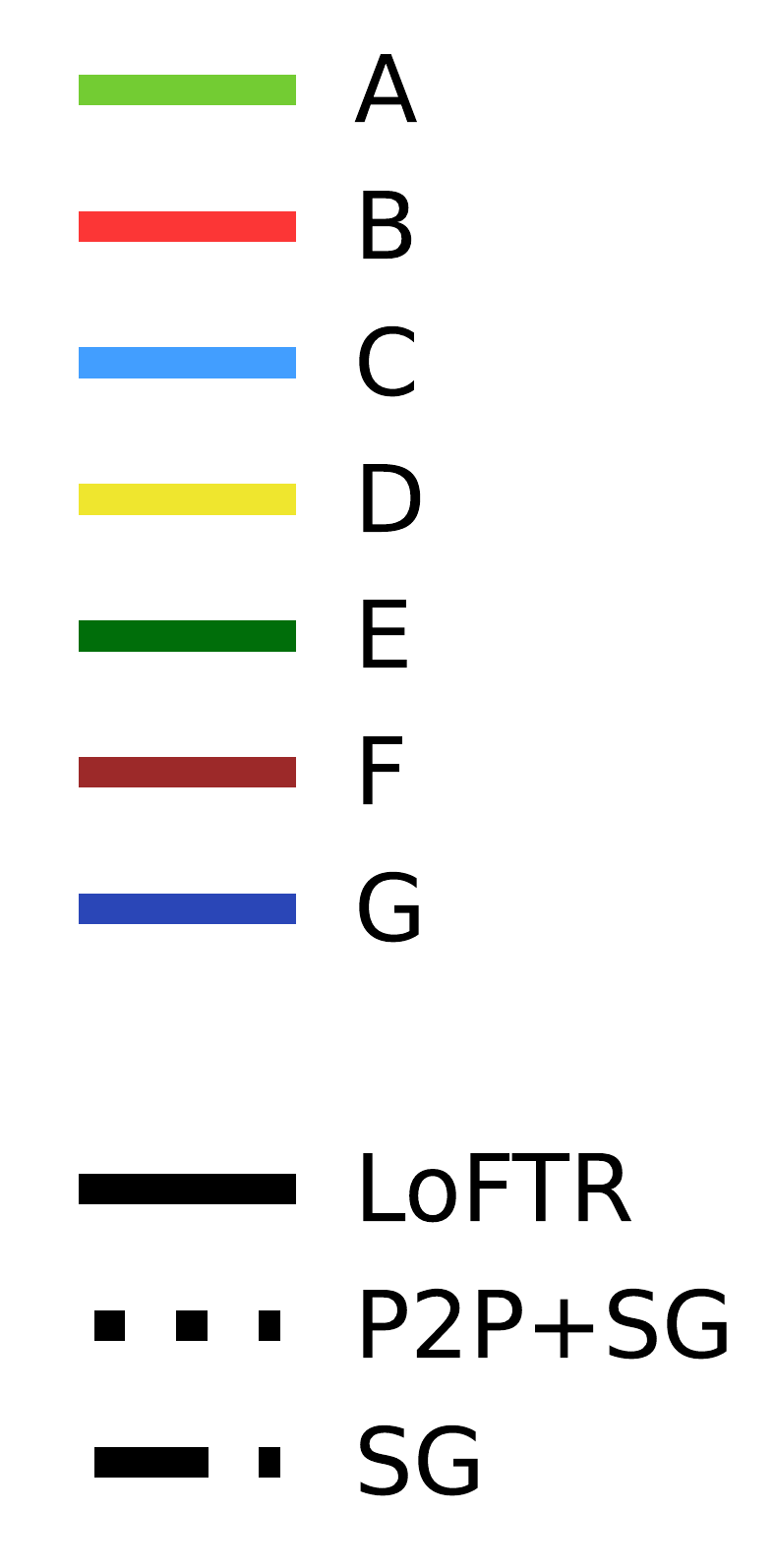}
    \end{subfigure}
    \vspace{-10pt}
    \caption{Cumulative histograms of the mean DCRE over all query images in a scene for the ground truth poses obtained via global alignment (GA) and local refinement (LR). We show the DCRE as the percentage of the image diagonal.}
    \label{fig:eval_dcre}
\end{figure*}

\section{Evaluation Protocol}
\label{sec:eval_protocol}
This paper studies how the level of detail of 3D meshes downloaded from the Internet impacts visual localization performance. 
The goal is to answer the question to which degree such 3D models can be used to replace the classical scene representations, constructed from images, used in the literature. 
The following describes the localization system we use and how we measure localization performance. 

\PAR{Visual localization.}
We use the state-of-the-art MeshLoc approach~\cite{Panek2022ECCV} for our experiments.
MeshLoc represents the scene via a 3D mesh and a set of database images with known poses. 
Image retrieval is used to find the top-$k$ most similar database images for each query.  
2D-2D matches between the query and the retrieved images are lifted to 2D-3D matches using the 3D mesh and the known poses of the database images. 
The camera pose of the query is then estimated from these 2D-3D matches using a P3P solver~\cite{Persson2018ECCV} inside RANSAC~\cite{Fischler81CACM} with local optimization~\cite{Lebeda2012BMVC}. 

The 3D meshes used in~\cite{Panek2022ECCV} align well with the real scene geometry. 
\cite{Panek2022ECCV} show that using renderings of the meshes instead of the original photos leads to comparable pose accuracy. 
They further show that even renderings of uncolored meshes can lead to accurate poses, as long as the meshes contain enough geometric detail.
This observation motivated us to explore the more abstract (less geometrically detailed) CAD models available on the Internet. 

\begin{figure*}[th!]
    \centering
    \begin{subfigure}[b]{0.28\textwidth}
        \centering
        \includegraphics[width=\textwidth,trim={-0.4cm 0cm -0.4cm 0.1cm},clip]{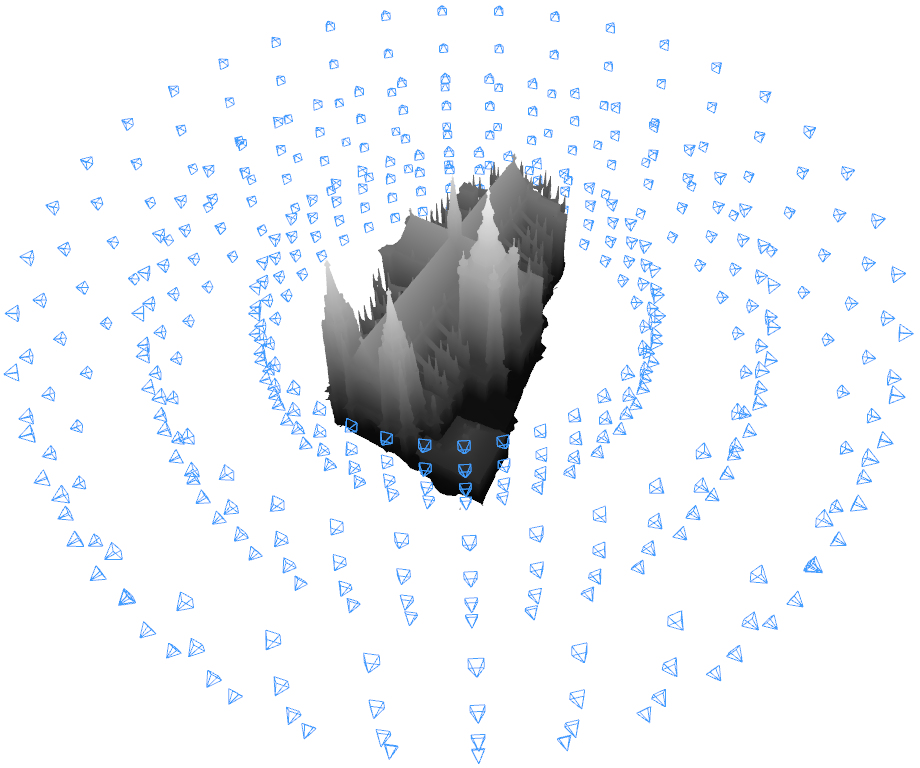}
        \vspace{-6pt}
        \caption{Synthetic cameras placement.}
        \label{fig:cam_poses}
    \end{subfigure}
    \hfill
    \begin{subfigure}[b]{0.68\textwidth}
    \includegraphics[width=\textwidth]{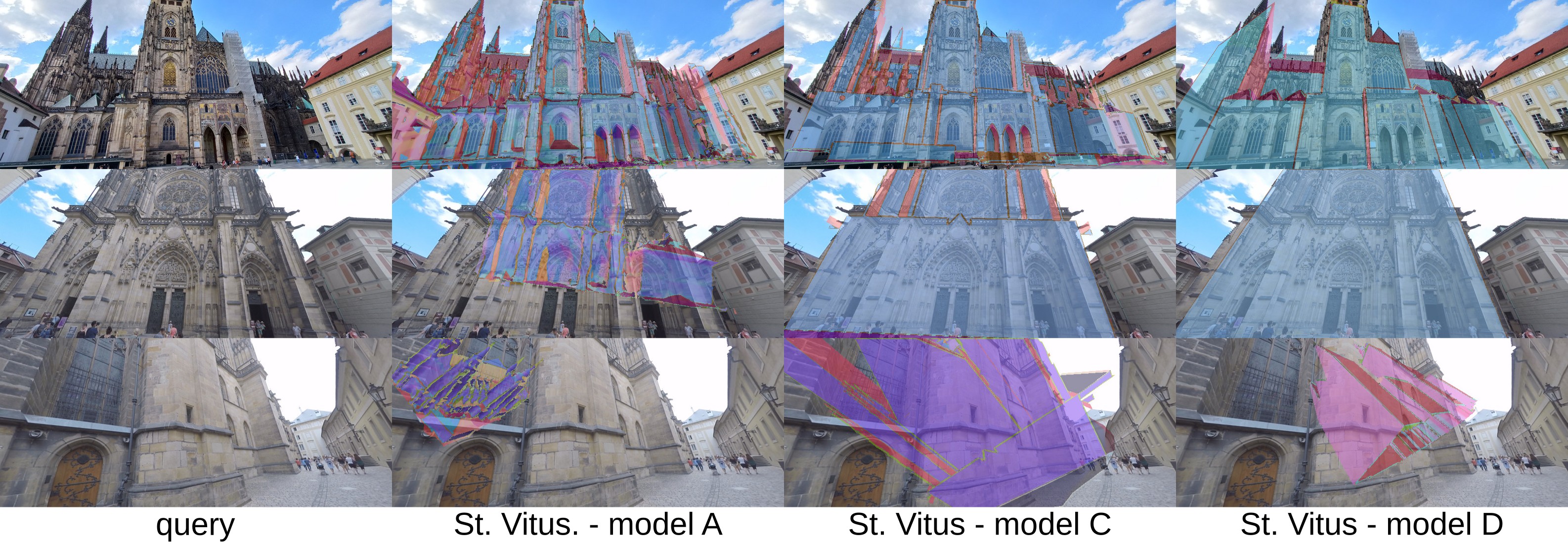}
        \centering
        \vspace{-6pt}
        \caption{Visualization of estimated poses for the St. Vitus Cathedral scene, obtained with LoFTR~\cite{Sun2021CVPR}.}
        \label{fig:st_vitus_align}
    \end{subfigure}
    \vspace{-6pt}
    \caption{The rendering setup for St. Vitus model and 3D model alignment based on estimated camera poses.}
\end{figure*}

In~\cite{Panek2022ECCV}, the synthetic images are rendered from the poses of the original database images.  
However, we only have a 3D model of the scene and no database images.
We thus use a simple approach to sampling camera poses around the model, from which we then render database images for each Internet model:  
We place cameras on the surfaces of multiple spheres with different radii, all centered around the center of gravity of the 3D model (\cf~\cref{fig:cam_poses}). 
All cameras are looking at this center.
The center and radii of the spheres, along with elevation limits and angular sampling periods, are manually adjusted to fit a particular model's geometry.

\PAR{Performance measures.}
A commonly used measure for localization performance computes the percentage of query images with camera pose estimates that are within given error thresholds of the ground truth~\cite{Sattler2018CVPR,Shotton2013CVPR,Brachmann2021ICCV,Jafarzadeh2021ICCV}. 
This requires that the scale of the scene is known, which is the case for the Aachen Day-Night v1.1 dataset~\cite{Sattler2012BMVC,Sattler2018CVPR,Zhang2020IJCV} that is used to define the queries for the Aachen scene. 
By aligning the CAD model to the MVS mesh obtained from the database images of the Aachen dataset, we can use the ground truth poses from~\cite{Sattler2018CVPR,Zhang2020IJCV} for evaluation. 

However, this measure is not directly applicable to the other datasets, where the scale of the scene is not given. 
Rather than measuring errors in meters or degrees, we thus measure reprojection errors, which are independent of the scale of the scene.
We use the Dense Correspondence Re-Projection Error (DCRE) measure from~\cite{Wald2020ECCV}. 
Given a ground truth and an estimated pose, and a depth map of the 3D model from the view of the ground truth pose, the DCRE is measured as follows~\cite{Wald2020ECCV}: for each pixel in the depth map, we obtain a 3D point in the world coordinate system of the 3D model.
Each 3D point is then projected into the image using the ground truth and the estimated pose, resulting in a set of 2D-2D correspondences. 
We measure the mean Euclidean distance between correspondences (mean DCRE) and the maximum Euclidean distance between correspondences (max. DCRE). 
Both DCRE variants measure how the change in pose between ground truth and estimate affects the quality of the alignment of the pose \wrt to the 3D model. 
Smaller DCREs correspond to better alignments and are a direct measure of pose quality for AR applications. 

We consider two ways to define ground truth poses required for measuring the DCREs: 
For each scene, we rigidly align the Internet models to a MVS model computed from the query images (the query model). 
For the alignment, we use ICP~\cite{rusinkiewicz2001efficient}, starting from a manual initialization using 3D point correspondences.
This alignment defines the poses of the query images \wrt the Internet models and we use these poses as one set of ground truth poses. 
We refer to these poses as \textit{global alignment (GA) poses}. 

DCREs computed \wrt to the GA poses measure how well the poses estimated via visual localization align with the poses from which the photos were taken in the scene. 
The underlying assumption is that the Internet model can be aligned well to the query model using a rigid transformation.
However, the geometry of the query and the Internet models might differ, \eg, the ratios of the 3D model width and height might be different.
Thus, a rigid alignment between the model types might be insufficient and the GA poses will not reflect the ``optimal" pose of the query \wrt the Internet model. 
We thus consider a second set of ground truth poses, obtained by refining the GA poses \wrt each Internet model: 
given the GA pose and the depth map (generated by rendering the query model) for a query image, we align this depth map to the Internet model via ICP~\cite{rusinkiewicz2001efficient}. 
We will refer to these poses as \emph{locally refined (LR) poses}. 

\begin{figure*}[t!]
    \centering
    \begin{subfigure}[b]{0.2178\textwidth}
        \centering
        \includegraphics[width=\textwidth,trim={0 0 0 0},clip]{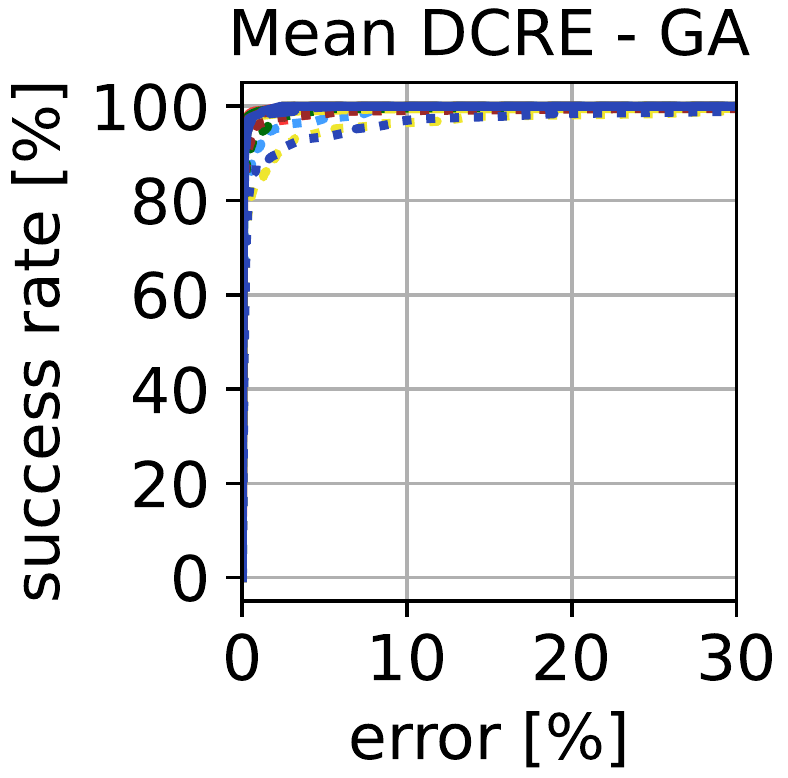}
        \caption{Notre Dame}
    \end{subfigure}
    \begin{subfigure}[b]{0.16\textwidth}
        \centering
        \includegraphics[width=\textwidth,trim={0 0 0 0},clip]{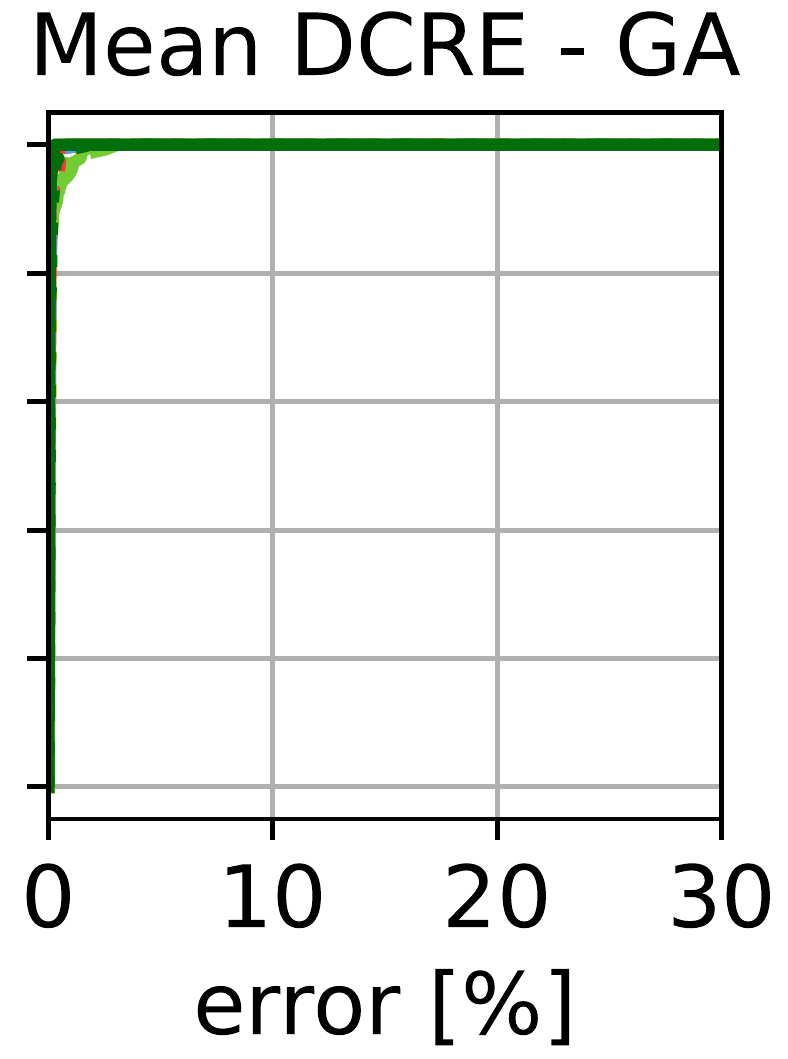}
        \caption{Pantheon}
    \end{subfigure}
    \begin{subfigure}[b]{0.16\textwidth}
        \centering
        \includegraphics[width=\textwidth,trim={0 0 0 0},clip]{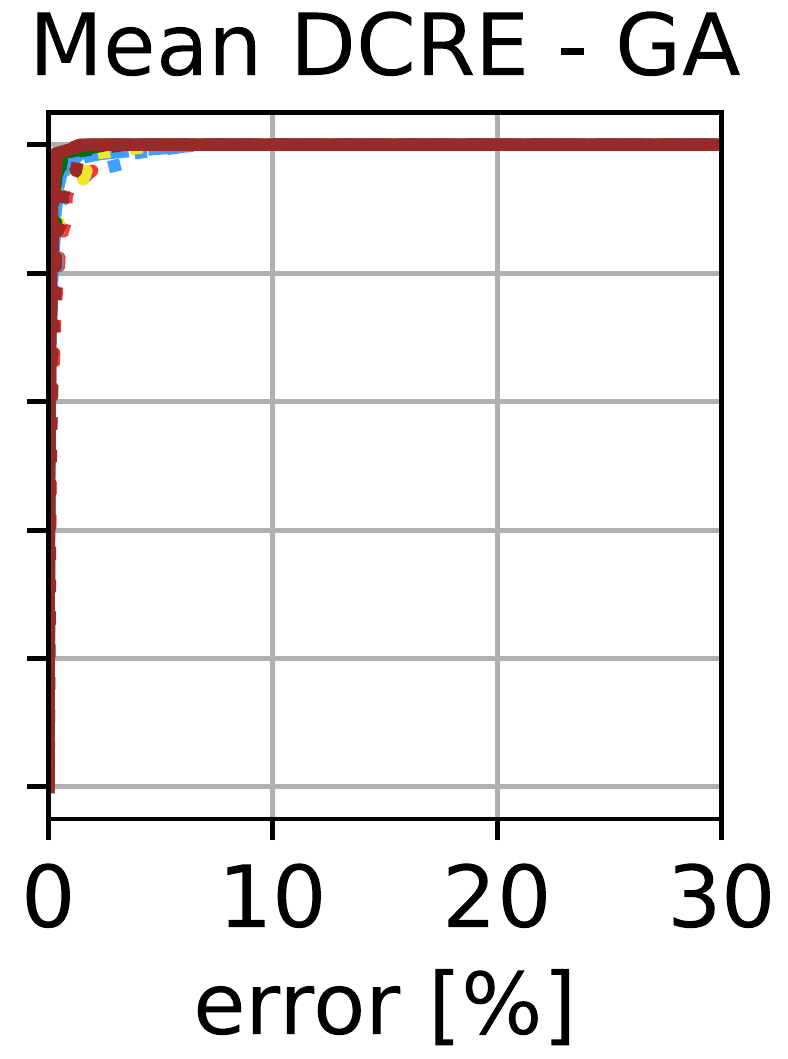}
        \caption{Reichstag}
    \end{subfigure}
    \begin{subfigure}[b]{0.16\textwidth}
        \centering
        \includegraphics[width=\textwidth,trim={0 0 0 0},clip]{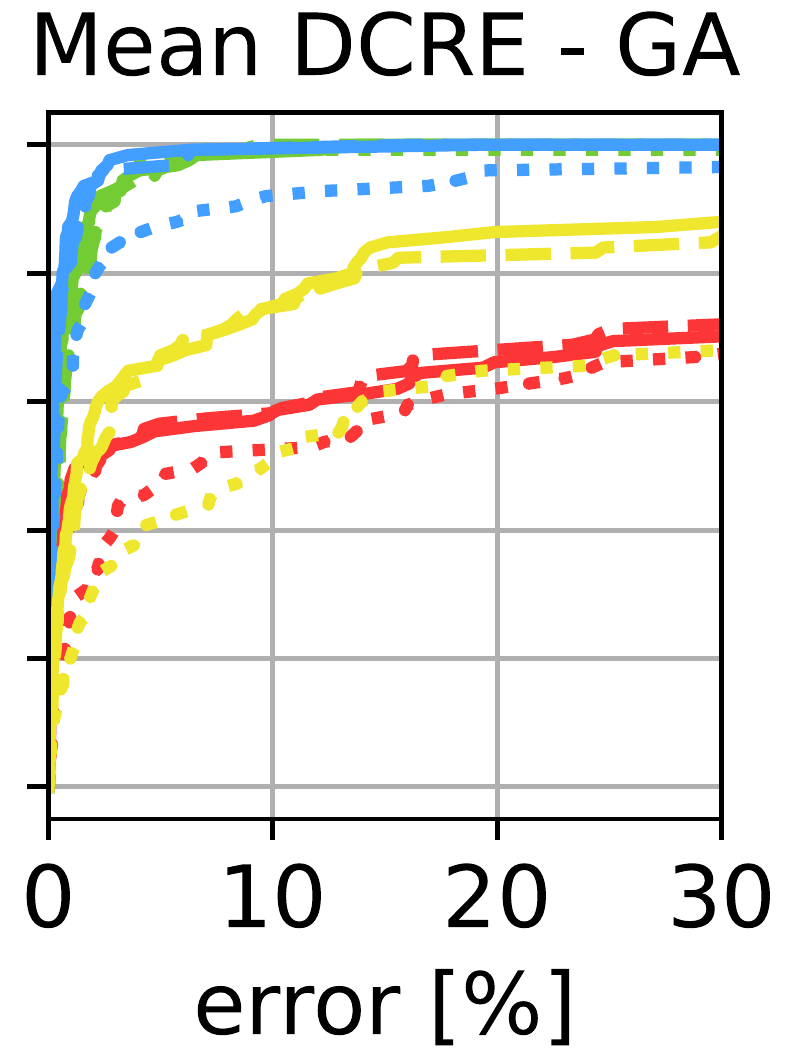}
        \caption{St. Peter's square}
    \end{subfigure}
    \begin{subfigure}[b]{0.16\textwidth}
        \centering
        \includegraphics[width=\textwidth,trim={0 0 0 0},clip]{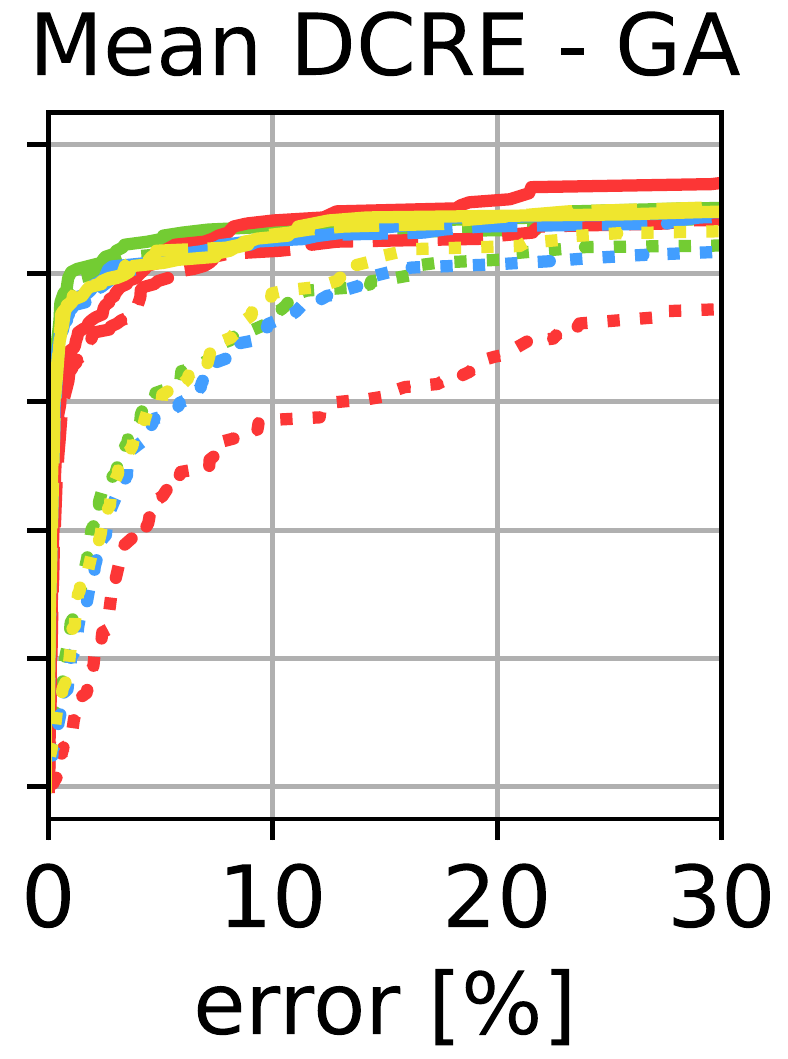}
        \caption{St. Vitus}
    \end{subfigure}
    \begin{subfigure}[b]{0.11\textwidth}
        \centering
        \includegraphics[width=\textwidth,trim={0cm -0.5cm 0cm 0cm},clip]{figures/example_legend_2x1.pdf}
    \end{subfigure}
    \vspace{-10pt}
    \caption{Isolating the impact of geometric fidelity by combining real images with geometry from the Internet models. We show cumulative histograms of the mean DCRE, as a percentage of the image diagonal, over all query images in a scene. See Sec.~\ref{sec:supp_exps} for results with LR.}
    \label{fig:eval_dcre_orig_imgs_cad_depth}
\end{figure*}

\section{Experimental Evaluation}
\label{sec:experiments}
We perform three sets of experiments: 
(1) we measure the level of geometric fidelity of the different models. This will later allow us to conclude how the level of appearance fidelity affects pose accuracy. 
(2) we measure localization accuracy via DCREs for the two definitions of ground truth poses, as well as pose accuracy in meters and degrees on the Aachen scene.
(3) we isolate the impact of geometric fidelity on the localization process.

\PAR{Measuring geometric fidelity.} 
We measure how accurately the Internet models capture the real geometry of the scene. 
To this end, we compute distances in 3D between the query model and the Internet models, with the query model serving as an approximation to the true scene geometry. 
For each 3D point in the query model mesh, we find the nearest vertex in an Internet model mesh. 
We appropriately subdivide the Internet models to handle scenarios where an Internet model consists of only a few large polygons. 
The query model often shows only part of the Internet model, which is why we do not measure distances in the other direction. 

Fig.~\ref{fig:three graphs} shows cumulative histograms over these distances. 
Here, smaller distances indicate a higher level of geometric fidelity. 
Note that for all scenes except Aachen, the 3D distances have no physical meaning. 
Within a scene, we can compare different models as they are all aligned to the query model and thus share the same scale. 
However, comparing models between scenes is not possible. 
For Aachen, due to the known scale of the Aachen Day-Night v1.1~\cite{Sattler2018CVPR,Sattler2012BMVC,Zhang2020IJCV} model, one model unit corresponds to one meter. 
As can be seen, the level of geometric fidelity of the Internet models used in our benchmark varies significantly. 

\PAR{Measuring localization performance via DCREs.}
We localize real images against the Internet models using the MeshLoc~\cite{Panek2022ECCV} pipeline. 
We evaluate different features and matchers inside MeshLoc:  
LoFTR~\cite{Sun2021CVPR}, SuperGlue~\cite{Sarlin2020CVPR}, and the combination of Patch2Pix~\cite{Zhou2021CVPR} and SuperGlue~\cite{Sarlin2020CVPR}.
The main paper only shows mean DCRE results. 
Max. DCRE results are shown in the Sec.~\ref{sec:supp_exps}.

\cref{fig:eval_dcre} shows cumulative distributions of the mean DCRE over all query images for a scene. Comparing the results with those shown in \cref{fig:three graphs} and the visualization from \cref{fig:mesh_all}, we can observe the following: 
(1) the best results are achieved by Internet models with both (relatively) high geometric and appearance level of detail: the Notre Dame A and B, and Pantheon A models are among the geometrically most accurate ones and provide high-quality textures. 
For all three, most images have a mean DCRE of 10\% or less, which is comparable to what has been reported in the literature for scene representations created from RGB(-D) images (\cf supp. mat. of~\cite{Brachmann2021ICCV}). 
This clearly shows that the approach of using models downloaded from the Internet as a scene representation for visual localization is feasible and can lead to high pose accuracy. 
\noindent (2) higher appearance fidelity can significantly compensate for a lower geometric fidelity. 
\Eg, the Pantheon B and Reichstag B models are both significantly less accurate than other models from the same scene. 
Still, both models lead to very good localization performance. 
\noindent (3) if there are large discrepancies between the Internet model and the real world, then localization quickly fails. 
\Eg, none of the models for the St. Vitus and St. Peter's square scenes performs well. 
For the St. Vitus A and B models, this can be explained by the fact that the models were generated from drone imagery, leading to severe texture and model distortions. 
As a result, images taken close to the scene from the ground lead to localization failures (\cf \cref{fig:st_vitus_align}). 
Similarly, the St. Vitus C and D models lack the geometric and appearance detail necessary to localize such images. 
However, it is still possible to localize images taken relatively far from the scene, where less geometric and appearance detail is needed (\cf \cref{fig:st_vitus_align}). 
\noindent (4) we generally observe a higher performance with the global alignment (GA) ground truth poses compared to the locally refined (LR) poses. 
We attribute this to a lack of geometric detail in some of the models (in particular for the Reichstag scene), which allows the ICP algorithm~\cite{rusinkiewicz2001efficient} to significantly alter poses. 
\noindent (5) there can be significant differences in performance depending on the type of local features used, \eg, the Reichstag E and St.~Peter's square D models. 
Both these models do not have any texture, indicating that this is still a very challenging problem. 
Since such 3D models are often very compact in terms of memory footprint, being able to use them for localization is a problem of interest.

\begin{table}[t!]
    \centering
    \scriptsize{
    \renewcommand{\arraystretch}{0.8}
    \begin{tabular*}{\linewidth}{@{\extracolsep{\fill}}lll@{\extracolsep{\fill}}}
        \hline\noalign{\smallskip}
        method & day & night \\ 
        \hline\noalign{\smallskip}
        CAD - LoFTR~\cite{Sun2021CVPR} & 0.4 / 6.8 / 71.3 & 0.6 / 7.0 / 72.5 \\
        CAD - Patch2Pix~\cite{Zhou2021CVPR} + SuperGlue~\cite{Sarlin2020CVPR} & 0.7 / 5.6 / 70.0 & 1.1 / 7.6 / 74.3 \\
        CAD - SuperGlue~\cite{Sarlin2020CVPR} & 0.6 / 6.1 / 72.2 & 1.8 / 8.8 / 73.3	\\
        \hline
    \end{tabular*}
    \vspace{-6pt}
    }
    \caption{Localization results for the Aachen scene for different types of local features.
    We report the \% of query images localized within (0.25m, 2°) / (0.5m, 5°) / (5m, 10°) of the ground truth pose.
    We report results only for queries that observe the CAD model.
    }
    \label{tab:aachen_results}
\end{table}
\begin{figure}[t]
    \centering
    \begin{subfigure}[b]{0.22\textwidth}
        \centering
        \includegraphics[width=\textwidth,trim={0 0 110 0},clip]{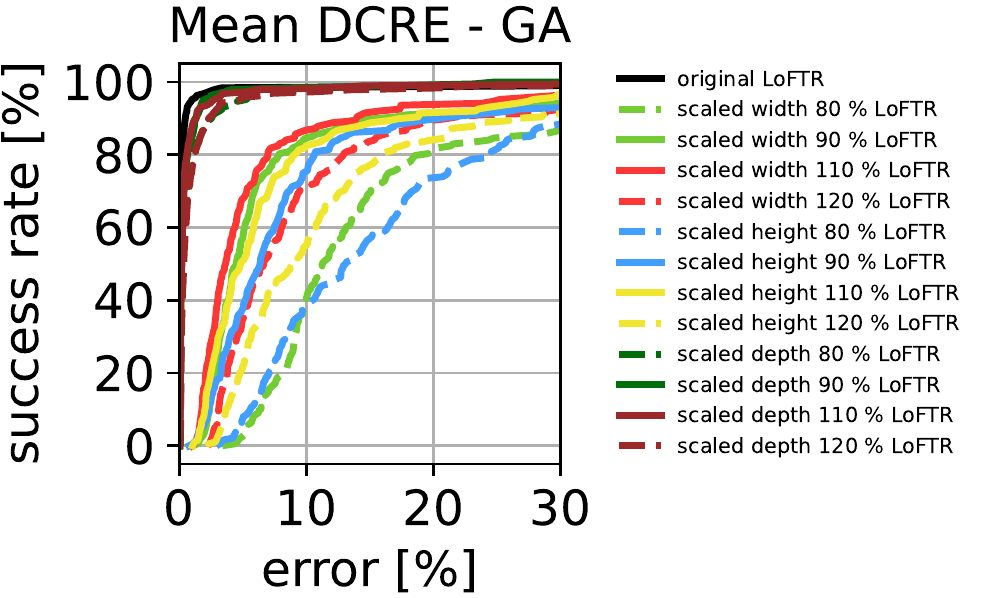}
    \end{subfigure}
    \hfill
    \begin{subfigure}[b]{0.19\textwidth}
        \centering
        \includegraphics[width=\textwidth,trim={180 40 0 0},clip]{figures/dcre_stretch/notre_dame_de_paris_stretch_20_loftr_eval_dcre_mean_diagnorm_inv_LOW.pdf}
    \end{subfigure}
    \vspace{-6pt}
    \caption{Influence of non-uniform scaling on the alignment success in the case of Notre Dame A model and LoFTR matcher.}
    \label{fig:scaling_exp}
\end{figure}

\PAR{Measuring via absolute errors.} 
\cref{tab:aachen_results} reports localization performance using the benchmark from~\cite{Sattler2018CVPR,Toft2022TPAMI} for evaluation.
The CAD model allows coarse localization despite its simplistic geometry and low-quality textures. 
Thus, this simple model is a viable alternative to a more detailed mesh if coarse pose accuracy is sufficient. 

\PAR{Isolating the impact of geometric fidelity.}
There are two main failure cases when localizing real images against Internet models: 
(1) Due to significant changes in appearance, insufficiently many matches are found. 
(2) Camera pose estimation computes a rigid alignment between the image and the Internet model. 
The pose estimation stage thus may fail if the geometry of the Internet model is too distorted. 
In this experiment, we isolate the impact of geometric fidelity on the localization.
The Internet models are only used to lift the 2D-2D matches between the real images to 2D-3D matches that can be used for pose estimation. 
As can be seen from \cref{fig:eval_dcre_orig_imgs_cad_depth}, the limiting factor often is the feature matching stage. 
Given enough feature matches, even geometrically less accurate models can produce relatively accurate poses (\eg, see results for St. Peter's Square in \cref{fig:eval_dcre} and \cref{fig:eval_dcre_orig_imgs_cad_depth}). 
Thus, research on features and matchers that can handle this challenging task is a very promising direction for future work. 

To better measure the influence of geometry inconsistency on the success of localization, we simulate different levels of distortion for one of the most precise models (Notre Dame A). 
We scale the model in a non-uniform way and use renderings of the resulting model for feature matching and pose estimation. Details on the scaling procedure are provided in Sec.~\ref{sec:geom_fidel}.
As can be seen in \cref{fig:scaling_exp}, increasing scale inconsistency significantly reduces localization success. The depth scaling does not influence the localization as much because all the query images were captured from the front facade side of the building, so only the depth changes of the more minor facade elements, such as doors, windows, or pillars, are visible. 
The results show that pose estimation strategies that actively account for non-uniform scaling are required for heavily distorted models.

We also investigate the influence of geometry and appearance simplification in Sec.~\ref{sec:simplification}.

\section{Conclusion}
\label{sec:conclusion}
In this paper, we have explored an alternative for scene representations in visual localization pipelines, \ie, 3D models that are readily available on the Internet. 
Such a representation is intriguing, as it, \eg, offers the promise of setting up AR experiences for users that are visiting places that the author creating the experience never visited. Yet, it also comes with many challenges due to imperfections of these 3D modes. We studied how such imperfections affect localization accuracy.
Detailed experiments show
that 3D models from the Internet represent a promising new
category of scene representations, which at the same time also open new directions for improvement in visual localization. 
In particular, research on matching real images against more abstract scene representation is an interesting direction for future work. 
To foster research on this interesting task, we publish the benchmark (\href{https://v-pnk.github.io/cadloc/}{v-pnk.github.io/cadloc}).

{\small
\noindent\textbf{Acknowledgements. }
This work was supported by
the European Regional Development Fund under project IMPACT (No. CZ.02.1.01/0.0/0.0/15\_003/0000468), 
a Meta Reality Labs research award under project call 'Benchmarking City-Scale 3D Map Making with Mapillary Metropolis', 
the Grant Agency of the Czech Technical University in Prague (No. SGS21/119/OHK3/2T/13),
the OP VVV funded project CZ.02.1.01/0.0/0.0/16\_019/0000765 “Research Center for Informatics”,
and the Czech Science Foundation (GAČR) EXPRO Grant No. 23-07973X.
}

\begin{appendix}
\section{Supplementary material}
This supplementary material is organized as follows:
Sec.~\ref{sec:model_issues} describes the practical issues we encountered with the downloaded Internet models, mentioned in Sec.~\ref{sec:datasets}.
Sec.~\ref{sec:supp_exps} shows mean and maximum Dense Correspondence Re-Projection Error (DCRE) plots for both versions of the ground truth (Sec.~\ref{sec:experiments}).
Sec.~\ref{sec:geom_fidel} describes in detail the setup of the experiments on the evaluation of geometric fidelity. It shows extended results from Sec.~\ref{sec:experiments} for both the experiment with the database of real images and the database images rendered from a stretched 3D model.
Sec.~\ref{sec:simplification} presents experiment on the influence of simplification of model geometry and appearance on localization accuracy.

\section{Issues with the Internet Models}
\label{sec:model_issues}

\begin{figure*}[t]
    \centering
    \parbox{\textwidth}{
    \begin{subfigure}[b]{0.32\textwidth}
    \centering
        \includegraphics[width=\textwidth]{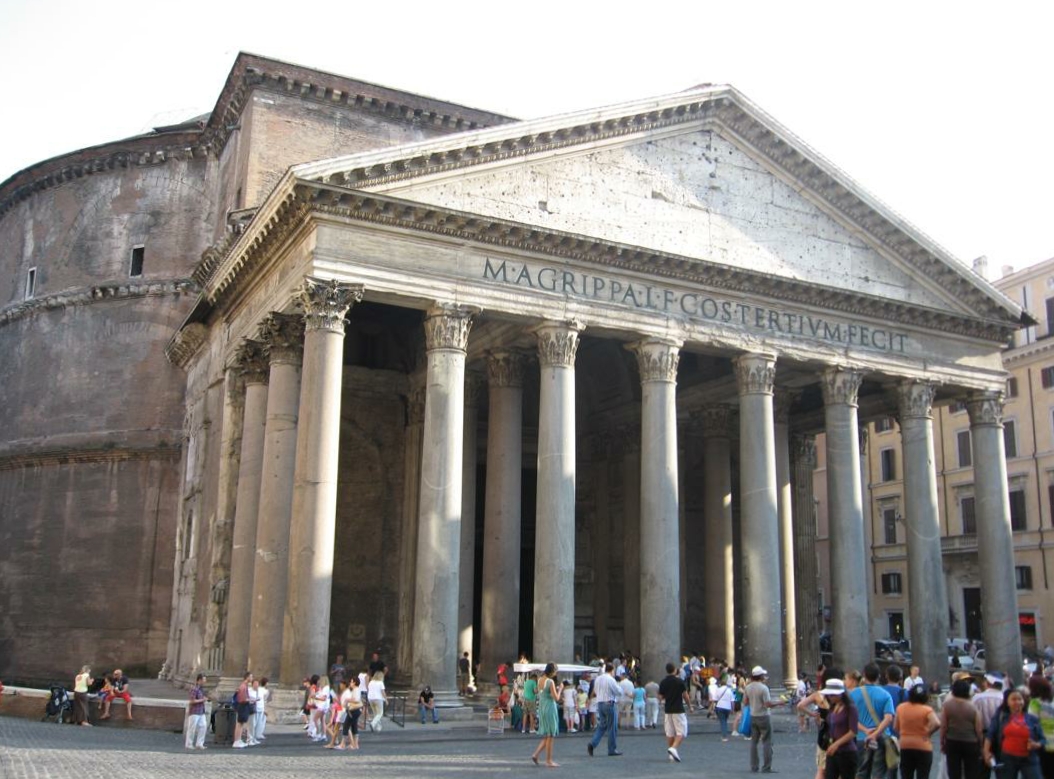}
        \caption{The real state of the landmark}
    \end{subfigure}
    \hfill
    \begin{subfigure}[b]{0.32\textwidth}
    \centering
        \includegraphics[width=\textwidth]{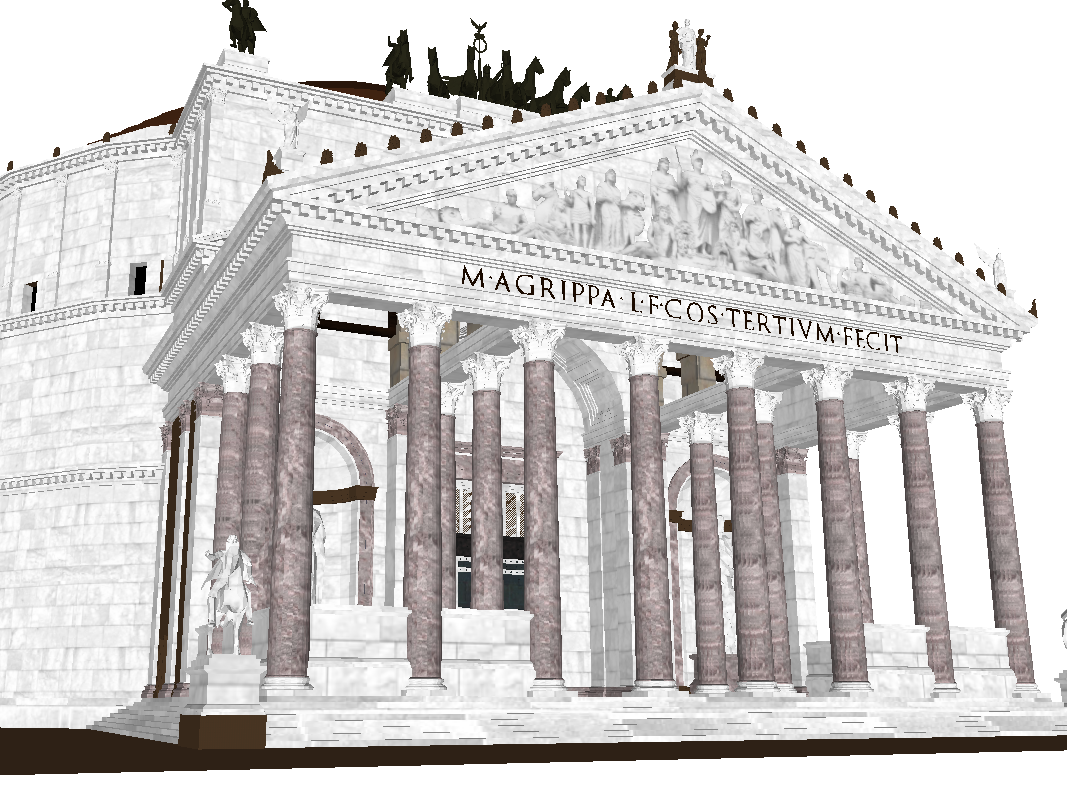}
        \caption{Pantheon - model C}
    \end{subfigure}
    \hfill
    \begin{subfigure}[b]{0.32\textwidth}
    \centering
        \includegraphics[width=\textwidth]{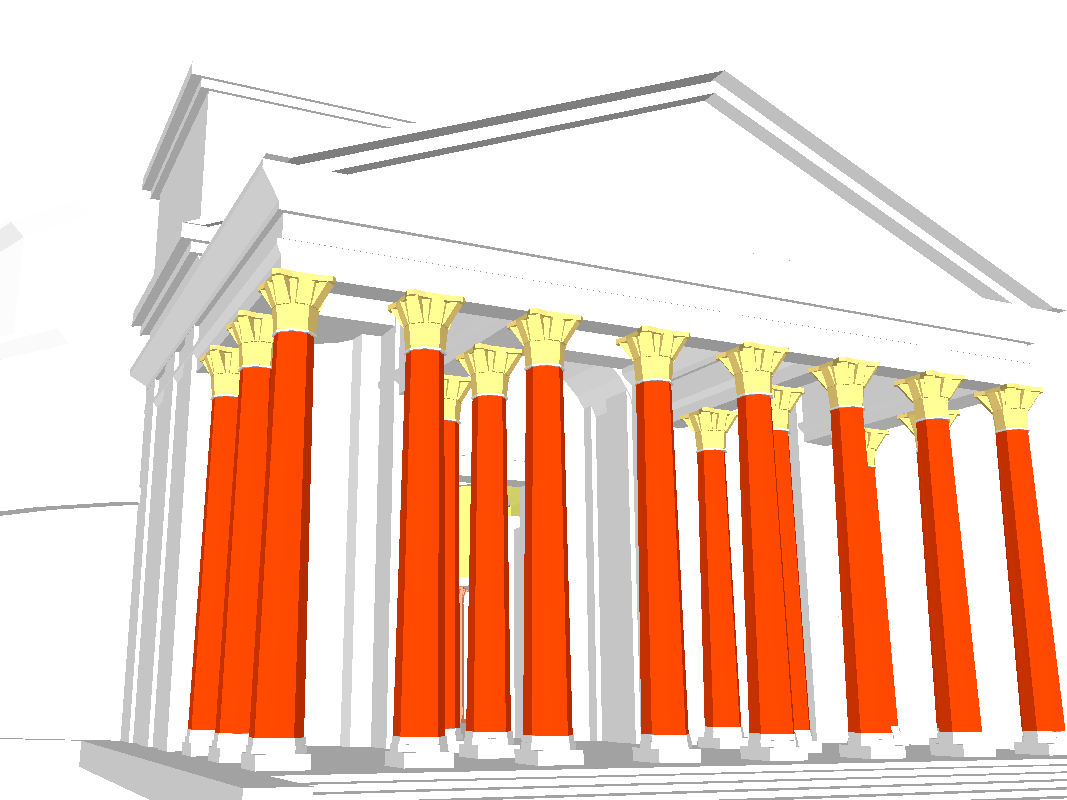}
    \caption{Pantheon - model D}
    \end{subfigure}
    \caption{Comparison of the real state of the Pantheon landmark to  models containing generic textures.}
    \label{fig:low_fidelity_meshes}
    }
\end{figure*}

\begin{figure*}[t]
    \centering
    \parbox{\textwidth}{
    \begin{subfigure}[b]{0.32\textwidth}
    \centering
        \includegraphics[width=\textwidth,trim={7.0cm 0.0cm 5.0cm 0.0cm},clip]{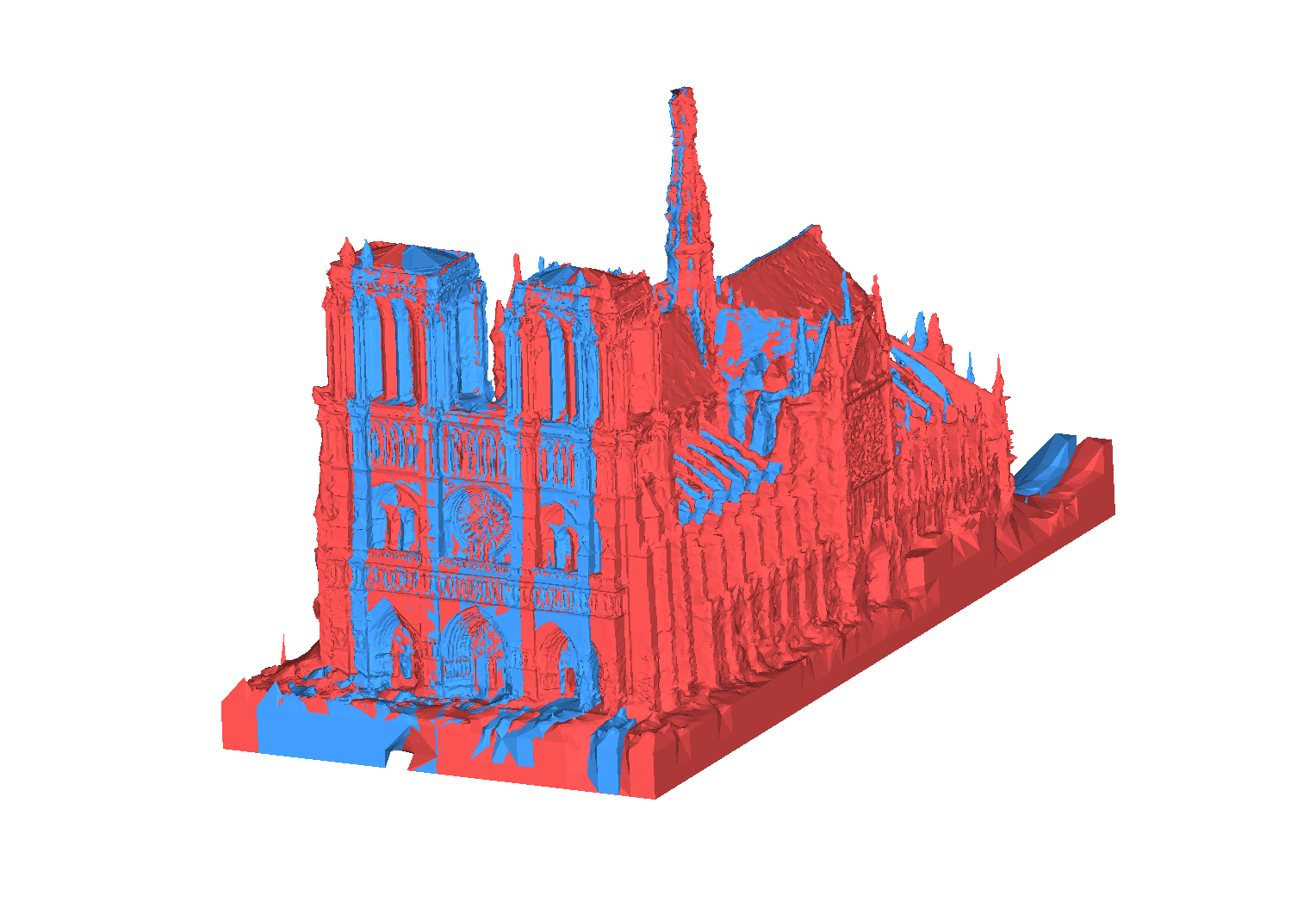}
        \caption{Stretching model width (120\%)}
    \end{subfigure}
    \hfill
    \begin{subfigure}[b]{0.32\textwidth}
    \centering
        \includegraphics[width=\textwidth,trim={7.0cm 0.0cm 5.0cm 0.0cm},clip]{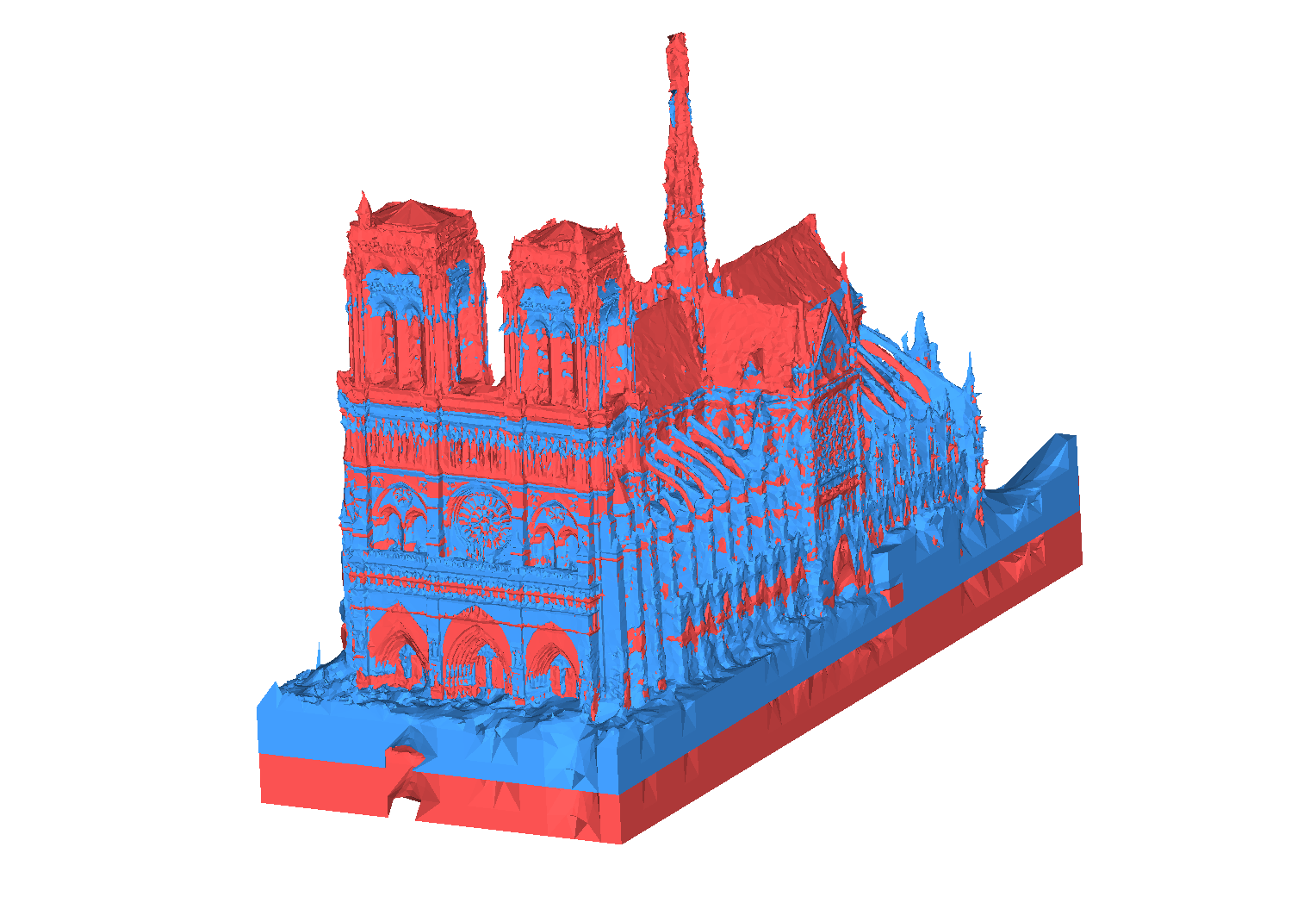}
        \caption{Stretching model height (120\%)}
    \end{subfigure}
    \hfill
    \begin{subfigure}[b]{0.32\textwidth}
    \centering
        \includegraphics[width=\textwidth,trim={7.0cm 0.0cm 5.0cm 0.0cm},clip]{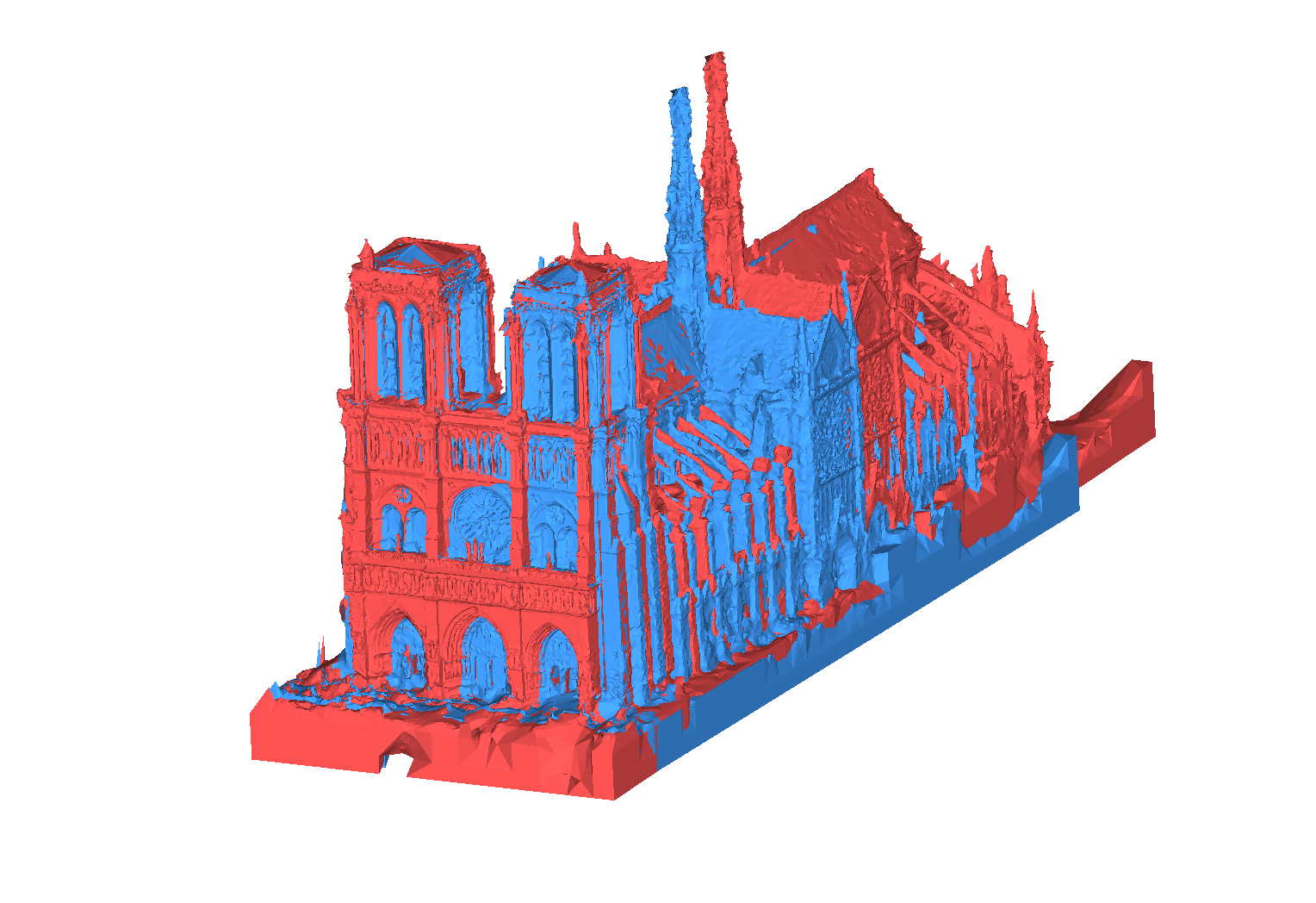}
    \caption{Stretching model depth (120\%)}
    \end{subfigure}
    \caption{Visualization of the non-uniform scaling used for the evaluation of the impact of geometric fidelity. Original model in \textcolor{meshblue}{blue}, scaled model in \textcolor{meshred}{red}.}
    \label{fig:stretch_comparison}
    }
\end{figure*}

\begin{figure*}[t]
    \centering
    \includegraphics[width=\textwidth,trim={0 0 0 0},clip]{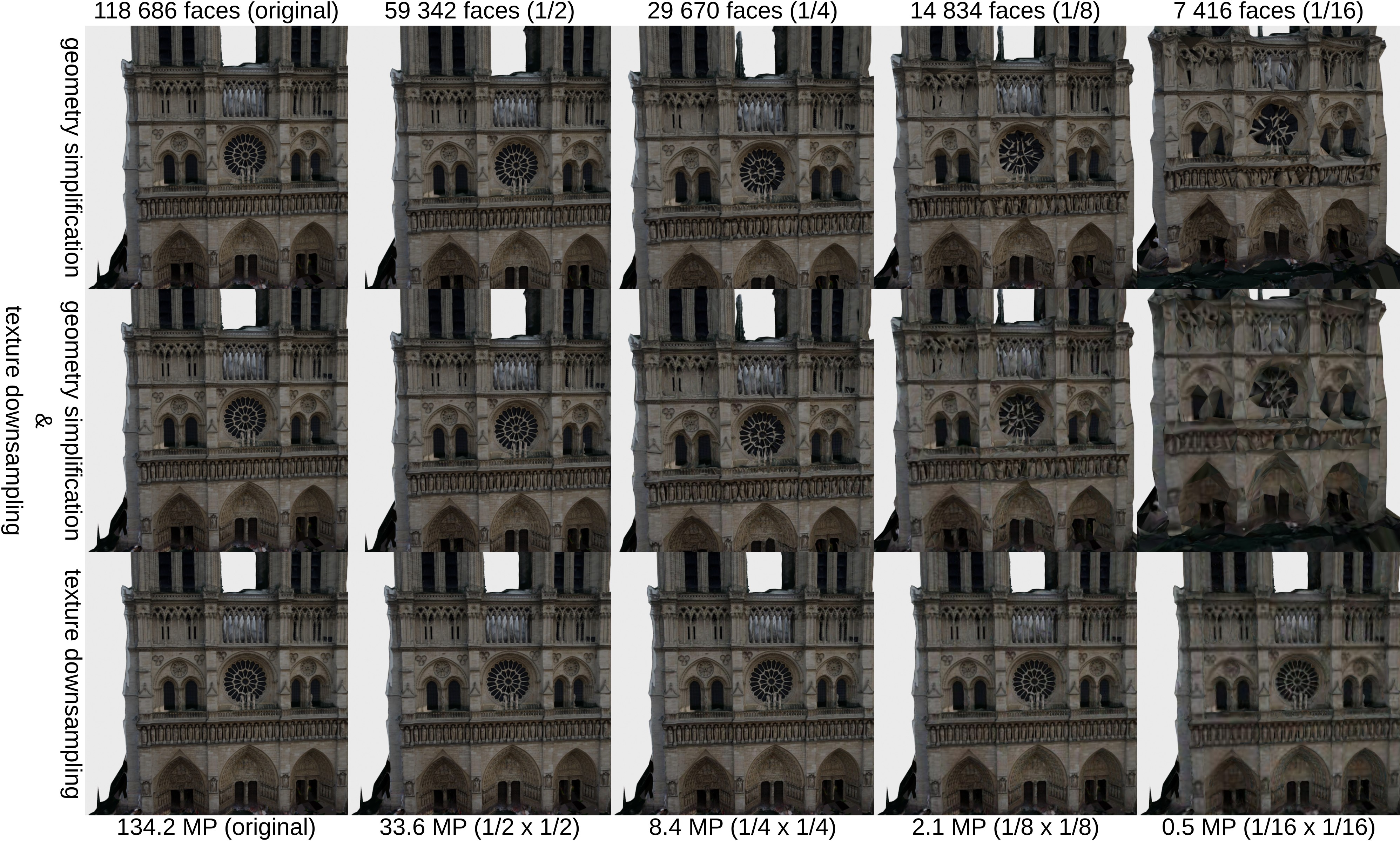}
    \vspace{-20pt}    
    \caption{Renderings of the simplified Notre Dame A model. The first column contains renderings of the original model. All other columns correspond to models with reduced geometric and / or texture detail.}
    \label{fig:simplification_comparison}
\end{figure*}

\begin{figure*}
    \centering
    \includegraphics[width=\textwidth,trim={-2cm 0cm -2cm 0cm},clip]{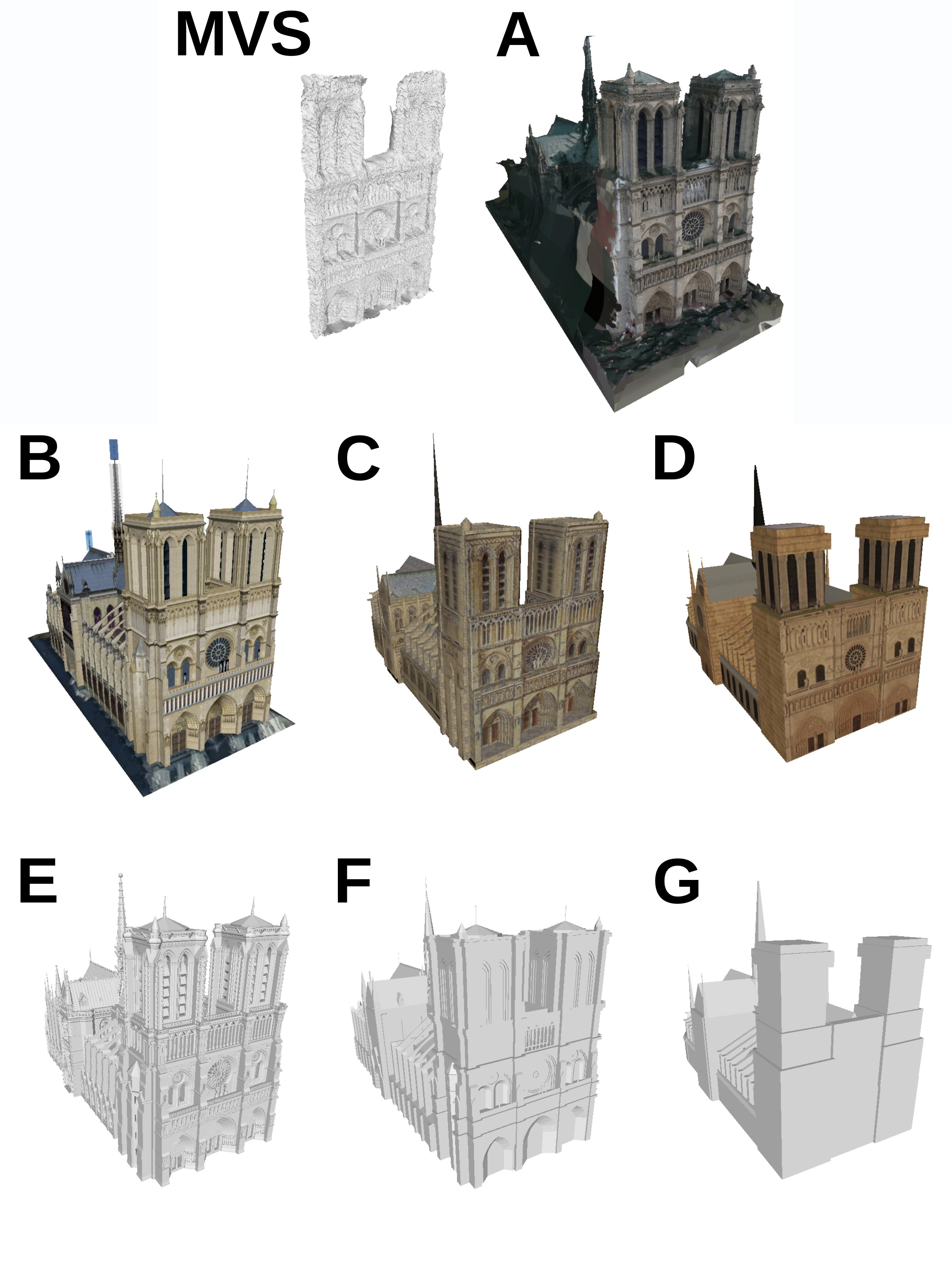}
    \caption{Enlarged Notre Dame models from Fig.~\ref{fig:mesh_all}.}
    \label{fig:models_notre_dame}
\end{figure*}

\begin{figure*}
    \centering
    \includegraphics[width=\textwidth,trim={0cm 0cm 0cm 0cm},clip]{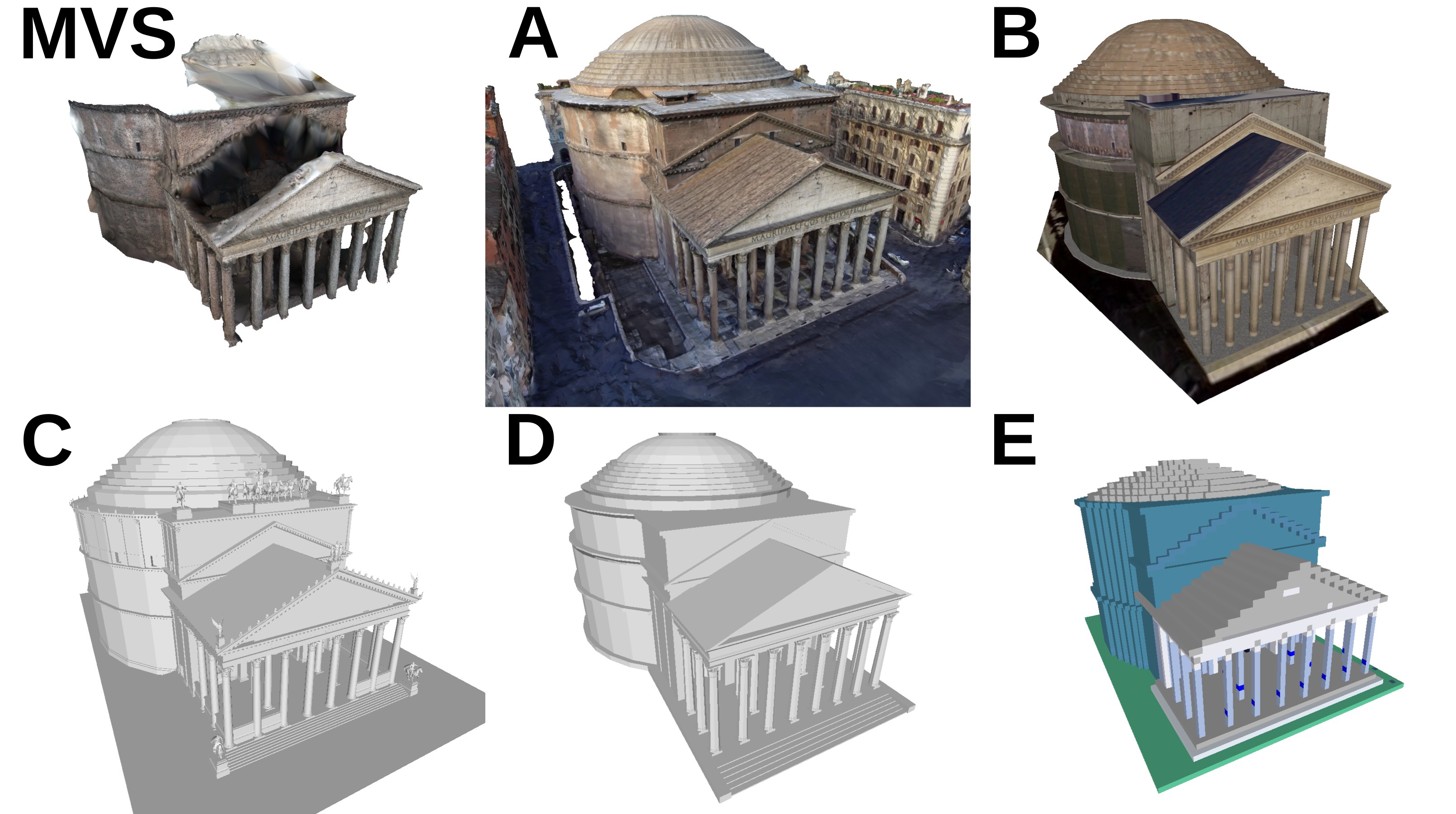}
    \caption{Enlarged Pantheon models from Fig.~\ref{fig:mesh_all}.}
    \label{fig:models_pantheon}
\end{figure*}

\begin{figure*}
    \centering
    \includegraphics[width=\textwidth,trim={-2cm 0cm -2cm 0cm},clip]{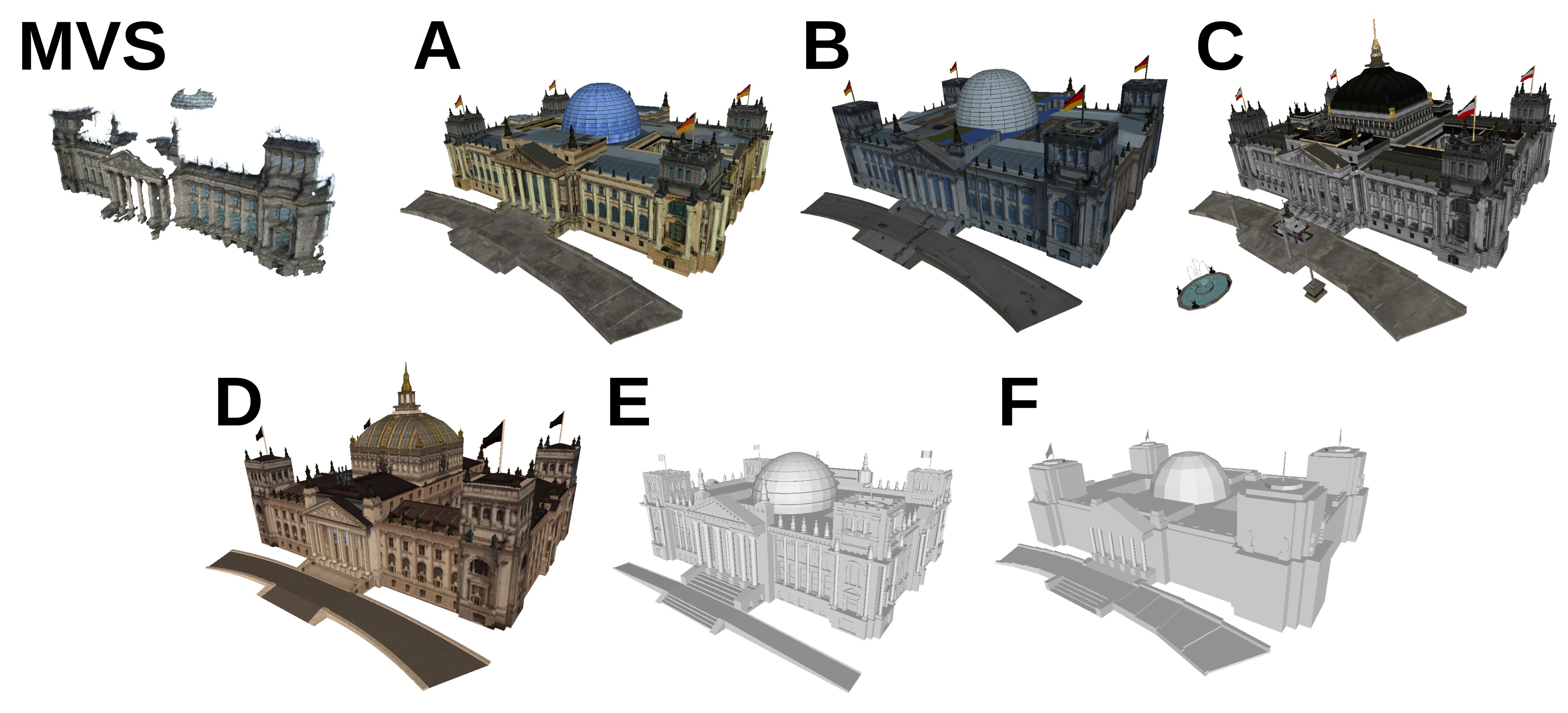}
    \caption{Enlarged Reichstag models from Fig.~\ref{fig:mesh_all}.}
    \label{fig:models_reichstag}
\end{figure*}

\begin{figure*}
    \centering
    \includegraphics[width=\textwidth,trim={-2cm 0cm -2cm 0cm},clip]{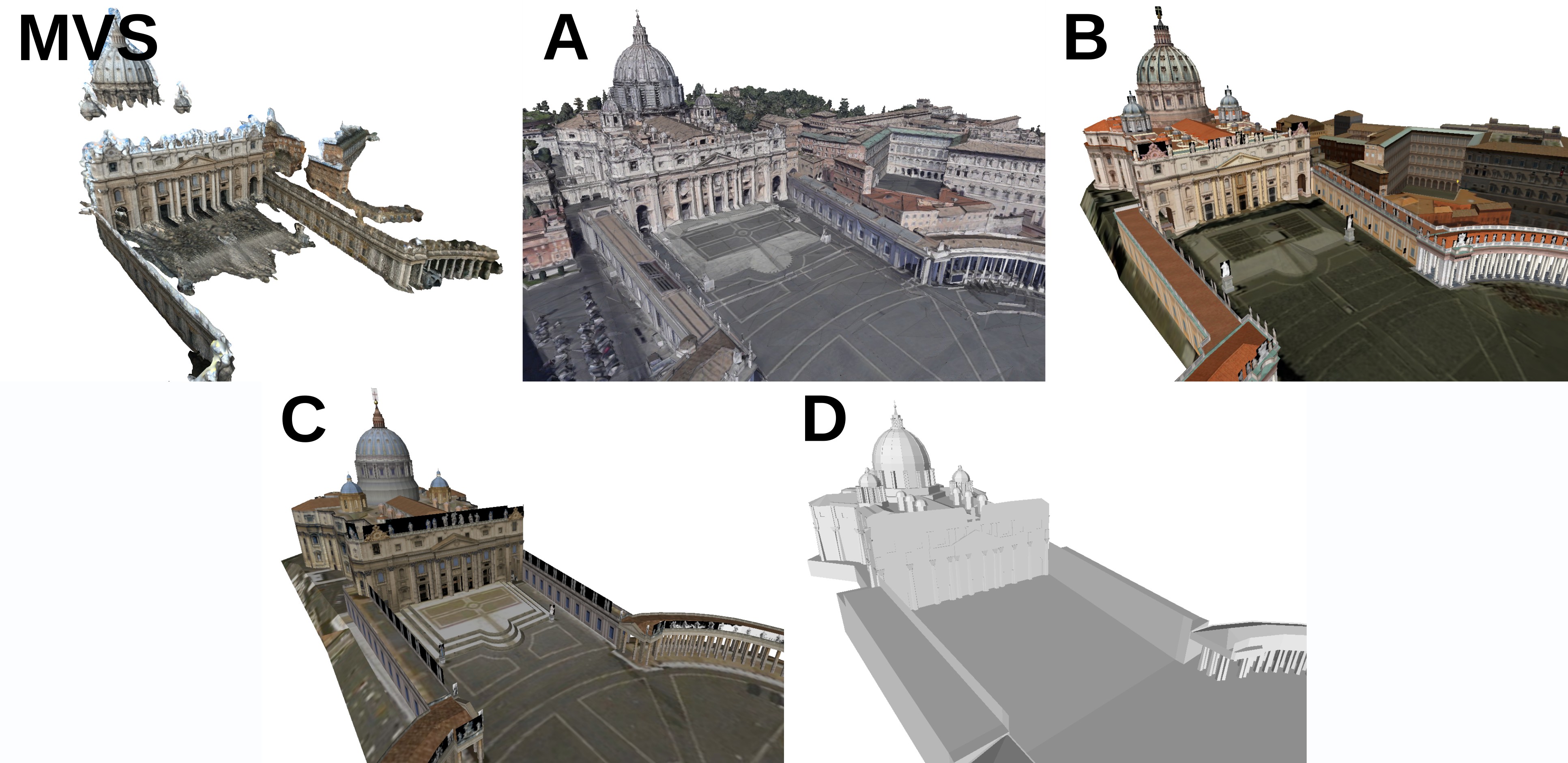}
    \caption{Enlarged St. Peter's Square models from Fig.~\ref{fig:mesh_all}.}
    \label{fig:models_st_peters_square}
\end{figure*}

\begin{figure*}
    \centering
    \includegraphics[width=\textwidth,trim={-2cm 0cm -2cm 0cm},clip]{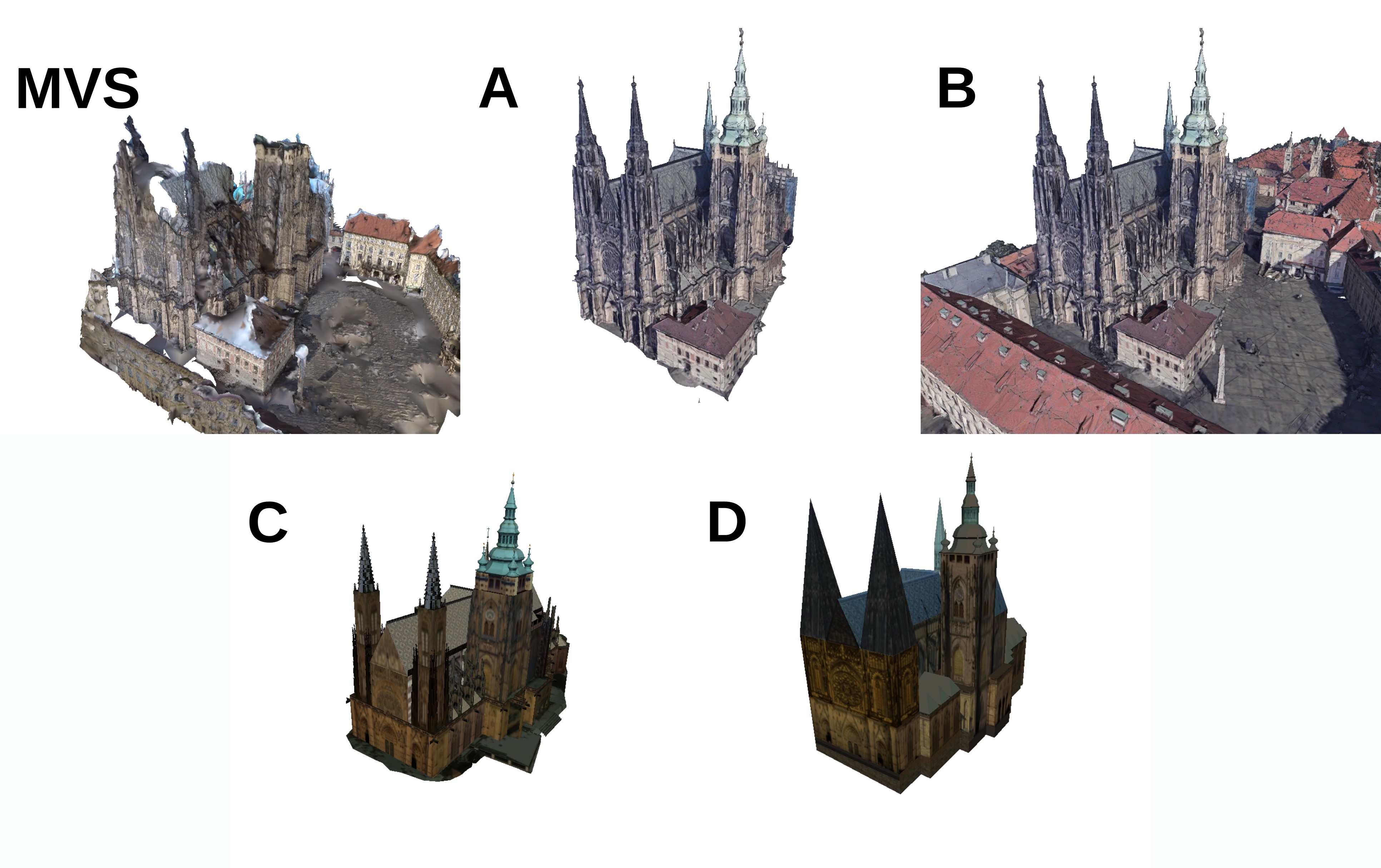}
    \caption{Enlarged St. Vitus Cathedral models from Fig.~\ref{fig:mesh_all}.}
    \label{fig:models_st_vitus}
\end{figure*}

\begin{figure*}
    \centering
    \includegraphics[width=\textwidth,trim={-2cm 0cm -2cm 0cm},clip]{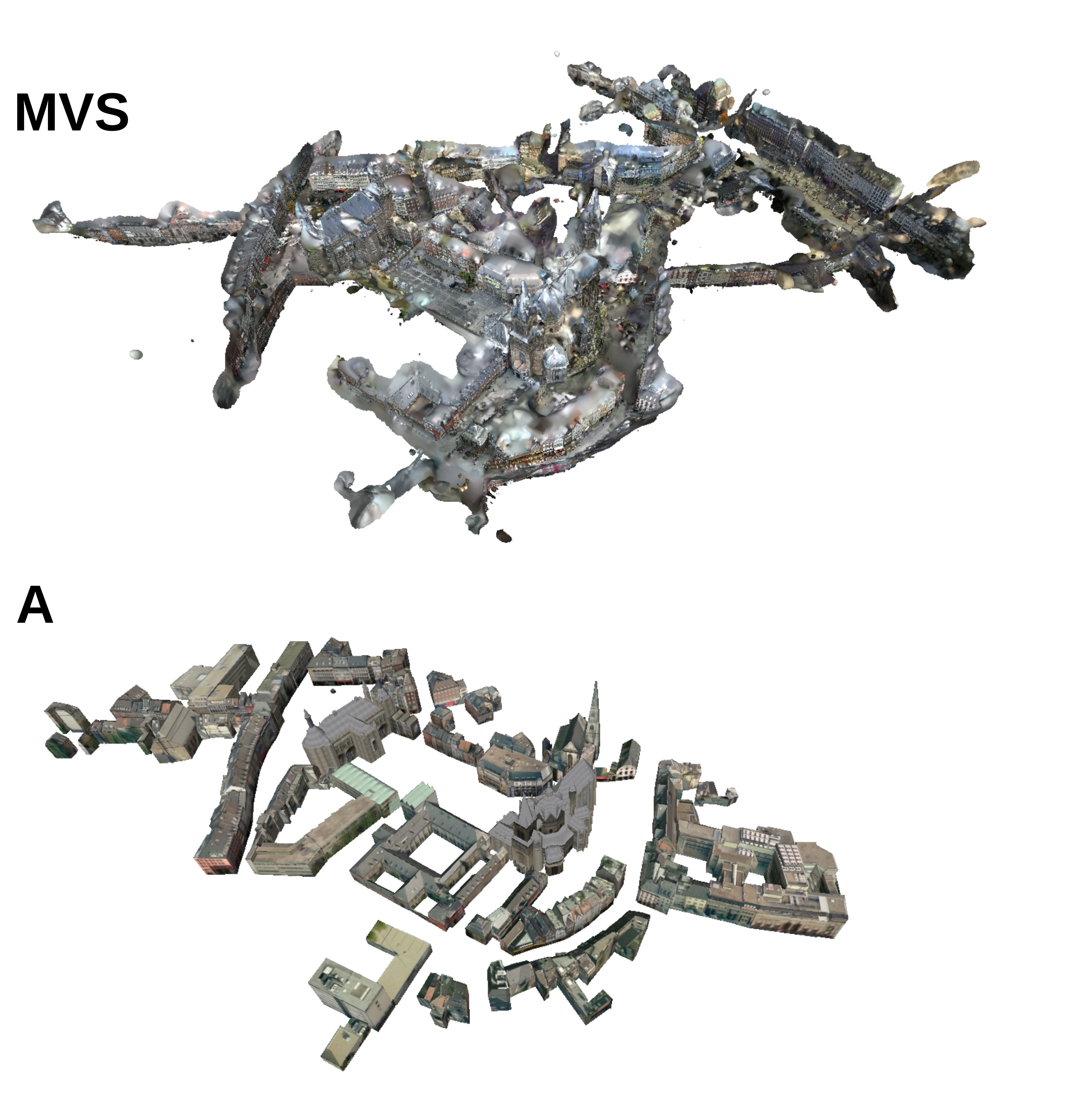}
    \caption{Enlarged Aachen models from Fig.~\ref{fig:mesh_all}.}
    \label{fig:models_aachen}
\end{figure*}

In this section, we describe the practical issues we encountered when collecting the models described in Sec.~\ref{sec:datasets}. For convenience we the individual scenes from Fig.~\ref{fig:mesh_all} from the original paper in \cref{fig:models_notre_dame} (Notre Dame), \cref{fig:models_pantheon} (Pantheon), \cref{fig:models_reichstag} (Reichstag), \cref{fig:models_st_peters_square} (St. Peter's Square), \cref{fig:models_st_vitus} (St. Vitus Cathedral) and \cref{fig:models_aachen} (Aachen).

All models downloaded from 3D Warehouse were created using the SketchUp 3D modeling software. 
The software allows mesh faces to have a material (color or texture) from both sides. Other formats, \eg, Wavefront OBJ, are not able to represent the double-sided textures, and therefore the texture of the face's back side is lost during the format conversion. The problem can be prevented during modeling time by orienting the front side of the faces outwards from the model and assigning the texture only to those. This is considered good practice in the SketchUp community; however a large fraction of the models we downloaded did not follow it. The same problem prevented us from extracting textures from Reichstag model F and St. Peters Square model D.

Few other models (Pantheon C and D) contained generic textures, which did not correspond to reality (see \cref{fig:low_fidelity_meshes}). Therefore we decided to use just the raw geometry of these models.

\section{Localization Accuracy with Internet Models and Used Metrics}
\label{sec:supp_exps}
\begin{figure*}[t!]
    \centering
    \begin{subfigure}[b]{0.404\textwidth}
        \centering
        \includegraphics[width=\textwidth,trim={0 2.9cm 8cm 0},clip]{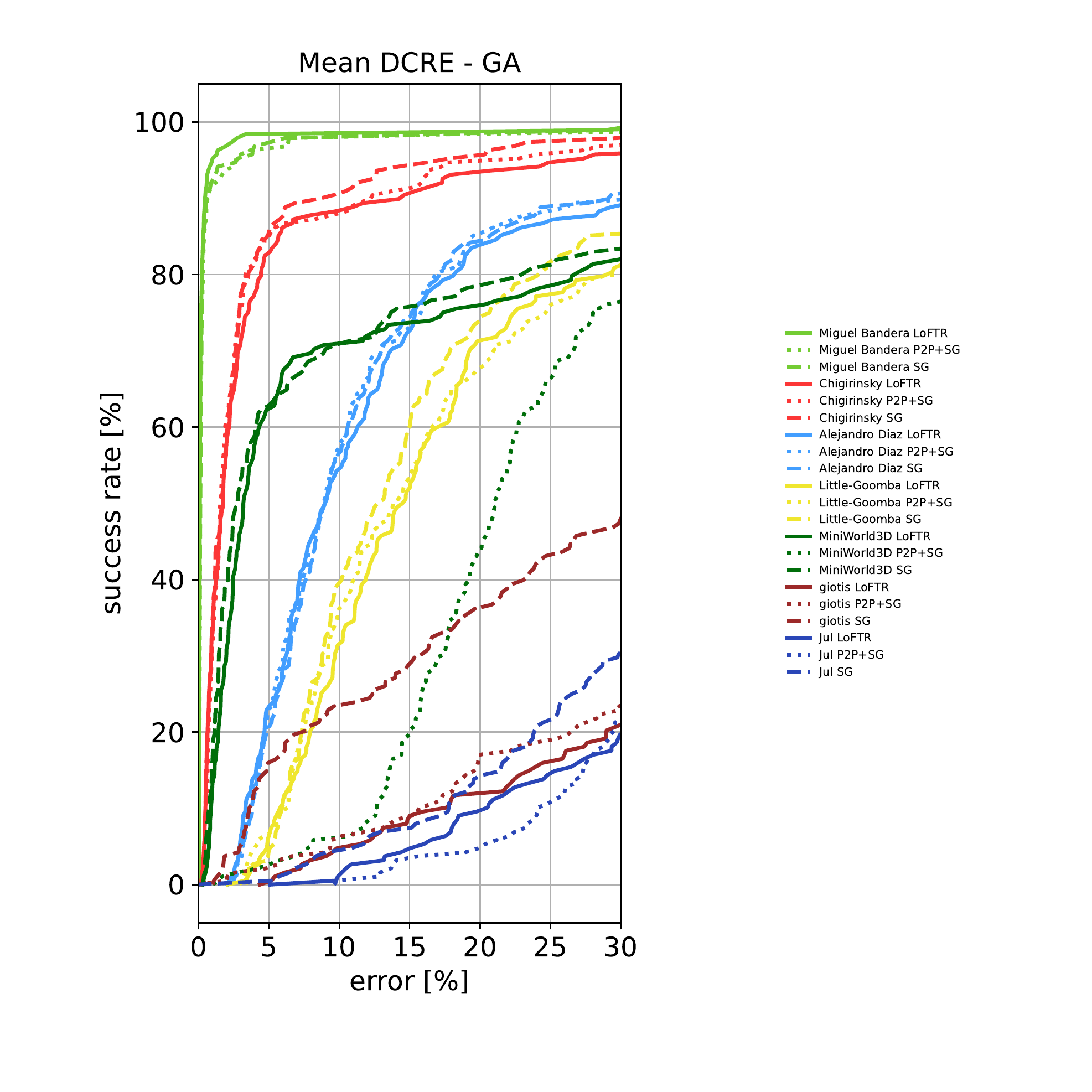}
        \caption{Notre Dame}
    \end{subfigure}
    \begin{subfigure}[b]{0.29\textwidth}
        \centering
        \includegraphics[width=\textwidth,trim={3.4cm 2.9cm 8cm 0},clip]{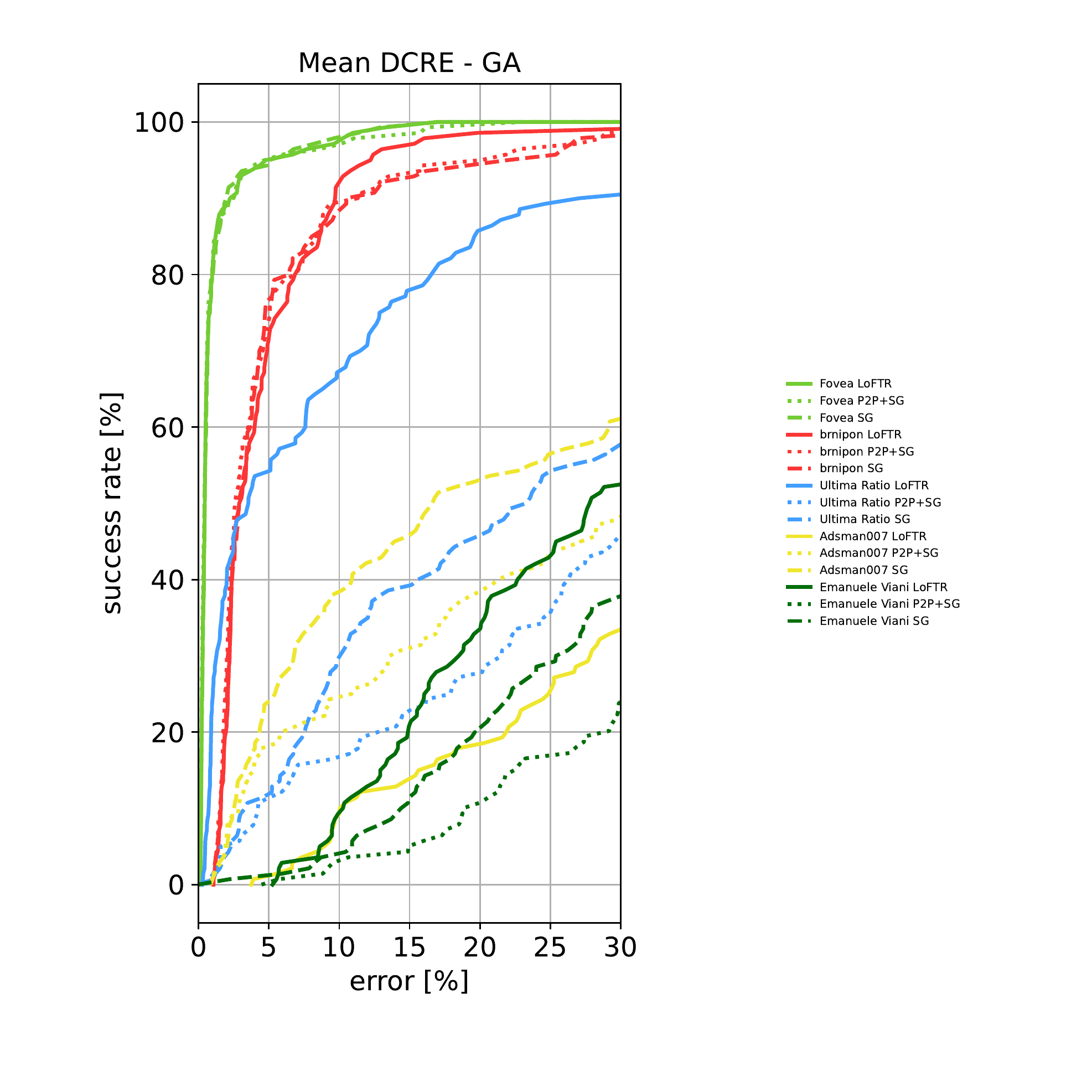}
        \caption{Pantheon}
    \end{subfigure}
    \begin{subfigure}[b]{0.29\textwidth}
        \centering
        \includegraphics[width=\textwidth,trim={-2cm -4cm -2cm 0cm},clip]{figures/example_legend_2x1.pdf}
    \end{subfigure}
    \\
    \begin{subfigure}[b]{0.404\textwidth}
        \centering
        \includegraphics[width=\textwidth,trim={0 1.7cm 8cm 1.4cm},clip]{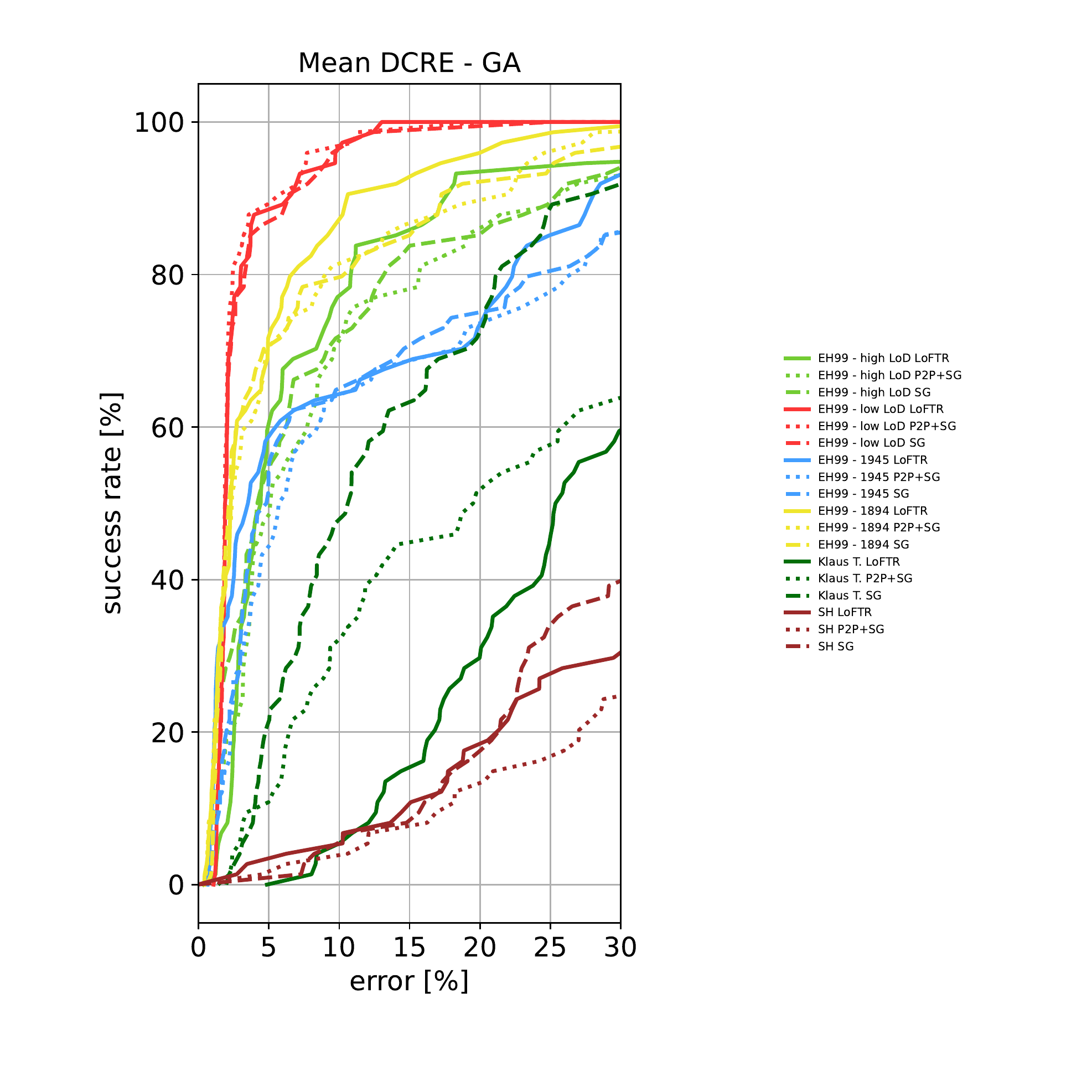}
        \caption{Reichstag}
    \end{subfigure}
    \begin{subfigure}[b]{0.29\textwidth}
        \centering
        \includegraphics[width=\textwidth,trim={3.4cm 1.7cm 8cm 1.4cm},clip]{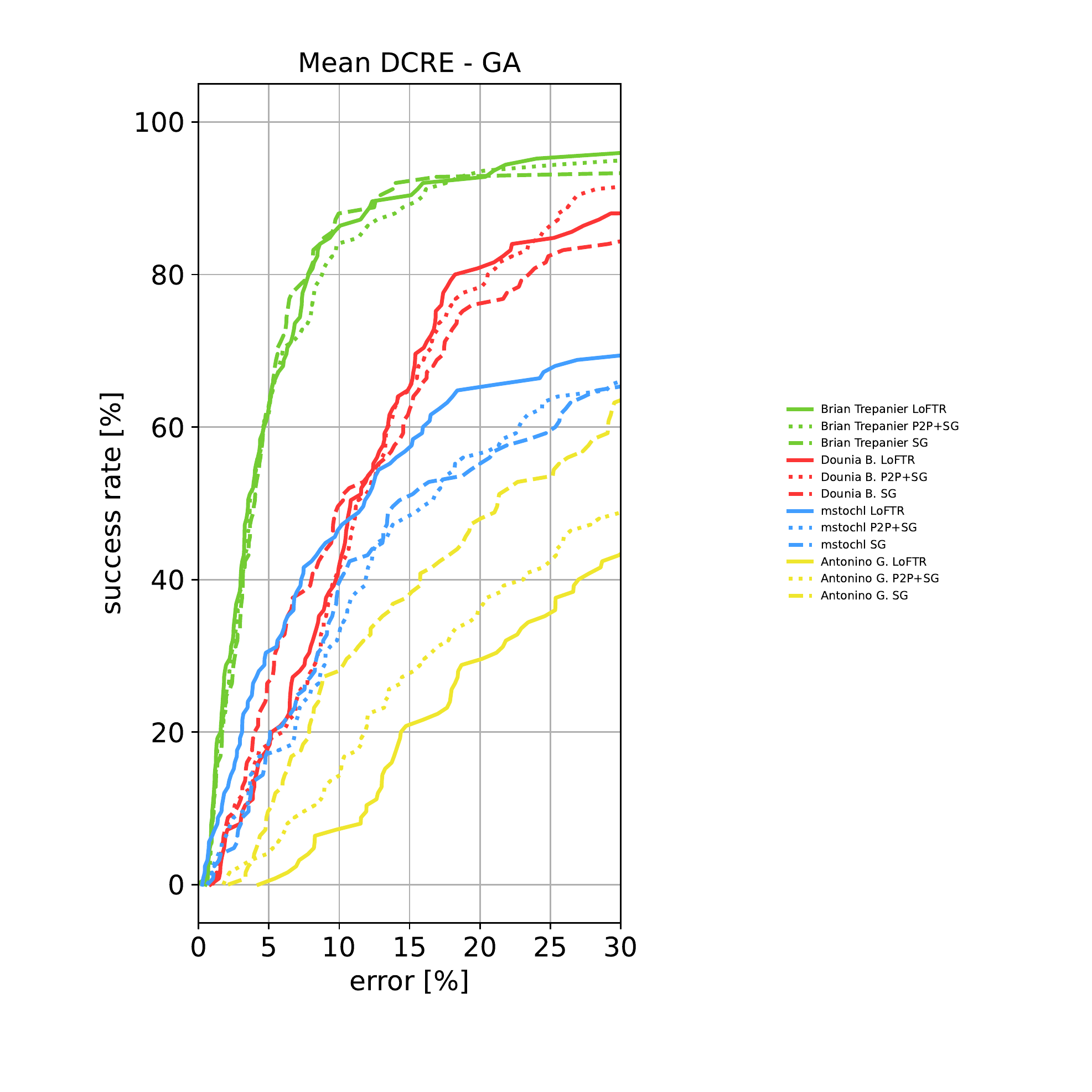}
        \caption{St. Peter's square}
    \end{subfigure}
    \begin{subfigure}[b]{0.29\textwidth}
        \centering
        \includegraphics[width=\textwidth,trim={3.4cm 1.7cm 8cm 0.8cm},clip]{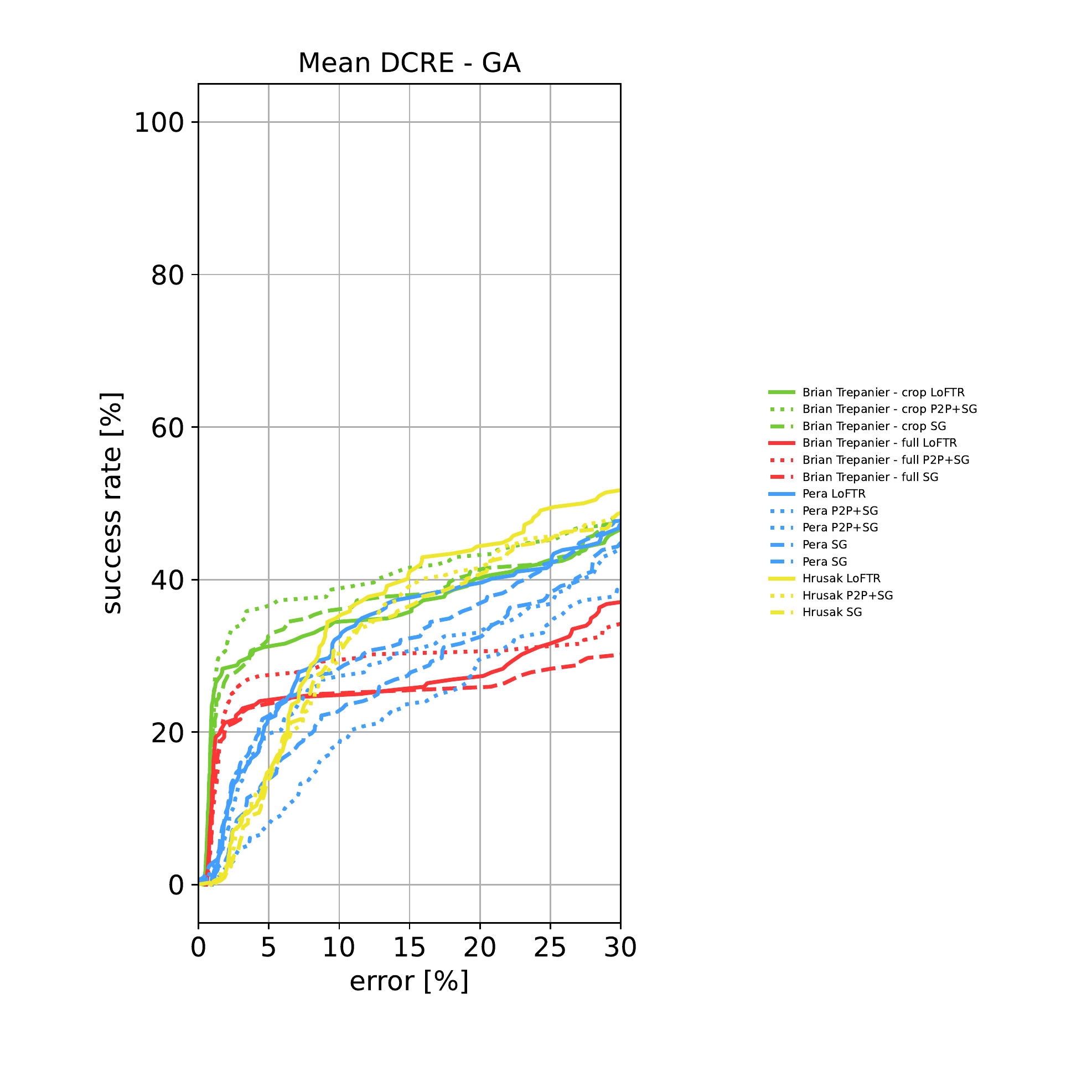}
        \caption{St. Vitus}
    \end{subfigure}
    \caption{Reproduction of the first row of Fig.~\ref{fig:eval_dcre}. Cumulative histograms of the mean DCRE over all query images in a scene for the ground truth poses obtained via global alignment (GA). We show the DCRE as the percentage of the image diagonal.}
    \label{fig:eval_dcre_mean_ga}
\end{figure*}

\begin{figure*}[t!]
    \centering
    \begin{subfigure}[b]{0.404\textwidth}
        \centering
        \includegraphics[width=\textwidth,trim={0 2.9cm 8cm 0},clip]{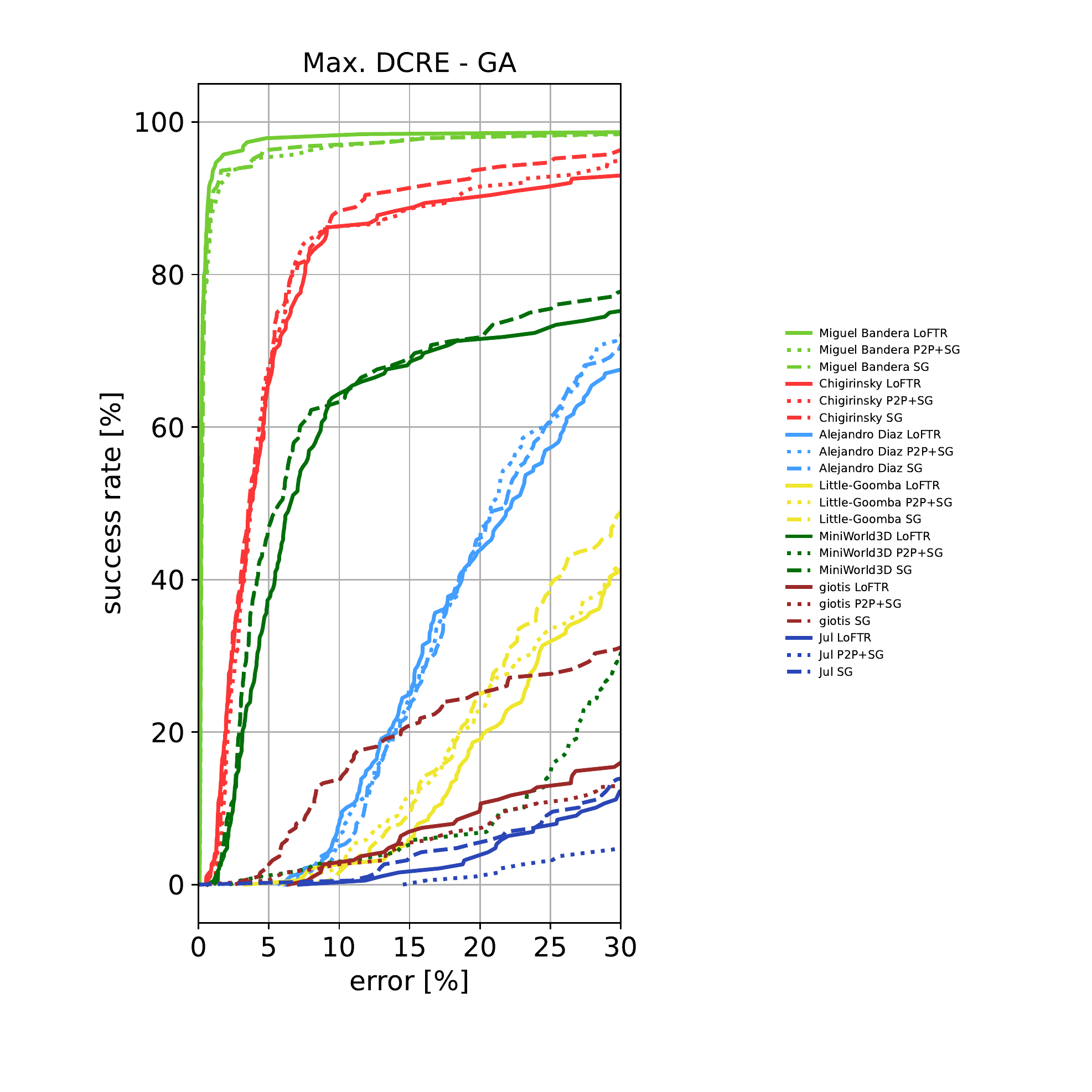}
        \caption{Notre Dame}
    \end{subfigure}
    \begin{subfigure}[b]{0.29\textwidth}
        \centering
        \includegraphics[width=\textwidth,trim={3.4cm 2.9cm 8cm 0},clip]{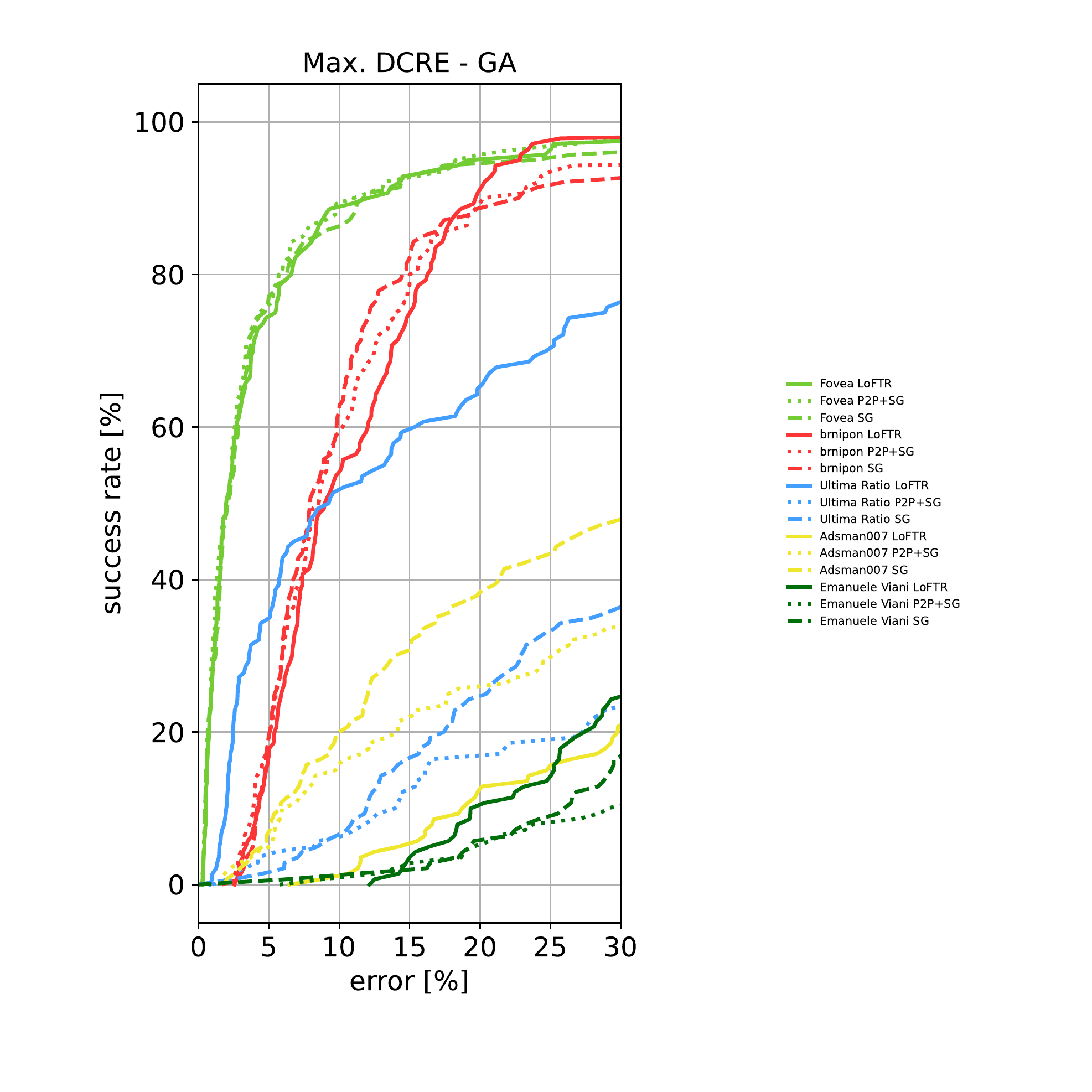}
        \caption{Pantheon}
    \end{subfigure}
    \begin{subfigure}[b]{0.29\textwidth}
        \centering
        \includegraphics[width=\textwidth,trim={-2cm -4cm -2cm 0cm},clip]{figures/example_legend_2x1.pdf}
    \end{subfigure}
    \\
    \begin{subfigure}[b]{0.404\textwidth}
        \centering
        \includegraphics[width=\textwidth,trim={0 1.7cm 8cm 1.4cm},clip]{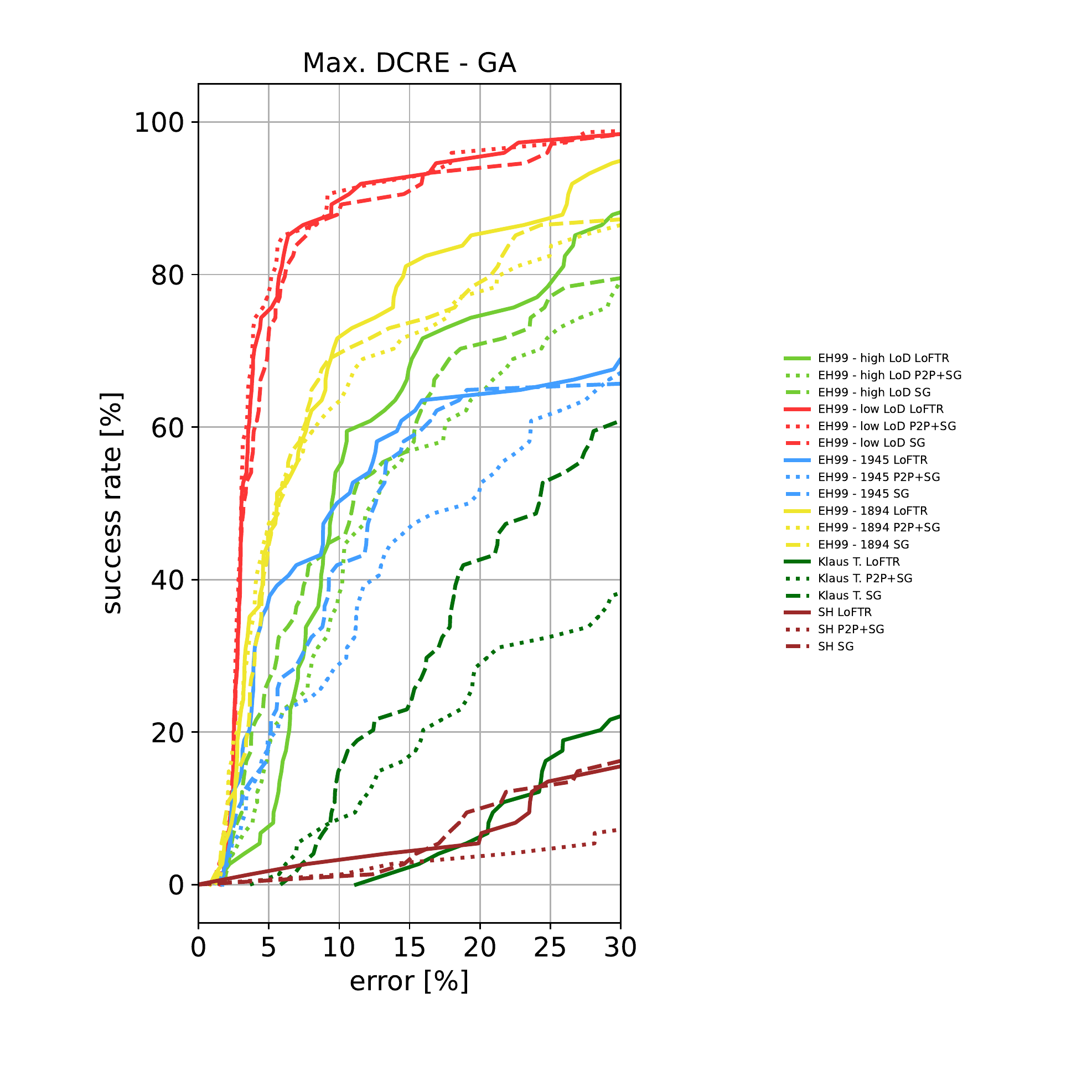}
        \caption{Reichstag}
    \end{subfigure}
    \begin{subfigure}[b]{0.29\textwidth}
        \centering
        \includegraphics[width=\textwidth,trim={3.4cm 1.7cm 8cm 1.4cm},clip]{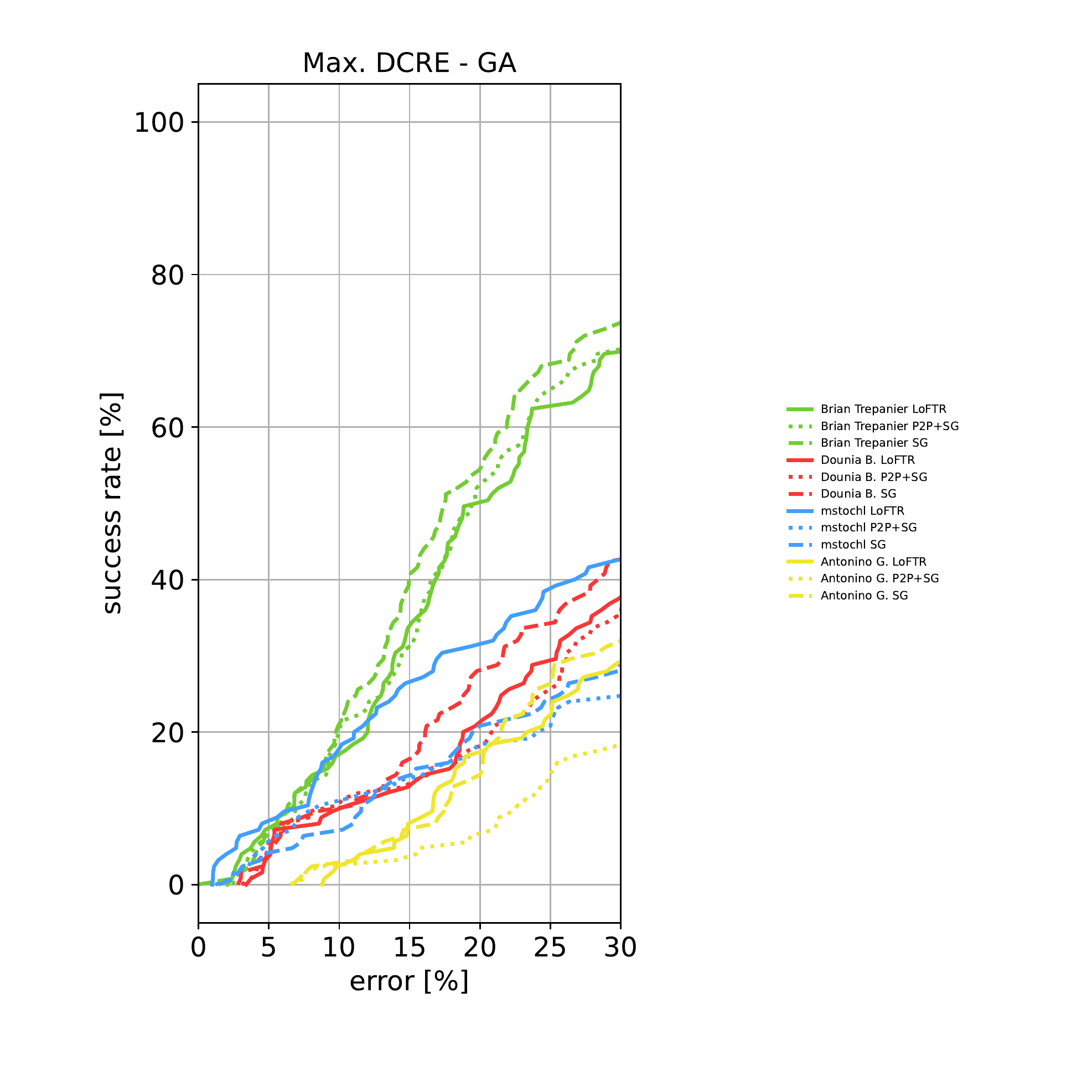}
        \caption{St. Peter's square}
    \end{subfigure}
    \begin{subfigure}[b]{0.29\textwidth}
        \centering
        \includegraphics[width=\textwidth,trim={3.4cm 1.7cm 8cm 0.8cm},clip]{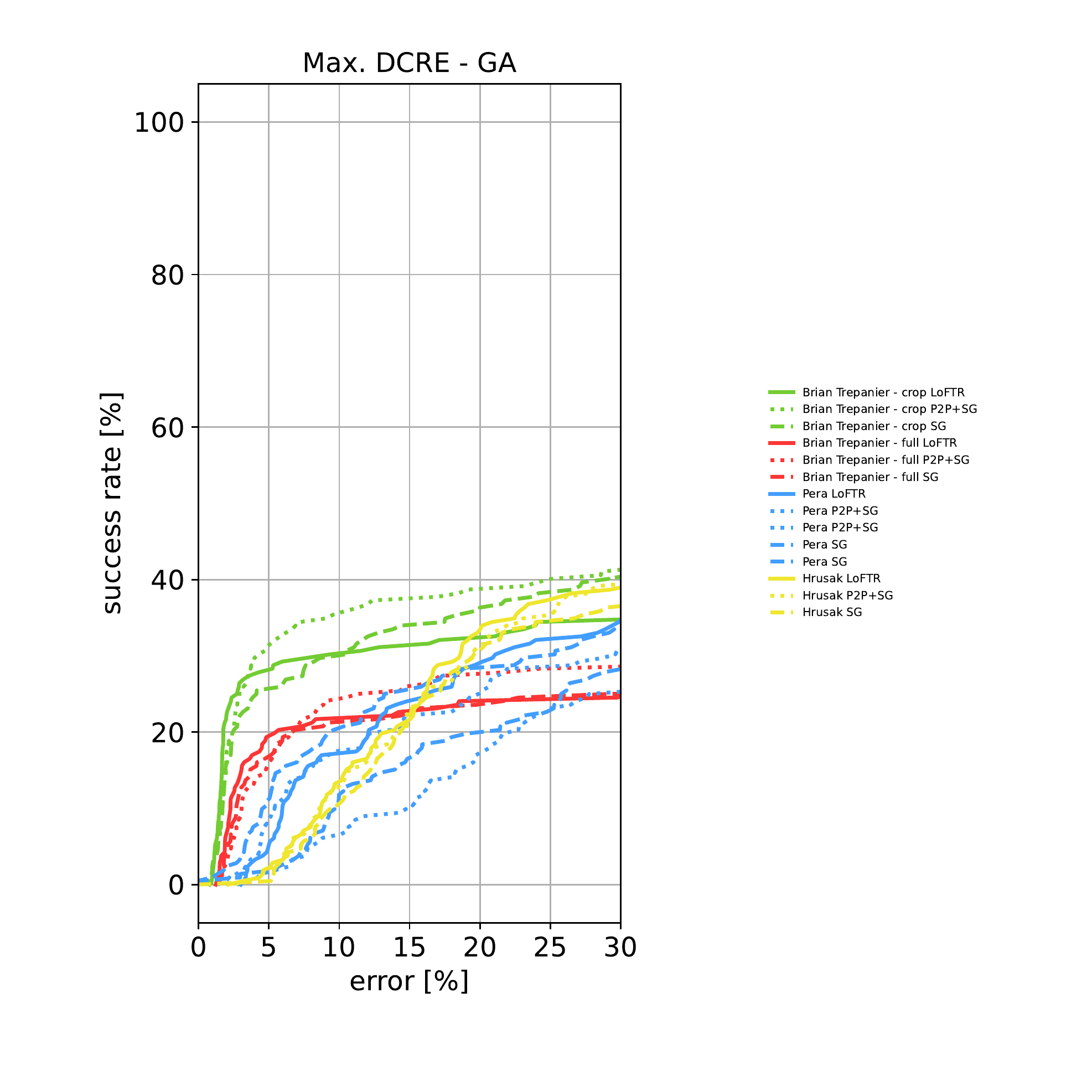}
        \caption{St. Vitus}
    \end{subfigure}
    \caption{Cumulative histograms of the maximum DCRE over all query images in a scene for the ground truth poses obtained via global alignment (GA). We show the DCRE as the percentage of the image diagonal.}
    \label{fig:eval_dcre_max_ga}
\end{figure*}

\begin{figure*}[t!]
    \centering
    \begin{subfigure}[b]{0.404\textwidth}
        \centering
        \includegraphics[width=\textwidth,trim={0 2.9cm 8cm 0},clip]{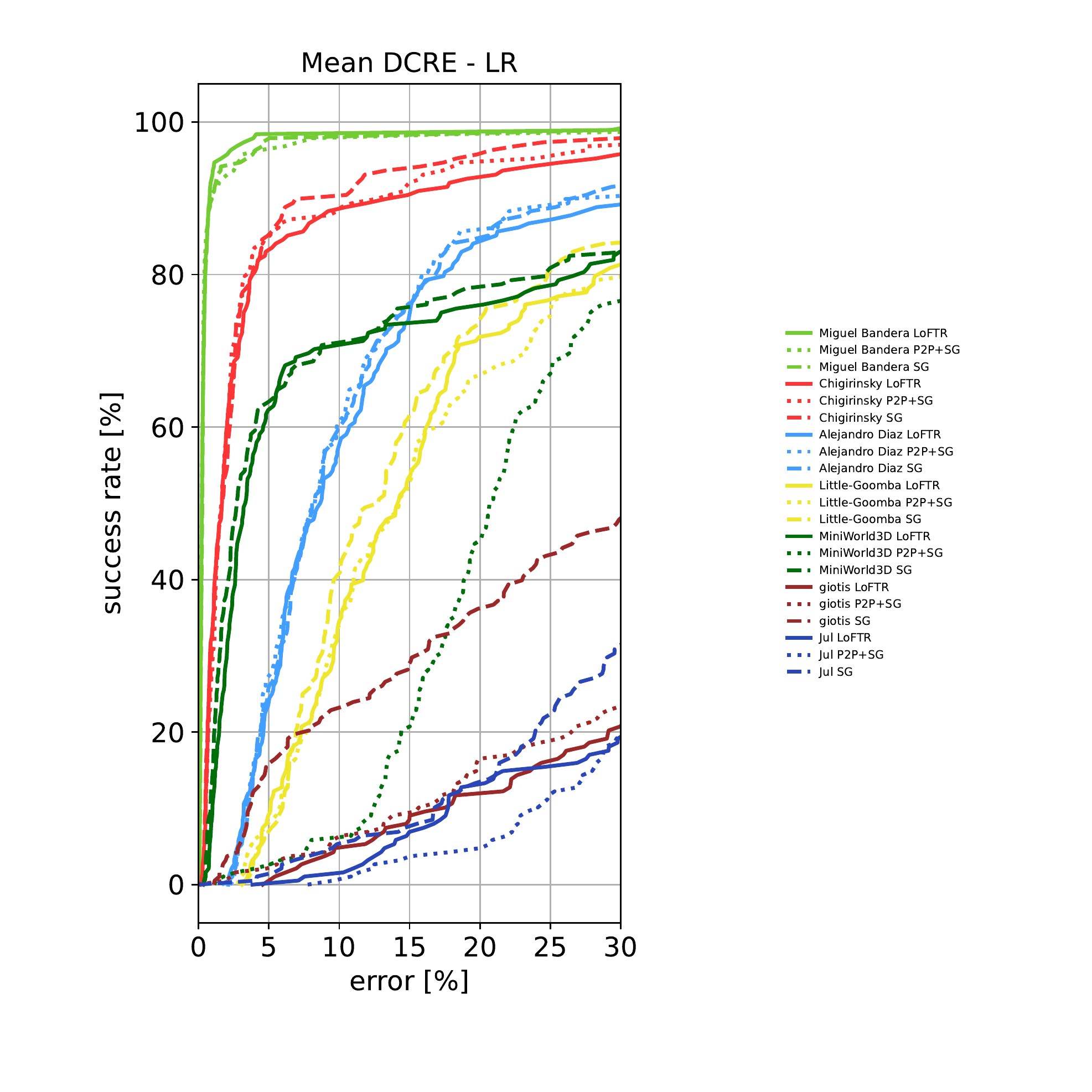}
        \caption{Notre Dame}
    \end{subfigure}
    \begin{subfigure}[b]{0.29\textwidth}
        \centering
        \includegraphics[width=\textwidth,trim={3.4cm 2.9cm 8cm 0},clip]{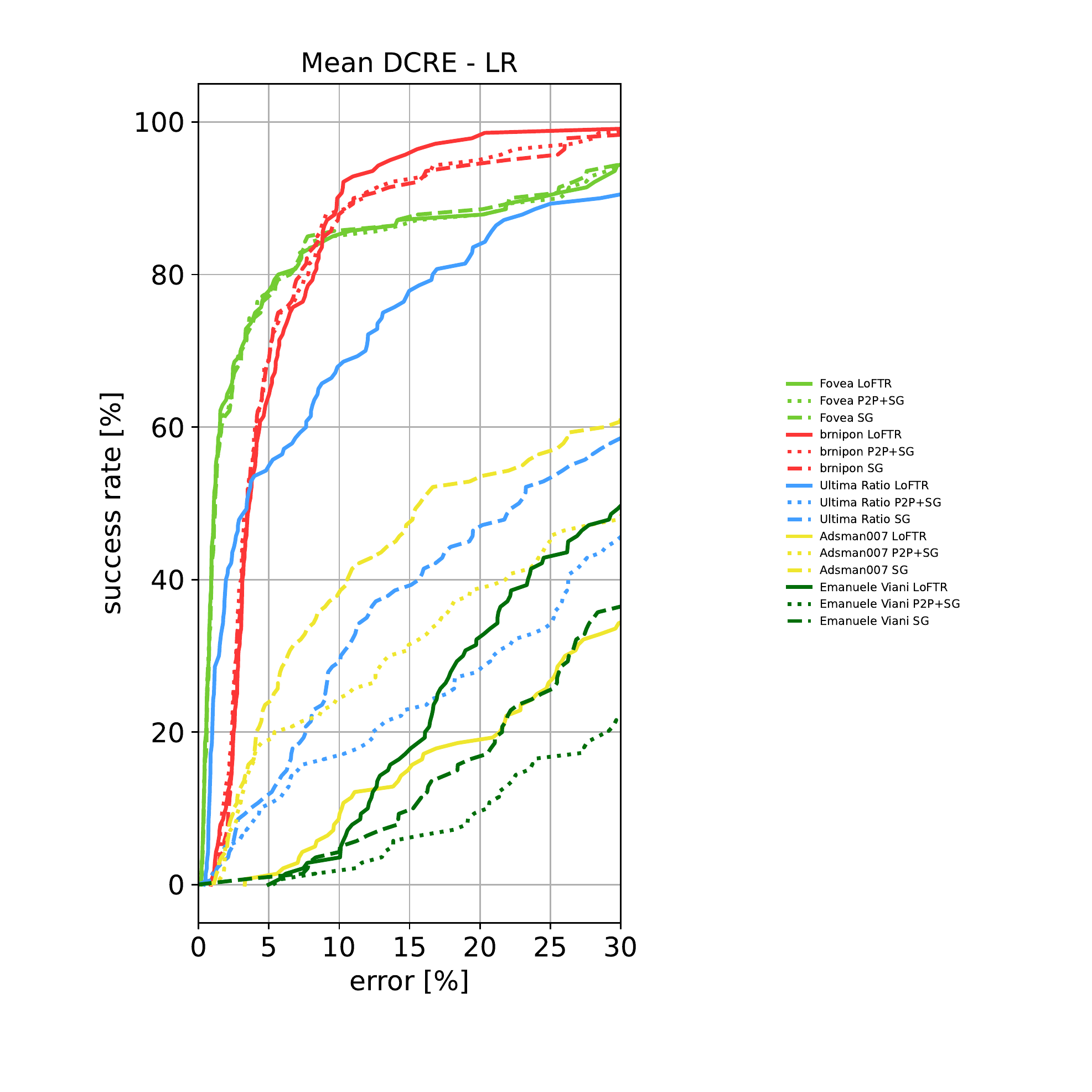}
        \caption{Pantheon}
    \end{subfigure}
    \begin{subfigure}[b]{0.29\textwidth}
        \centering
        \includegraphics[width=\textwidth,trim={-2cm -4cm -2cm 0cm},clip]{figures/example_legend_2x1.pdf}
    \end{subfigure}
    \\
    \begin{subfigure}[b]{0.404\textwidth}
        \centering
        \includegraphics[width=\textwidth,trim={0 1.7cm 8cm 1.4cm},clip]{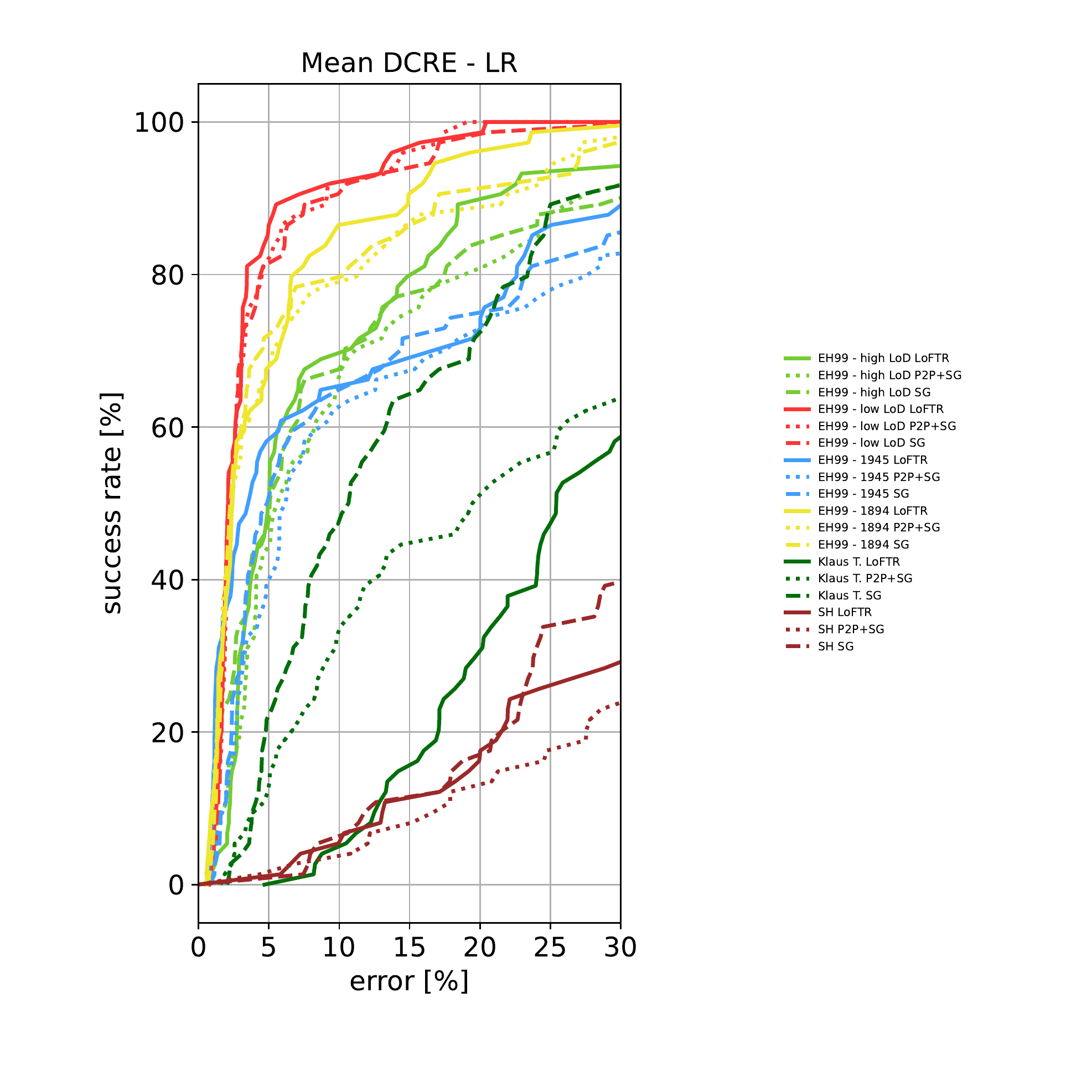}
        \caption{Reichstag}
    \end{subfigure}
    \begin{subfigure}[b]{0.29\textwidth}
        \centering
        \includegraphics[width=\textwidth,trim={3.4cm 1.7cm 8cm 1.4cm},clip]{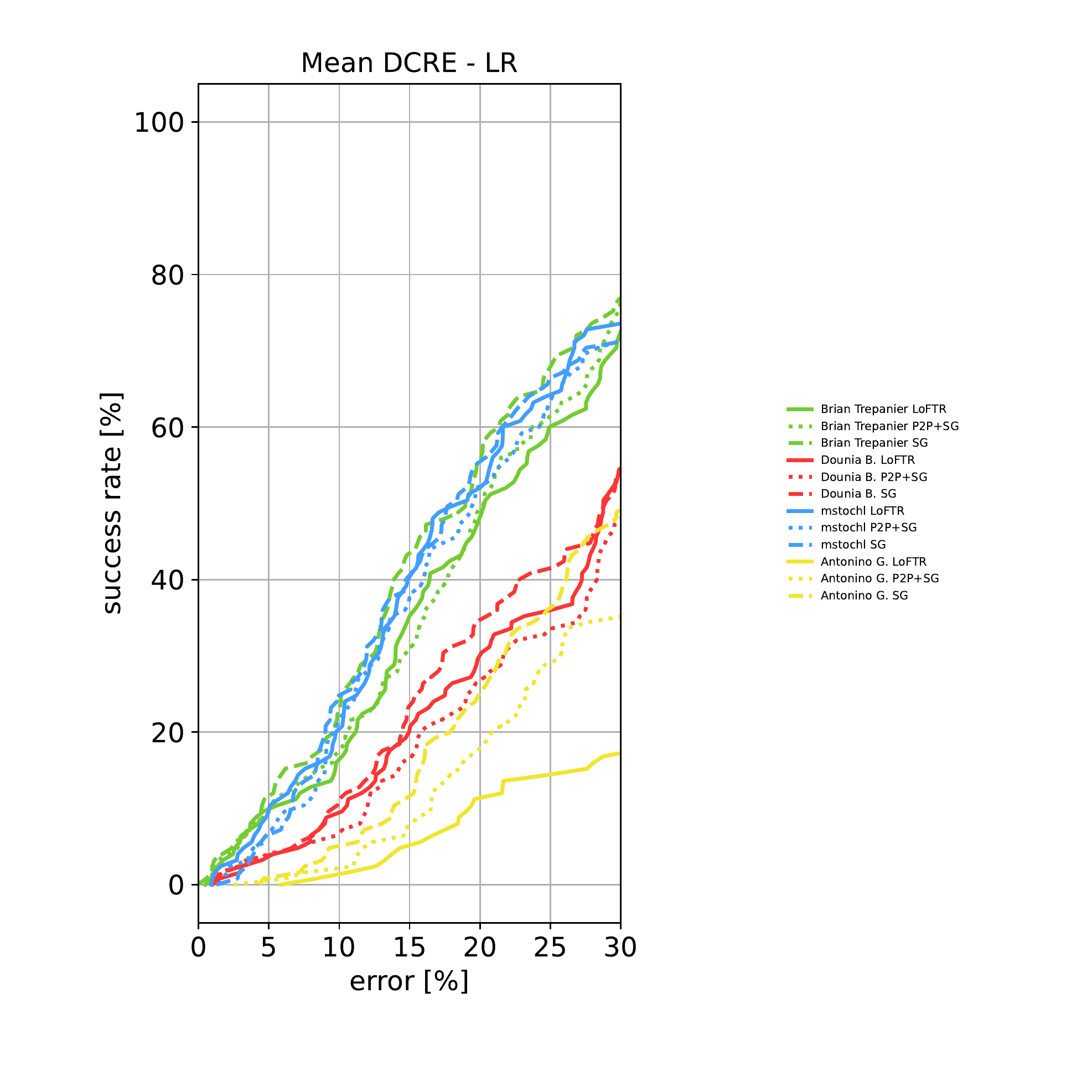}
        \caption{St. Peter's square}
    \end{subfigure}
    \begin{subfigure}[b]{0.29\textwidth}
        \centering
        \includegraphics[width=\textwidth,trim={3.4cm 1.7cm 8cm 0.8cm},clip]{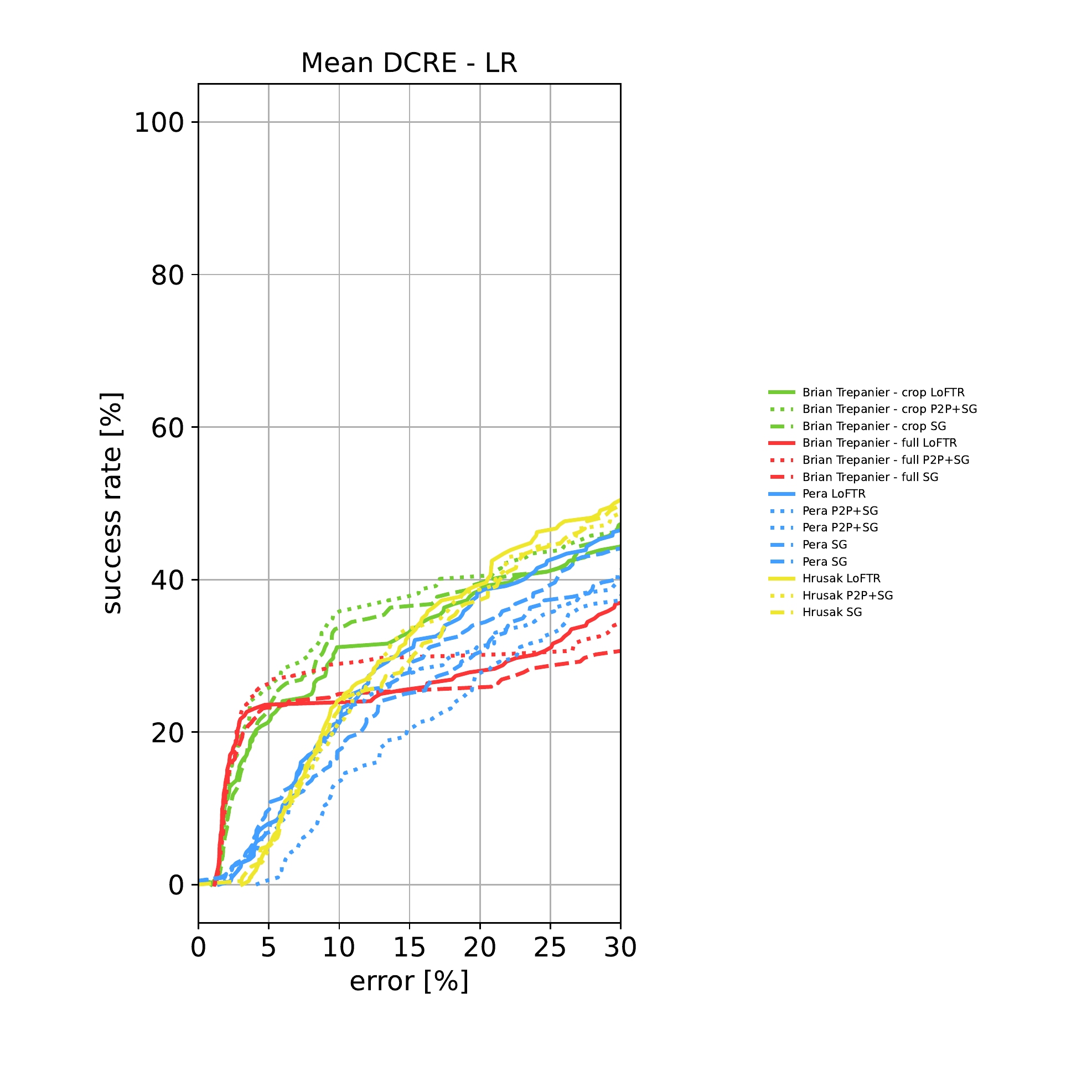}
        \caption{St. Vitus}
    \end{subfigure}
    \caption{Reproduction of the second row of Fig.~\ref{fig:eval_dcre}. Cumulative histograms of the mean DCRE over all query images in a scene for the ground truth poses obtained via local refinement (LR). We show the DCRE as the percentage of the image diagonal.}
    \label{fig:eval_dcre_mean_lr}
\end{figure*}

\begin{figure*}[t!]
    \centering
    \begin{subfigure}[b]{0.404\textwidth}
        \centering
        \includegraphics[width=\textwidth,trim={0 2.9cm 8cm 0},clip]{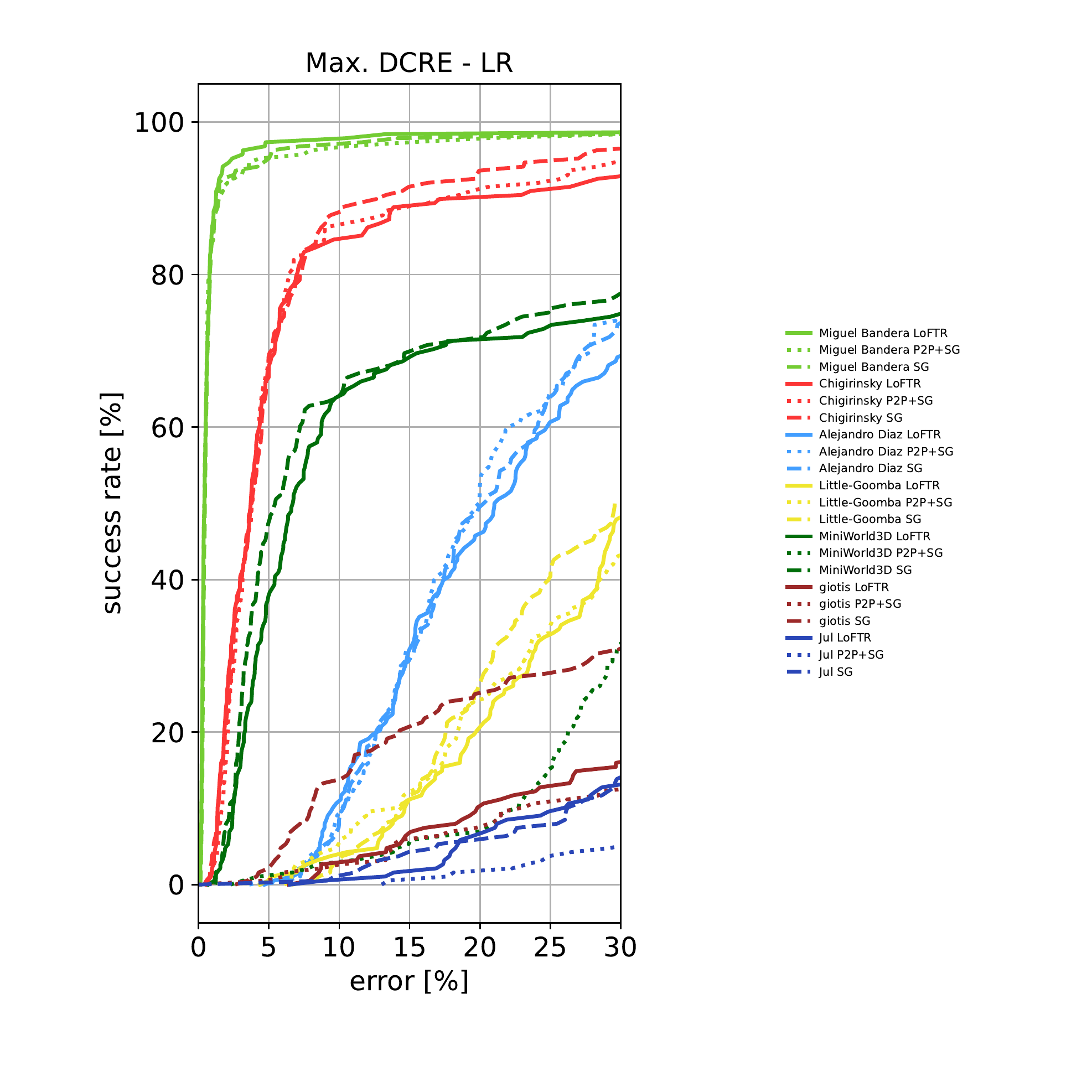}
        \caption{Notre Dame}
    \end{subfigure}
    \begin{subfigure}[b]{0.29\textwidth}
        \centering
        \includegraphics[width=\textwidth,trim={3.4cm 2.9cm 8cm 0},clip]{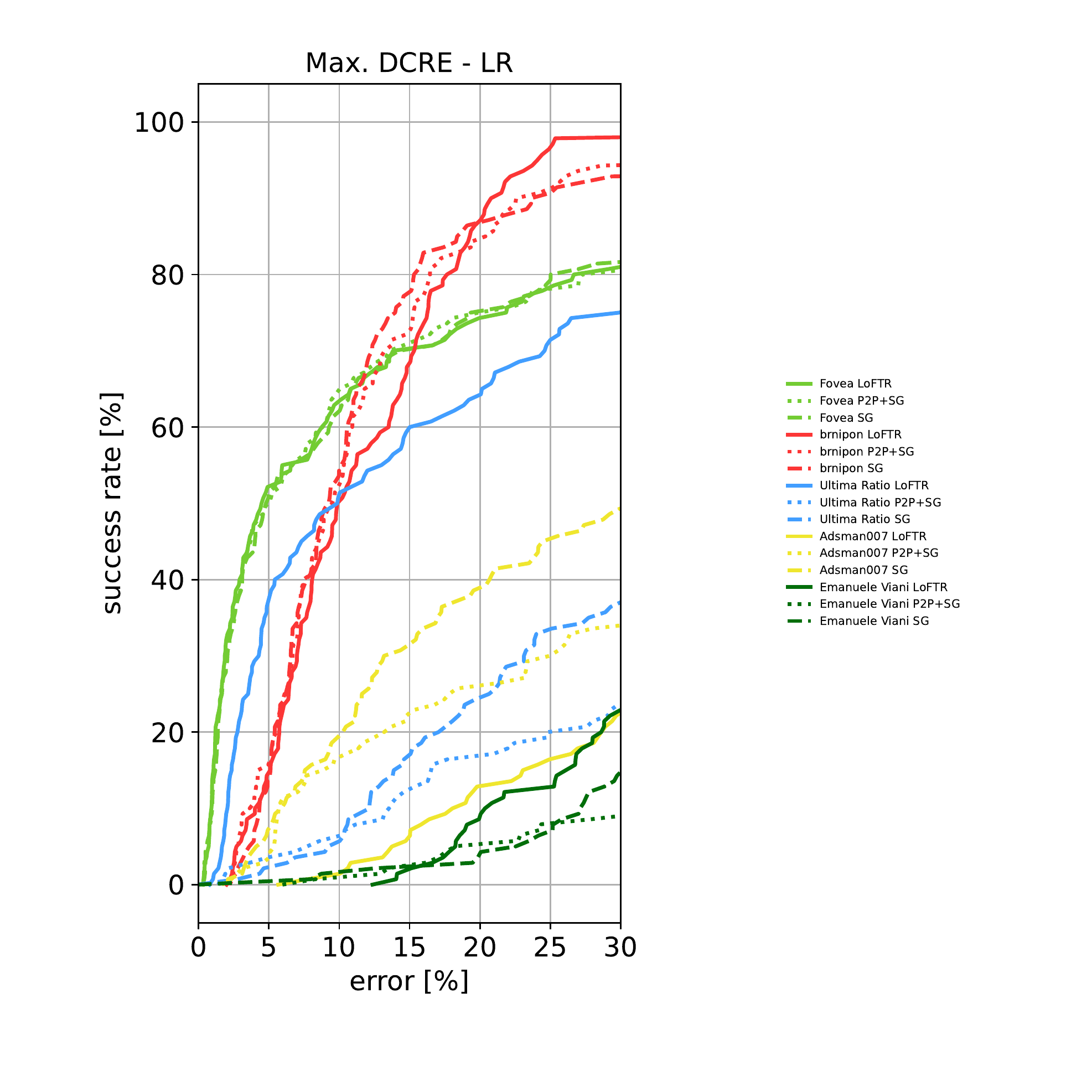}
        \caption{Pantheon}
    \end{subfigure}
    \begin{subfigure}[b]{0.29\textwidth}
        \centering
        \includegraphics[width=\textwidth,trim={-2cm -4cm -2cm 0cm},clip]{figures/example_legend_2x1.pdf}
    \end{subfigure}
    \\
    \begin{subfigure}[b]{0.404\textwidth}
        \centering
        \includegraphics[width=\textwidth,trim={0 1.7cm 8cm 1.4cm},clip]{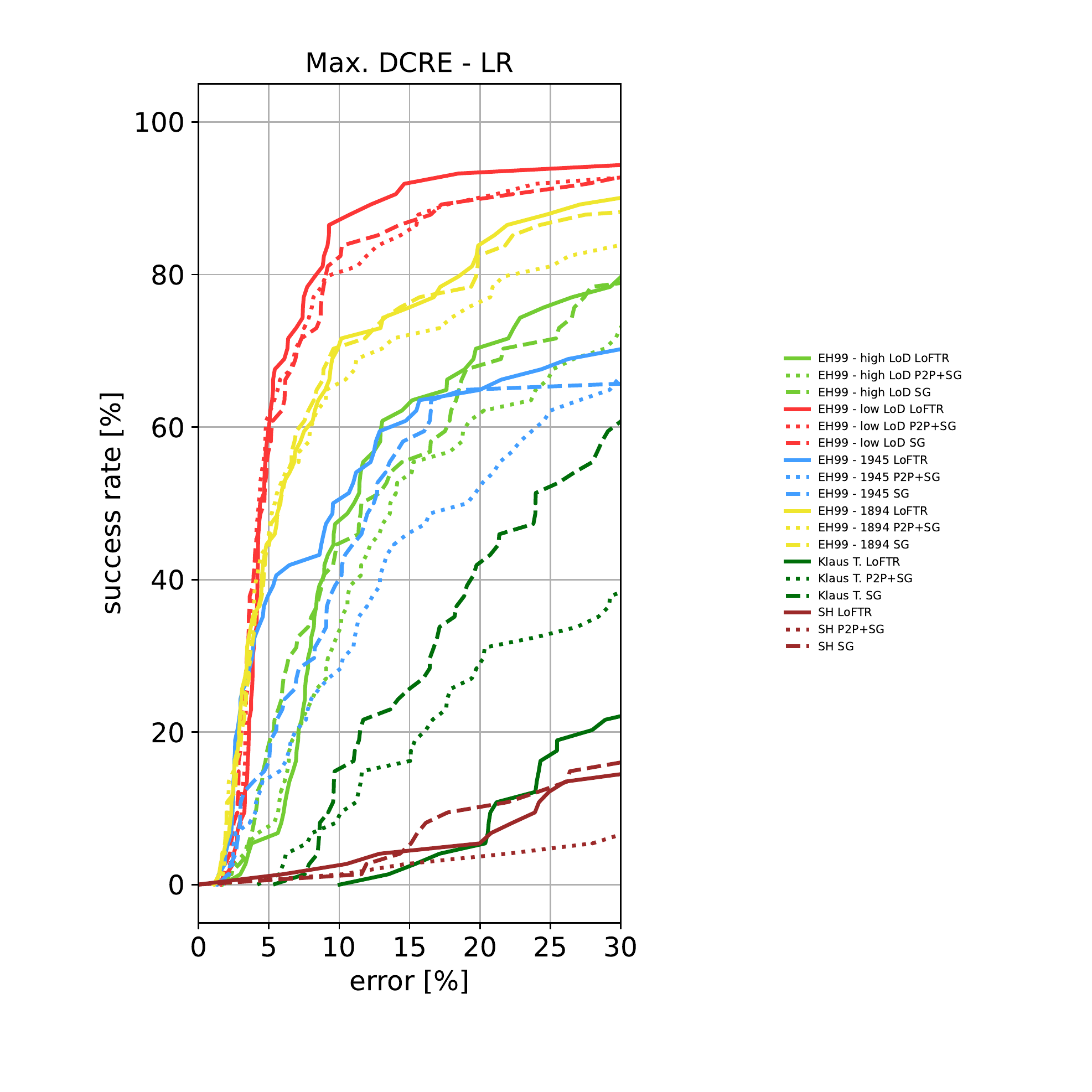}
        \caption{Reichstag}
    \end{subfigure}
    \begin{subfigure}[b]{0.29\textwidth}
        \centering
        \includegraphics[width=\textwidth,trim={3.4cm 1.7cm 8cm 1.4cm},clip]{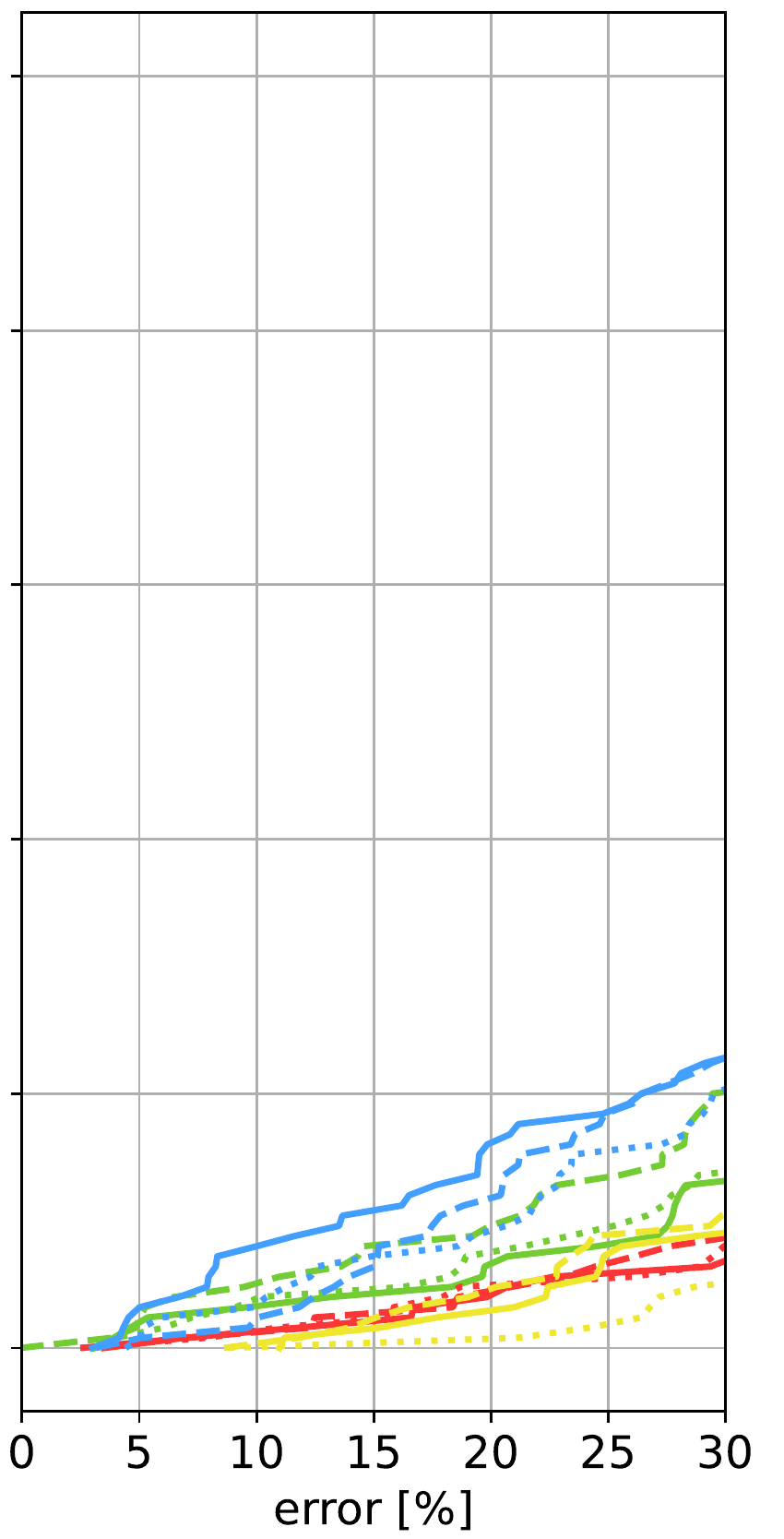}
        \caption{St. Peter's square}
    \end{subfigure}
    \begin{subfigure}[b]{0.29\textwidth}
        \centering
        \includegraphics[width=\textwidth,trim={3.4cm 1.7cm 8cm 0.8cm},clip]{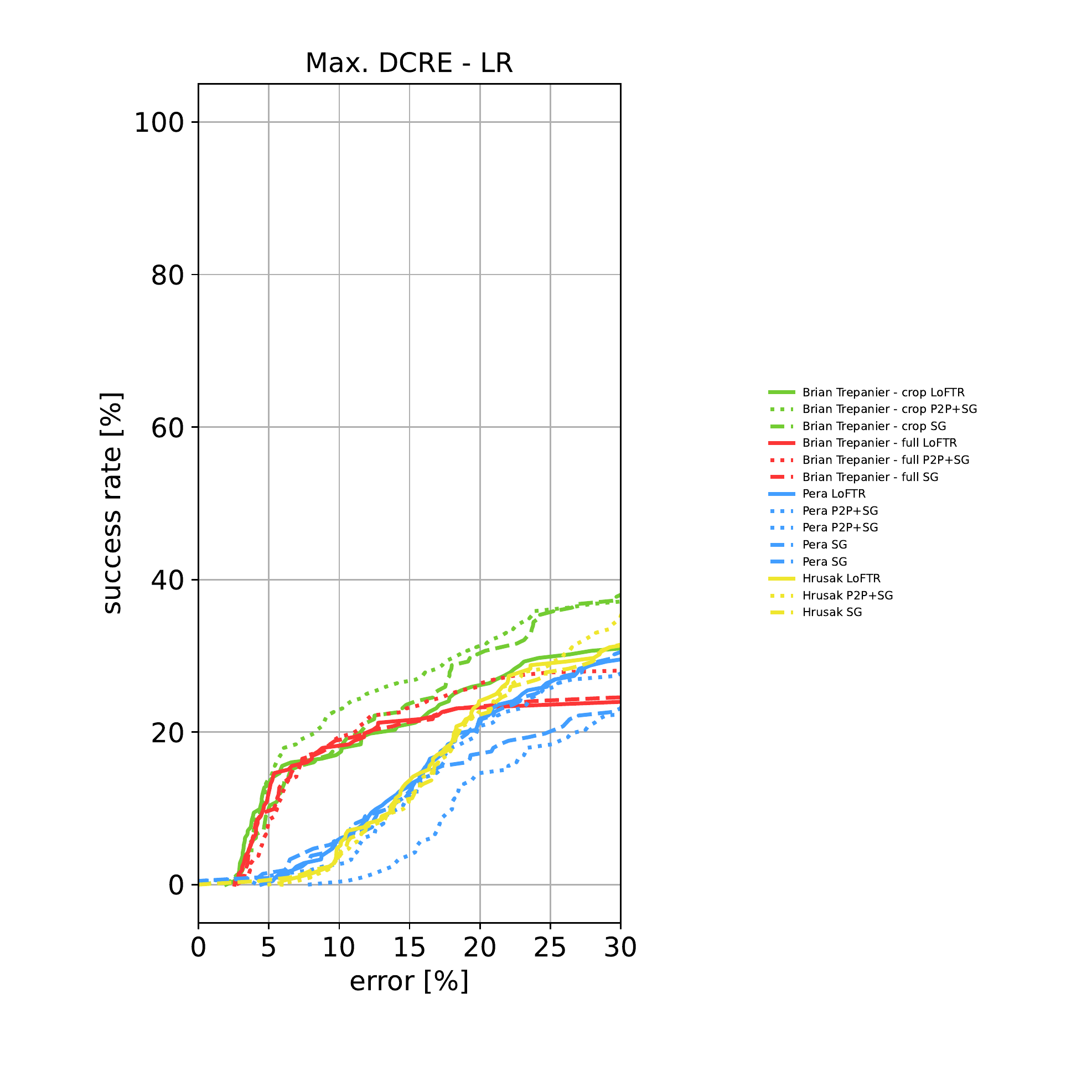}
        \caption{St. Vitus}
    \end{subfigure}
    \caption{Cumulative histograms of the maximum DCRE over all query images in a scene for the ground truth poses obtained via local refinement (LR). We show the DCRE as the percentage of the image diagonal.}
    \label{fig:eval_dcre_max_lr}
\end{figure*}

Sec.~\ref{sec:experiments} focused on showing results for the mean Dense Correspondence Re-Projection Error (mean DCRE) for both the global alignment (GA) and local refinement (LR) versions of the ground truth. 
For maximum DCRE results, Sec.~\ref{sec:experiments} pointed to the supplementary material. 
These results will be presented in this section. 
For convenience and to facilitate easier comparisons, the following shows results obtained using the MeshLoc pipeline~\cite{Panek2022ECCV} for both possible DCRE aggregation functions (mean and maximum) and both ground truth methods (global alignment (GA) and local refinement (LR)) (Sec.~\ref{sec:eval_protocol}).
We replicate the first row of Fig.~\ref{fig:eval_dcre}, which shows the experiments with mean DCRE aggregation and global alignment (GA) ground truth in \cref{fig:eval_dcre_mean_ga}, and the second row of Fig.~\ref{fig:eval_dcre}, which shows mean DCRE and local refinement (LR), in \cref{fig:eval_dcre_mean_lr}.
\cref{fig:eval_dcre_max_ga} shows max DCRE and global alignment (GA) results. 
\cref{fig:eval_dcre_max_lr} shows max DCRE and local refinement (LR) results.

Regarding the difference between mean DCRE and maximum DCRE curves, we can observe two types of behavior. For the first type, the maximum DCRE does not alter much from the mean DCRE curve (\eg, Notre Dame A, B, and E). For the second type, the drop is much more significant (\eg, Notre Dame C, D). The first type corresponds to the models with more accurate geometry (see Fig.~\ref{fig:three graphs}). 

Naturally, measuring the maximum instead of the mean DCRE per image leads to lower performance. Still, the relative ranking between different models is mostly preserved, especially for more accurate poses with a mean / max. DCRE of 15\% of the image diagonal or smaller.

\section{Isolating the Impact of Geometric Fidelity}
\label{sec:geom_fidel}
To isolate the impact of geometric fidelity on localization performance, we experimented with changing the geometry used in the MeshLoc pipeline while fixing the appearance. 
To this end, we matched each query image against other real images (see Sec.~\ref{sec:experiments}). 
Both image retrieval with AP-GeM~\cite{gordo2017end, revaud2019learning} descriptors and local feature matching is done using a database of real images.
Renderings of the different 3D models are only used to obtain the depth maps used by MeshLoc~\cite{Panek2022ECCV} to establish 2D-3D matches.

All IMC (Image Matching Challenge) 2021~\cite{Thomee2016YFCC100MTN,Heinly2015ReconstructingTW,Jin2020} scenes, except Reichstag, contain a large number of images, from which we use only a small part as queries (\cf Tab.~\ref{tab:datasets} for the resulting sizes of the query subsets). The rest of the images were used as an image database in this experiment. For Notre Dame and St. Peters Square, we use every 20th image as a query, for Pantheon every 10th, and for St. Vitus Cathedral every 4th. The Reichstag scene contains only 75 images; therefore, we decided to use all of them as queries and performed the experiment in a leave-one-out manner, \ie, when localizing one of the images, we used all the other queries as the image database.

\begin{figure*}[t!]
    \centering
    \begin{subfigure}[b]{0.404\textwidth}
        \centering
        \includegraphics[width=\textwidth,trim={0 2.9cm 8cm 0},clip]{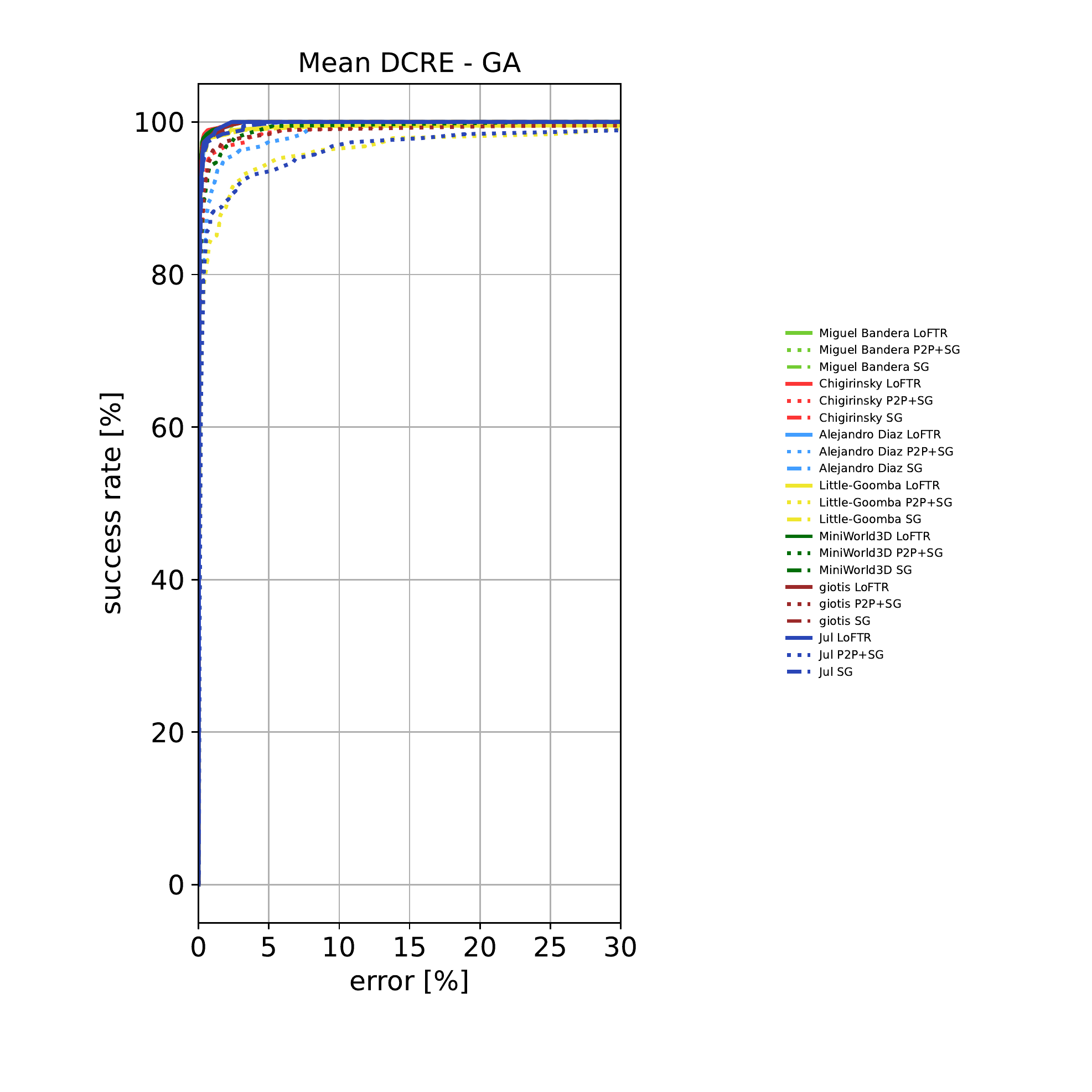}
        \caption{Notre Dame}
    \end{subfigure}
    \begin{subfigure}[b]{0.29\textwidth}
        \centering
        \includegraphics[width=\textwidth,trim={3.4cm 2.9cm 8cm 0},clip]{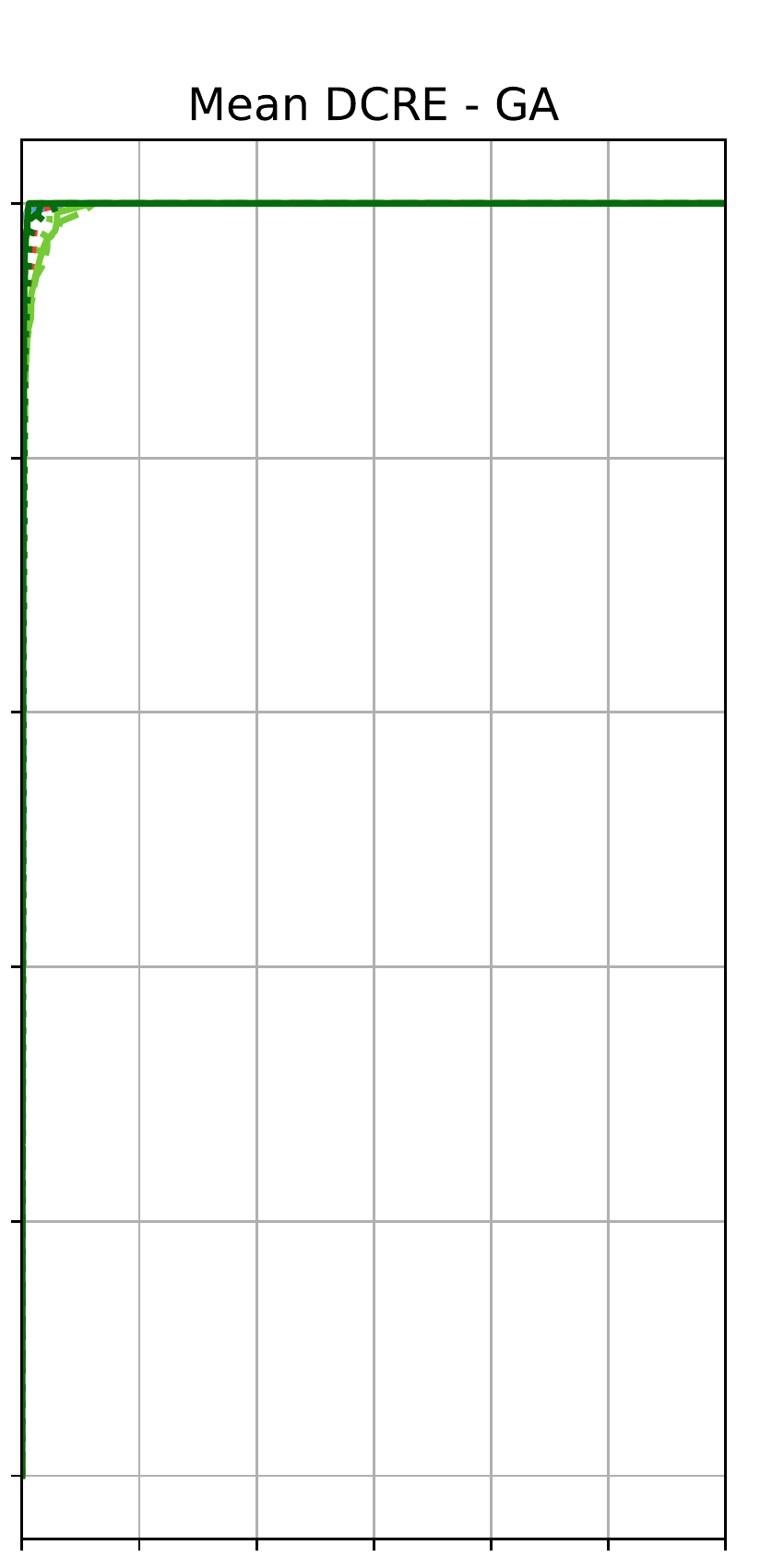}
        \caption{Pantheon}
    \end{subfigure}
    \begin{subfigure}[b]{0.29\textwidth}
        \centering
        \includegraphics[width=\textwidth,trim={-2cm -4cm -2cm 0cm},clip]{figures/example_legend_2x1.pdf}
    \end{subfigure}
    \\
    \begin{subfigure}[b]{0.404\textwidth}
        \centering
        \includegraphics[width=\textwidth,trim={0 1.7cm 8cm 1.4cm},clip]{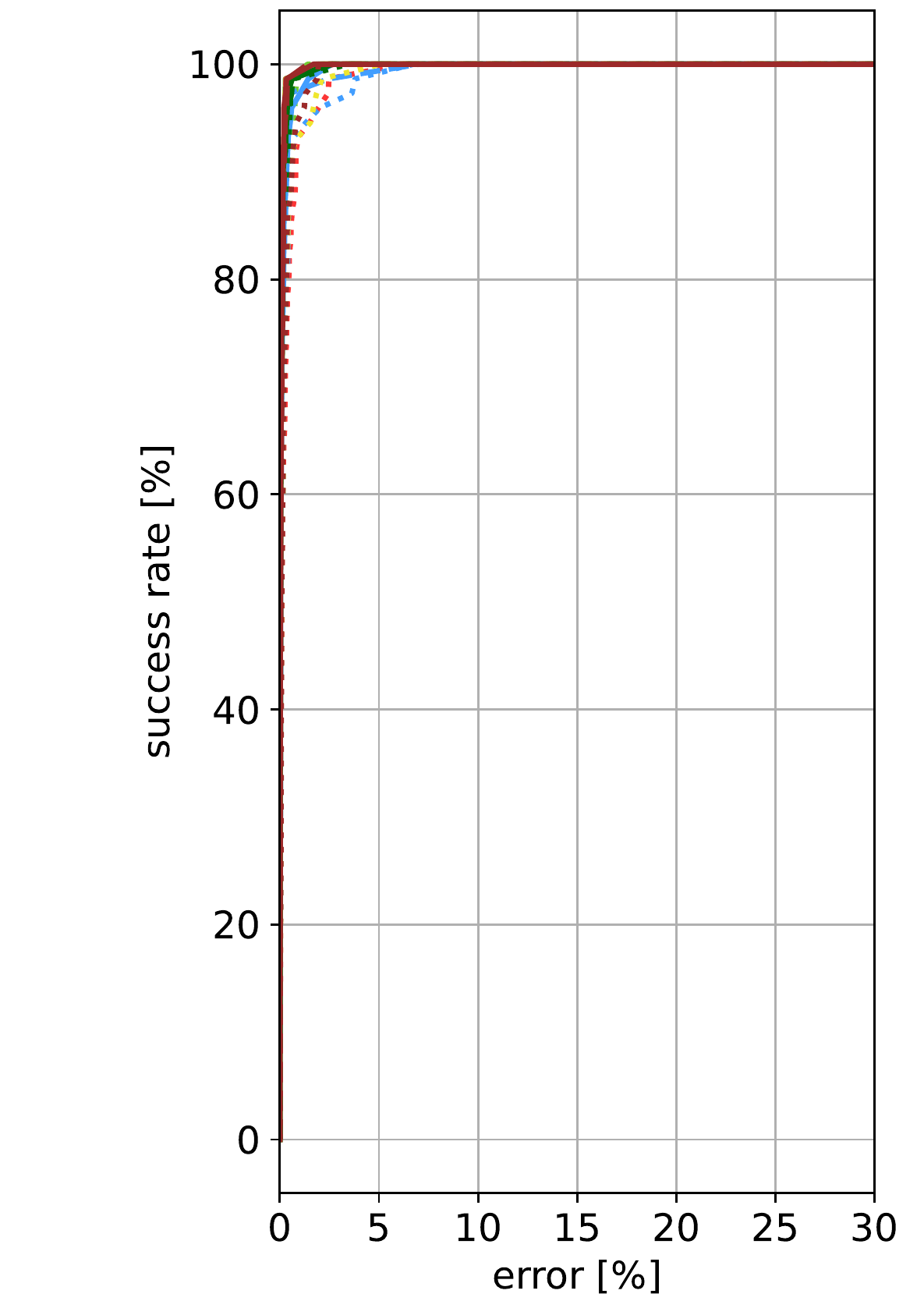}
        \caption{Reichstag}
    \end{subfigure}
    \begin{subfigure}[b]{0.29\textwidth}
        \centering
        \includegraphics[width=\textwidth,trim={3.4cm 1.7cm 8cm 1.4cm},clip]{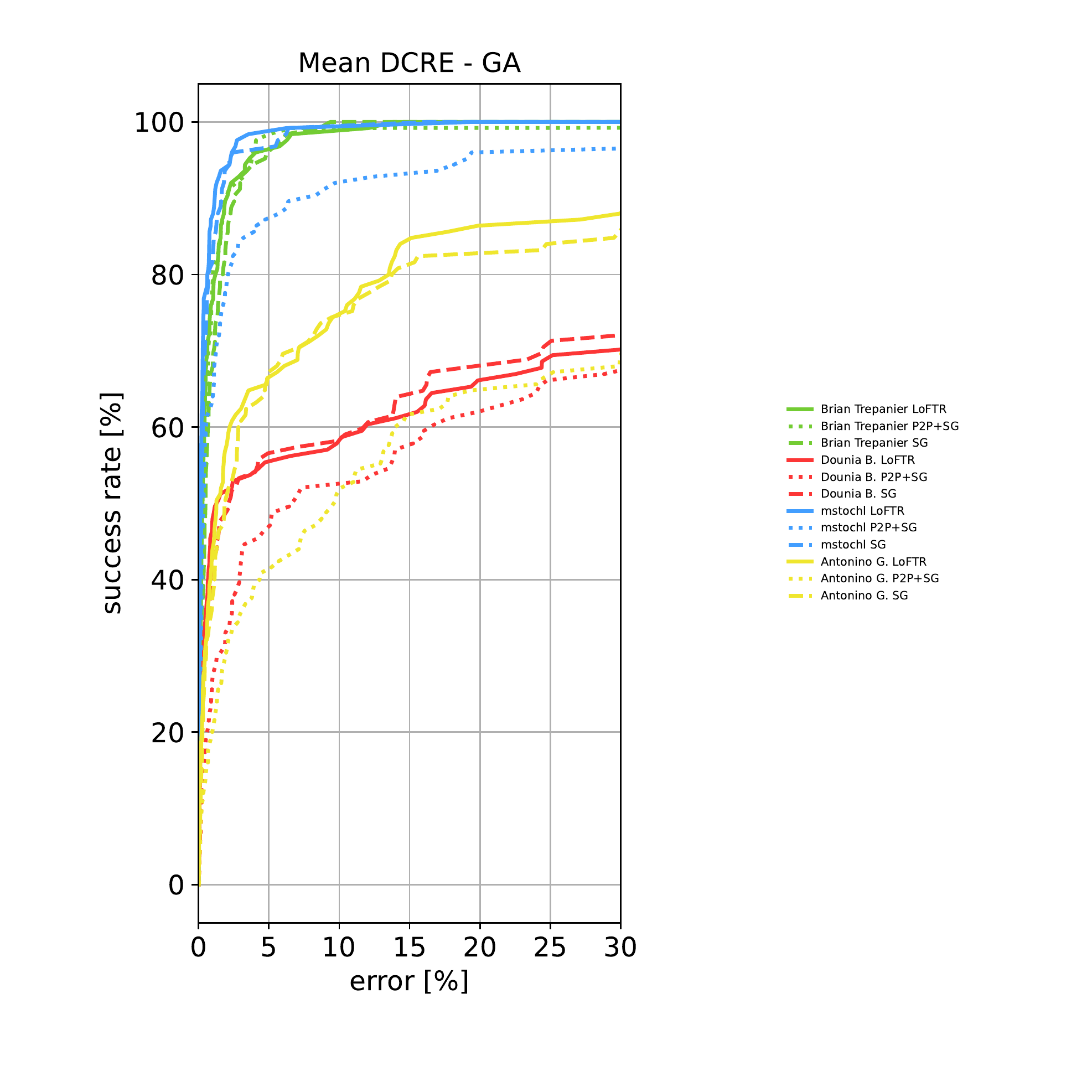}
        \caption{St. Peter's square}
    \end{subfigure}
    \begin{subfigure}[b]{0.29\textwidth}
        \centering
        \includegraphics[width=\textwidth,trim={3.4cm 1.7cm 8cm 0.8cm},clip]{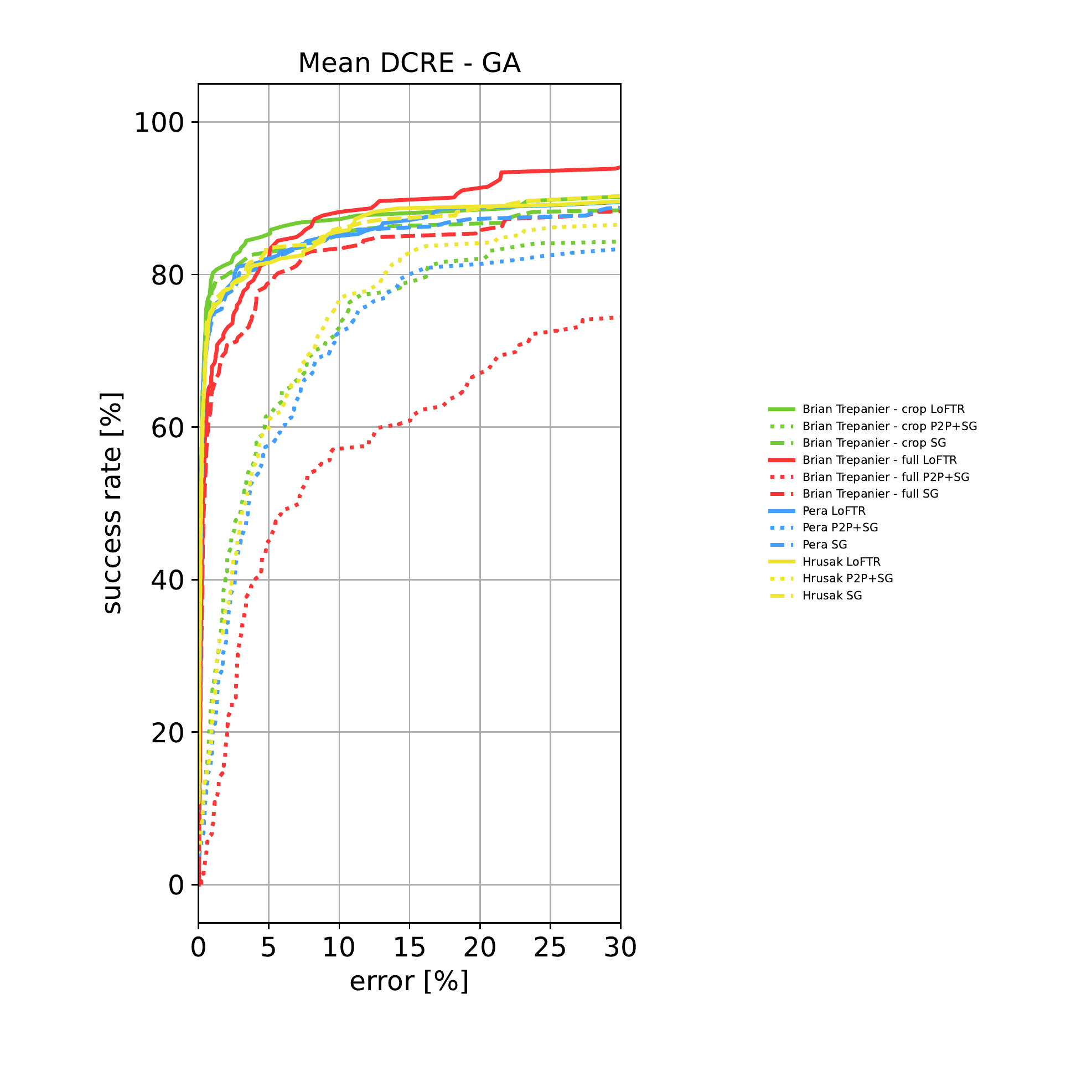}
        \caption{St. Vitus}
    \end{subfigure}
    \caption{Isolating the impact of geometric fidelity by combining real images with geometry from the Internet models. We show cumulative histograms of the mean DCRE, as a percentage of the image diagonal, over all query images in a scene for the ground truth poses obtained via global alignment (GA).}
    \label{fig:eval_dcre_orig_mean_ga}
\end{figure*}

\begin{figure*}[t!]
    \centering
    \begin{subfigure}[b]{0.404\textwidth}
        \centering
        \includegraphics[width=\textwidth,trim={0 2.9cm 8cm 0},clip]{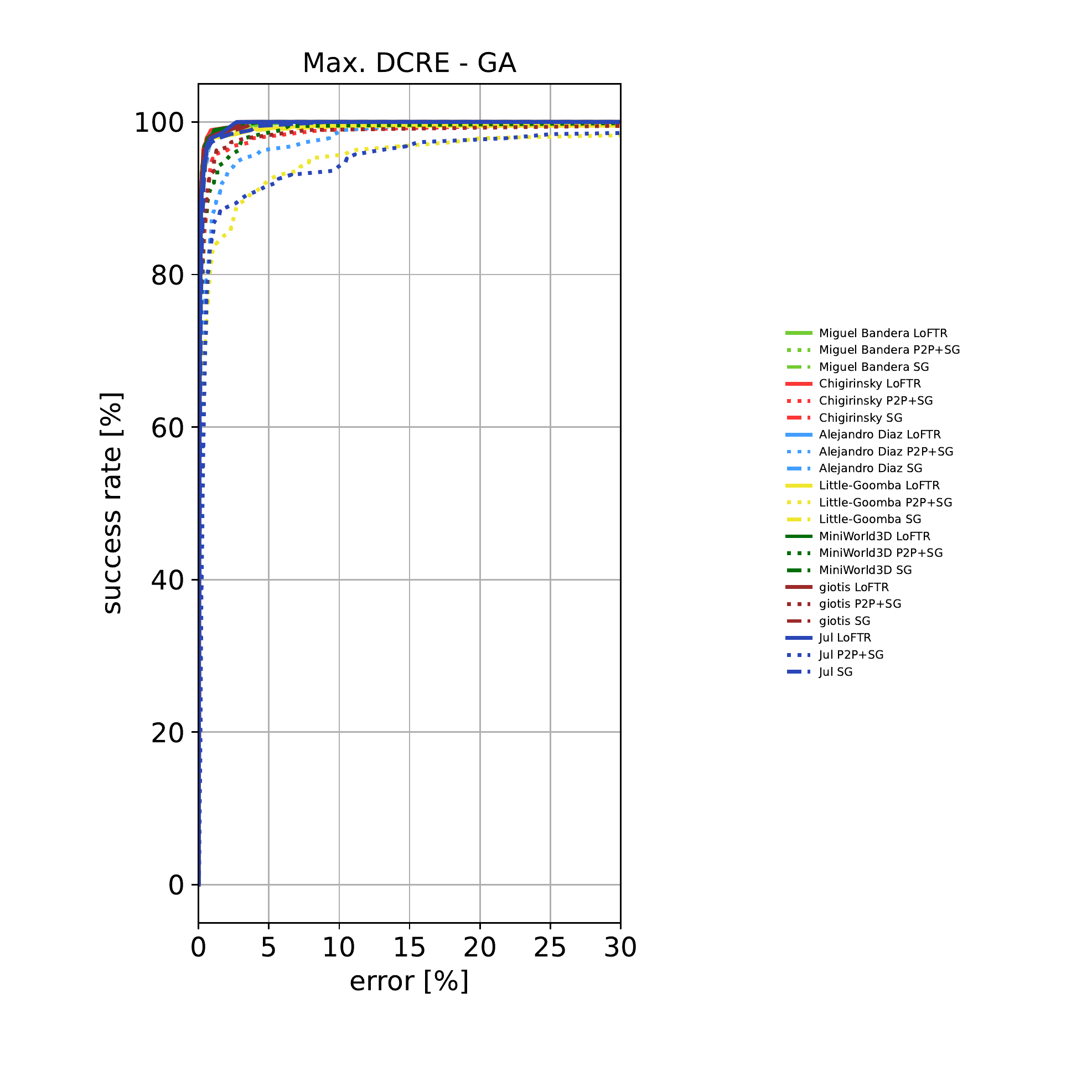}
        \caption{Notre Dame}
    \end{subfigure}
    \begin{subfigure}[b]{0.29\textwidth}
        \centering
        \includegraphics[width=\textwidth,trim={3.4cm 2.9cm 8cm 0},clip]{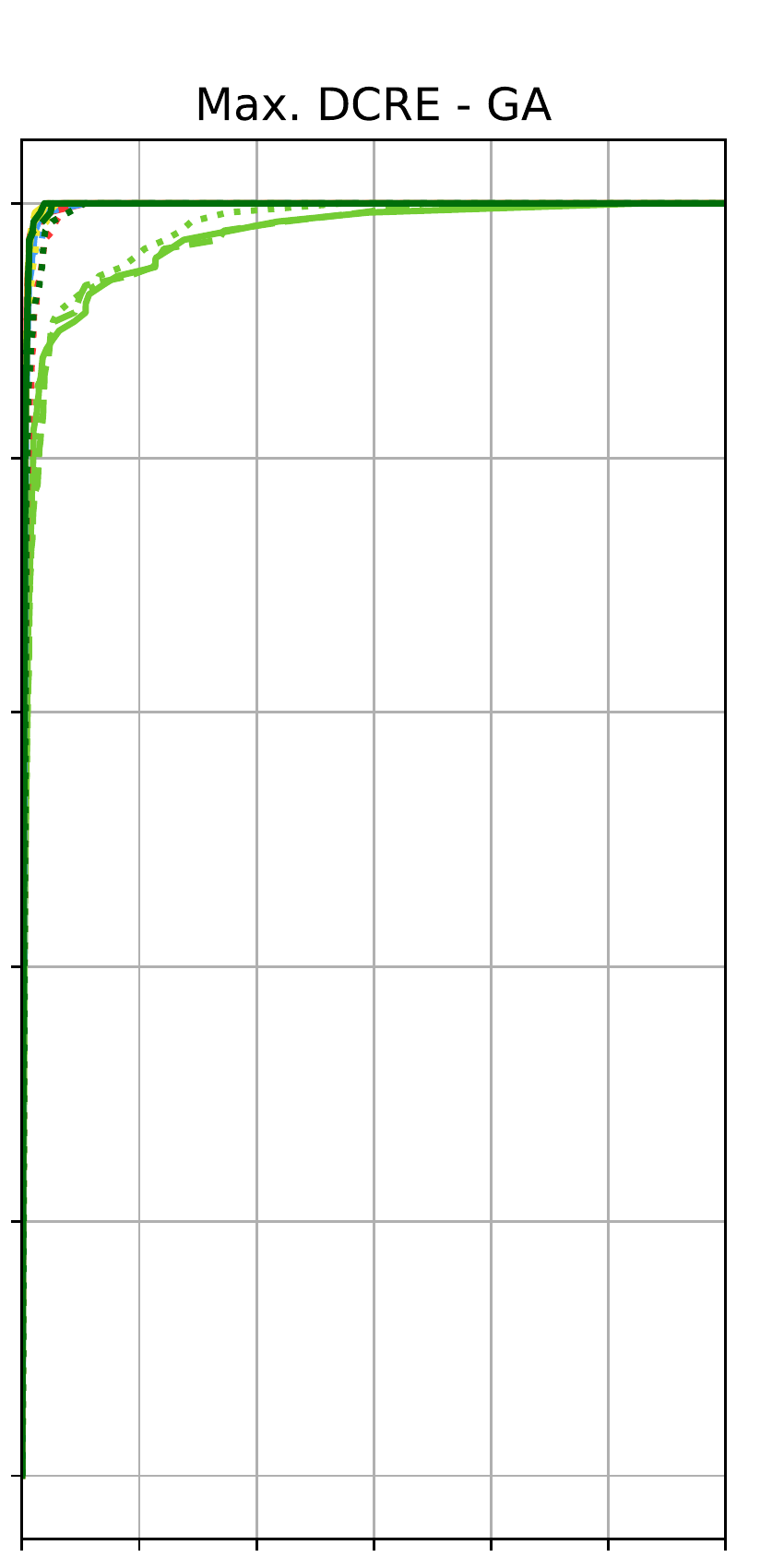}
        \caption{Pantheon}
    \end{subfigure}
    \begin{subfigure}[b]{0.29\textwidth}
        \centering
        \includegraphics[width=\textwidth,trim={-2cm -4cm -2cm 0cm},clip]{figures/example_legend_2x1.pdf}
    \end{subfigure}
    \\
    \begin{subfigure}[b]{0.404\textwidth}
        \centering
        \includegraphics[width=\textwidth,trim={0 1.7cm 8cm 1.4cm},clip]{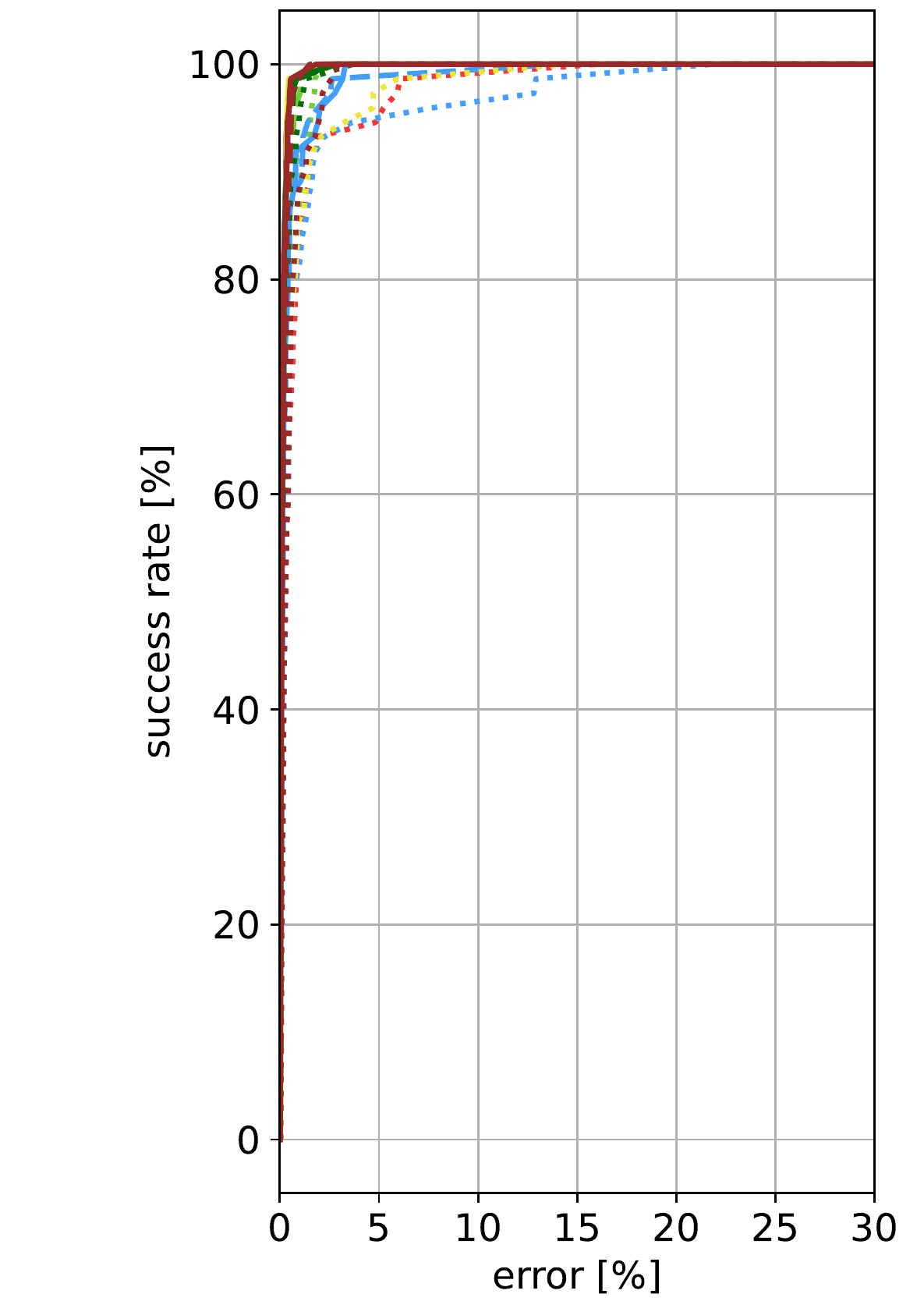}
        \caption{Reichstag}
    \end{subfigure}
    \begin{subfigure}[b]{0.29\textwidth}
        \centering
        \includegraphics[width=\textwidth,trim={3.4cm 1.7cm 8cm 1.4cm},clip]{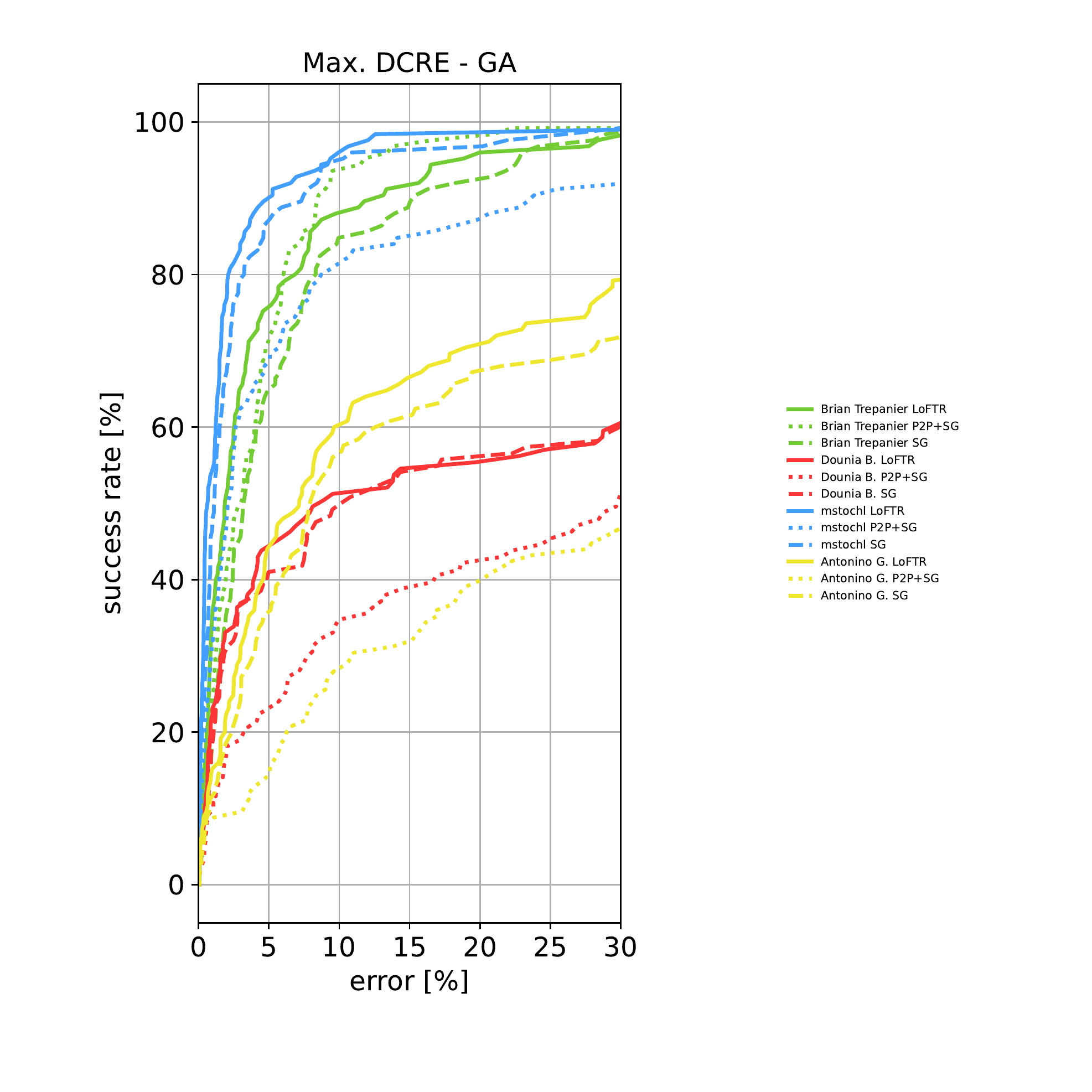}
        \caption{St. Peter's square}
    \end{subfigure}
    \begin{subfigure}[b]{0.29\textwidth}
        \centering
        \includegraphics[width=\textwidth,trim={3.4cm 1.7cm 8cm 0.8cm},clip]{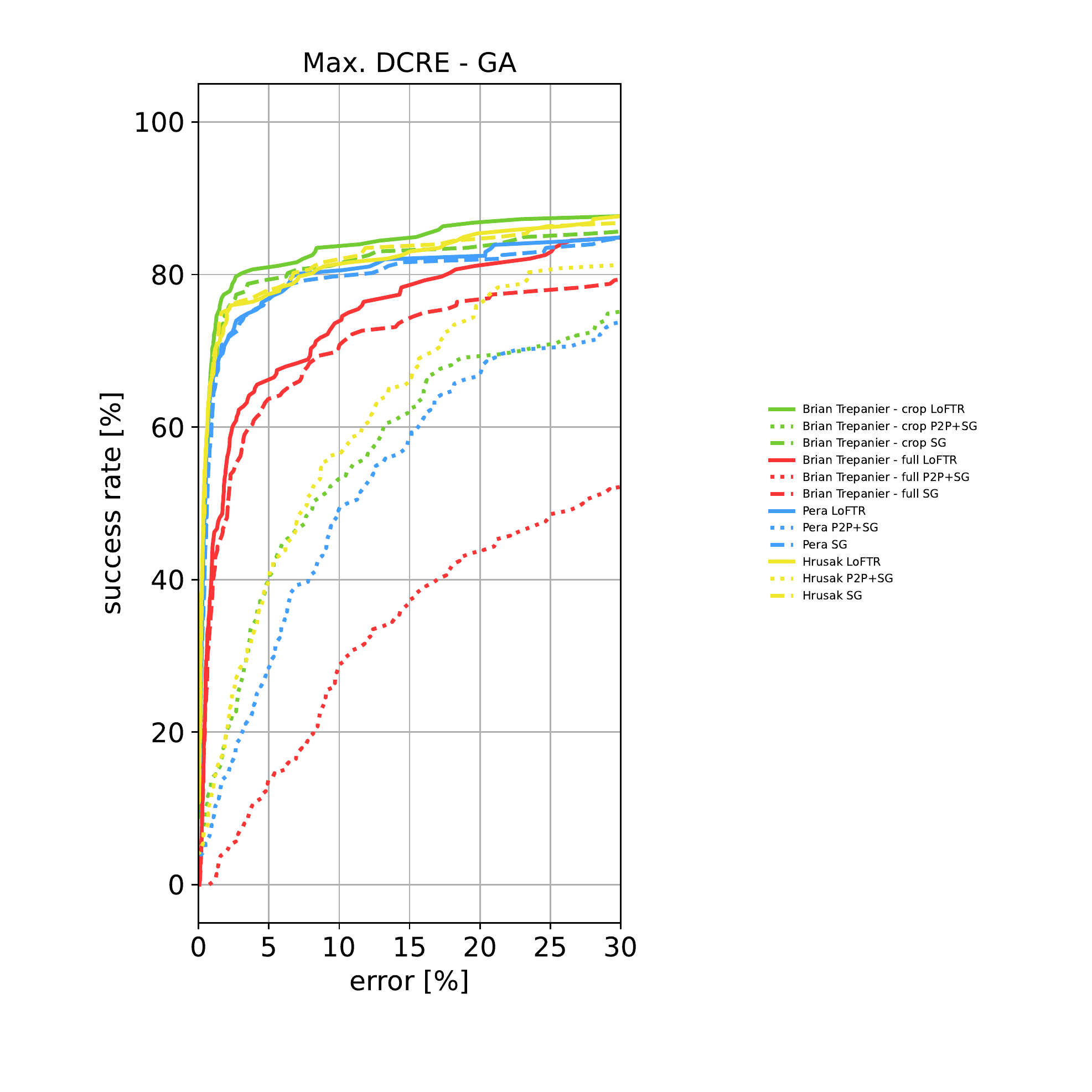}
        \caption{St. Vitus}
    \end{subfigure}
    \caption{Isolating the impact of geometric fidelity by combining real images with geometry from the Internet models. We show cumulative histograms of the maximum DCRE, as a percentage of the image diagonal, over all query images in a scene for the ground truth poses obtained via global alignment (GA).}
    \label{fig:eval_dcre_orig_max_ga}
\end{figure*}

\begin{figure*}[t!]
    \centering
    \begin{subfigure}[b]{0.404\textwidth}
        \centering
        \includegraphics[width=\textwidth,trim={0 2.9cm 8cm 0},clip]{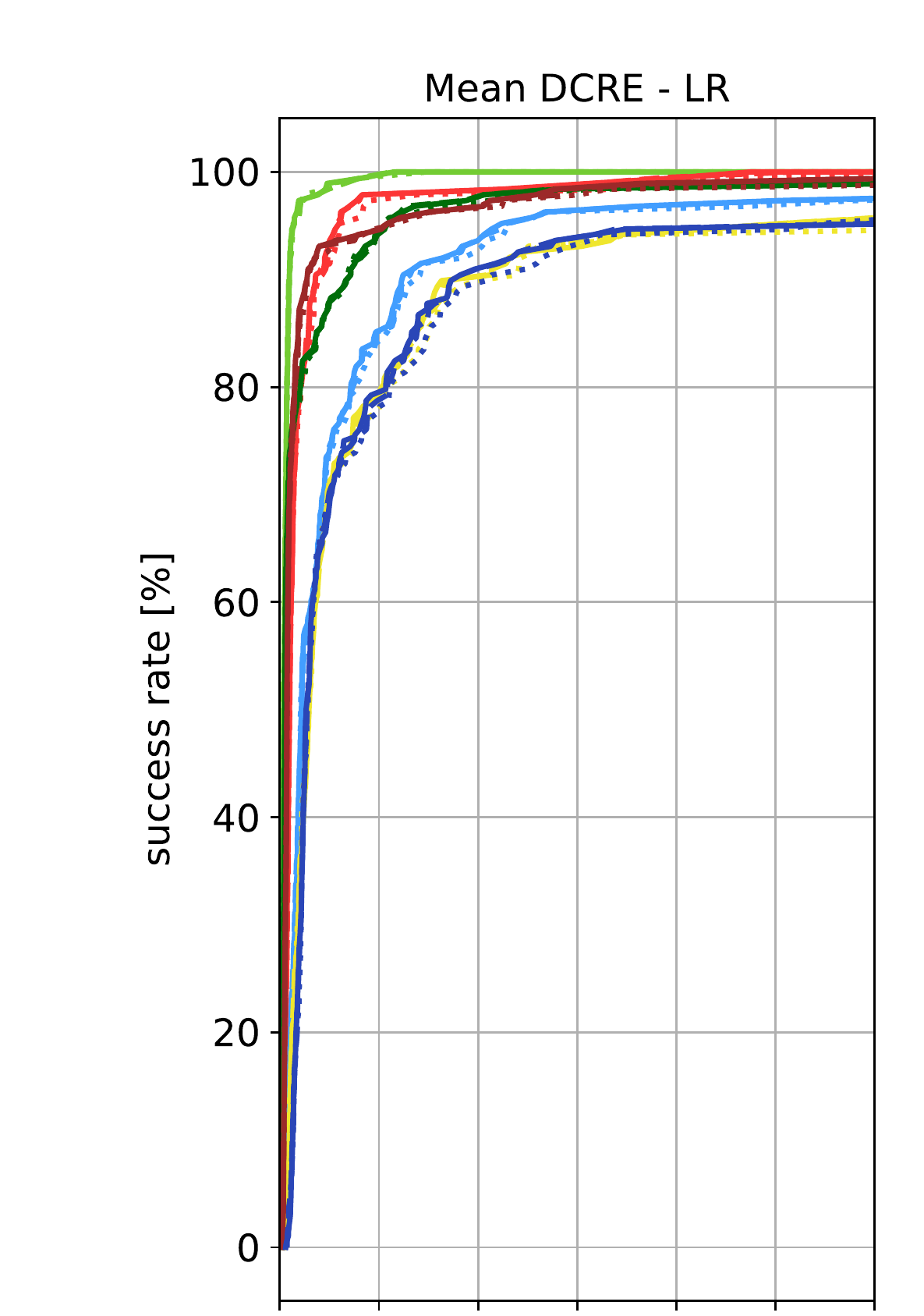}
        \caption{Notre Dame}
    \end{subfigure}
    \begin{subfigure}[b]{0.29\textwidth}
        \centering
        \includegraphics[width=\textwidth,trim={3.4cm 2.9cm 8cm 0},clip]{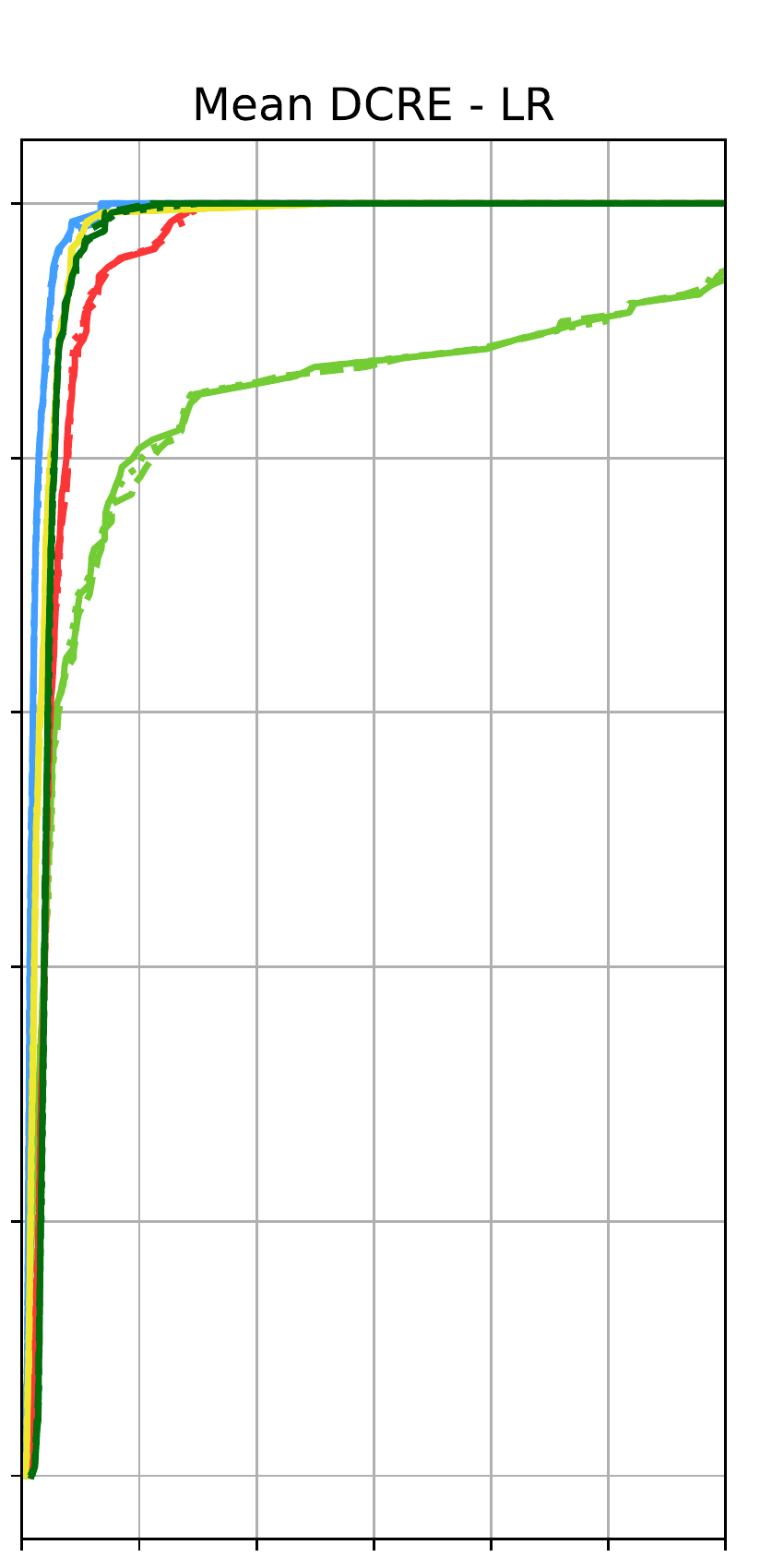}
        \caption{Pantheon}
    \end{subfigure}
    \begin{subfigure}[b]{0.29\textwidth}
        \centering
        \includegraphics[width=\textwidth,trim={-2cm -4cm -2cm 0cm},clip]{figures/example_legend_2x1.pdf}
    \end{subfigure}
    \\
    \begin{subfigure}[b]{0.404\textwidth}
        \centering
        \includegraphics[width=\textwidth,trim={0 1.7cm 8cm 1.4cm},clip]{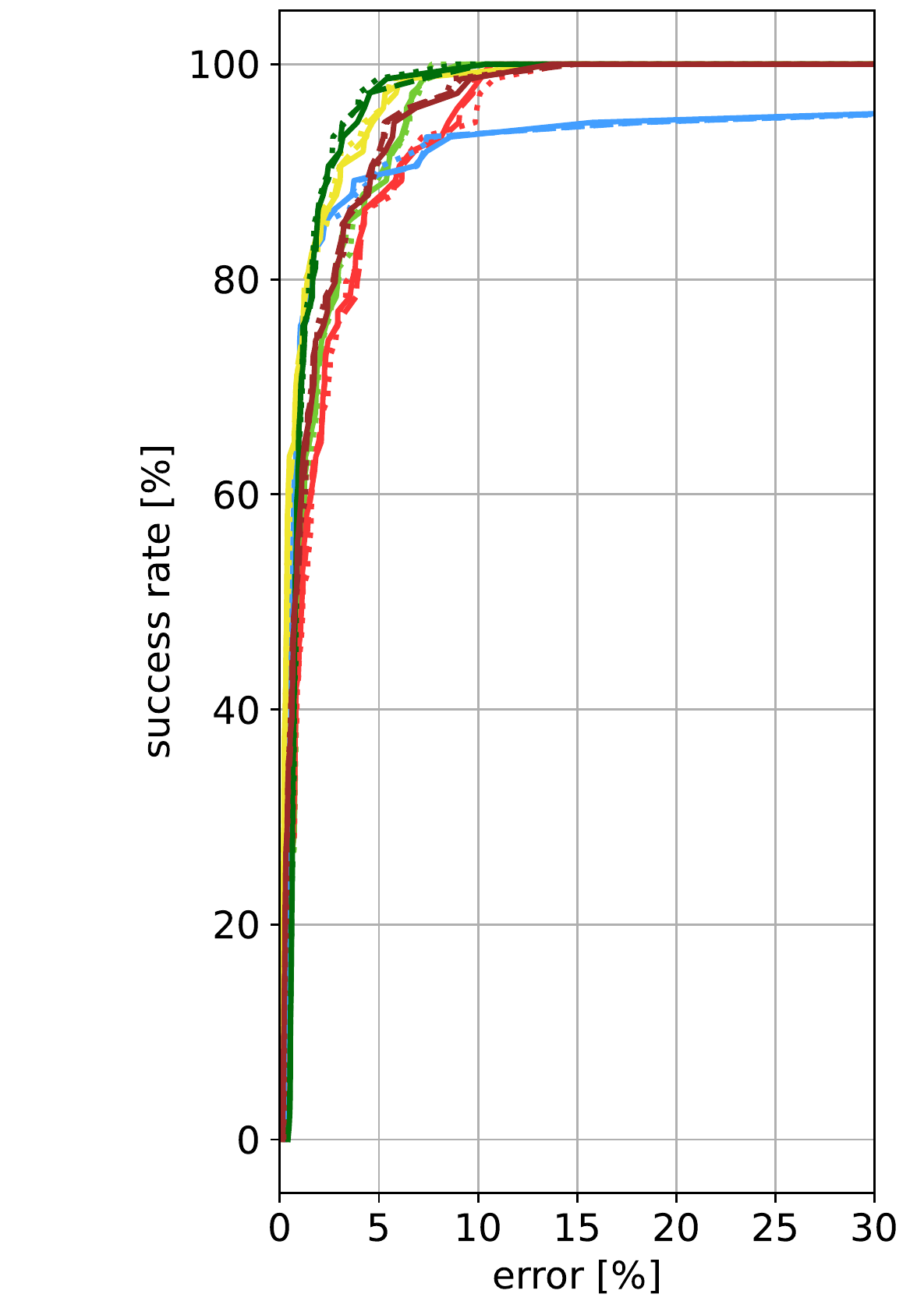}
        \caption{Reichstag}
    \end{subfigure}
    \begin{subfigure}[b]{0.29\textwidth}
        \centering
        \includegraphics[width=\textwidth,trim={3.4cm 1.7cm 8cm 1.4cm},clip]{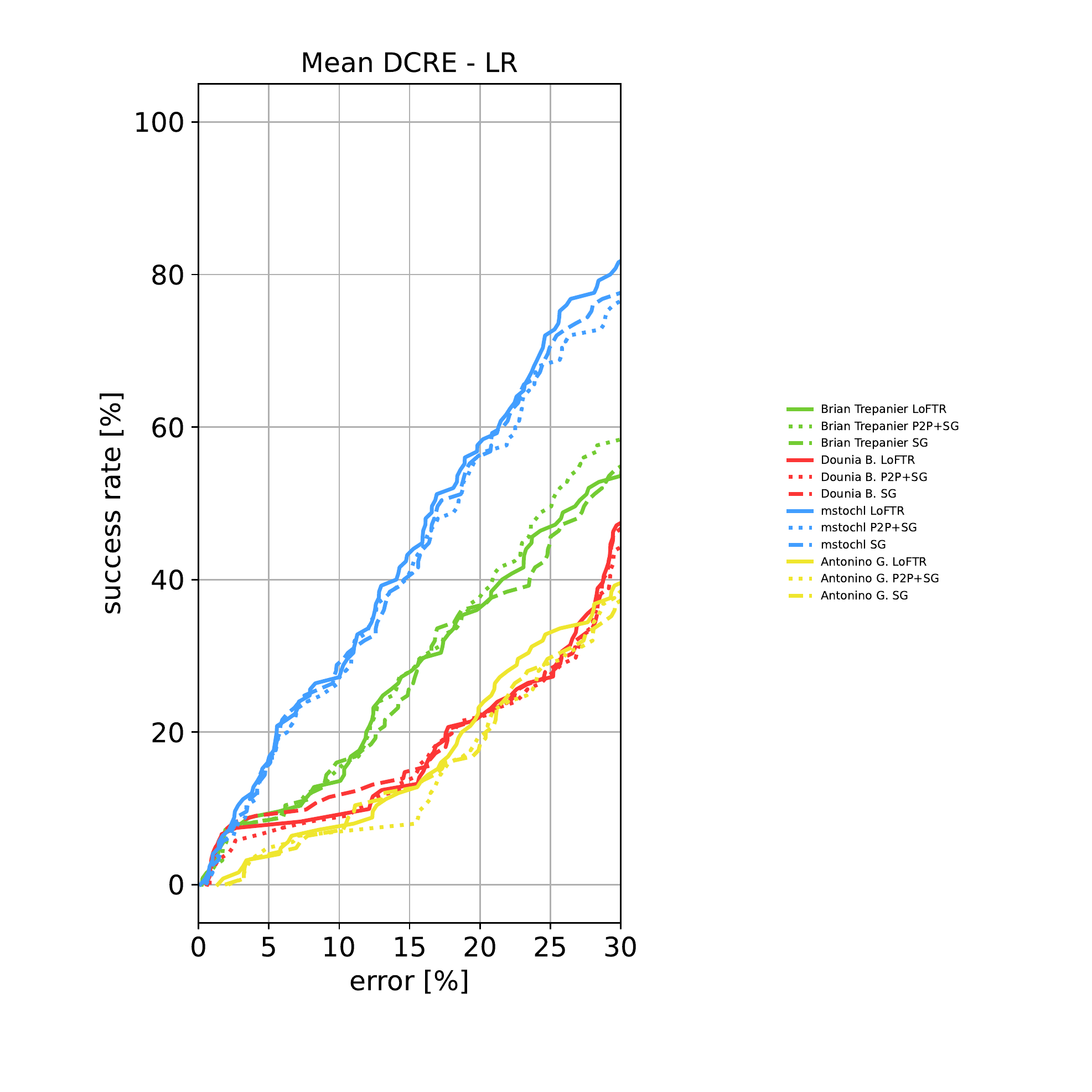}
        \caption{St. Peter's square}
    \end{subfigure}
    \begin{subfigure}[b]{0.29\textwidth}
        \centering
        \includegraphics[width=\textwidth,trim={3.4cm 1.7cm 8cm 0.8cm},clip]{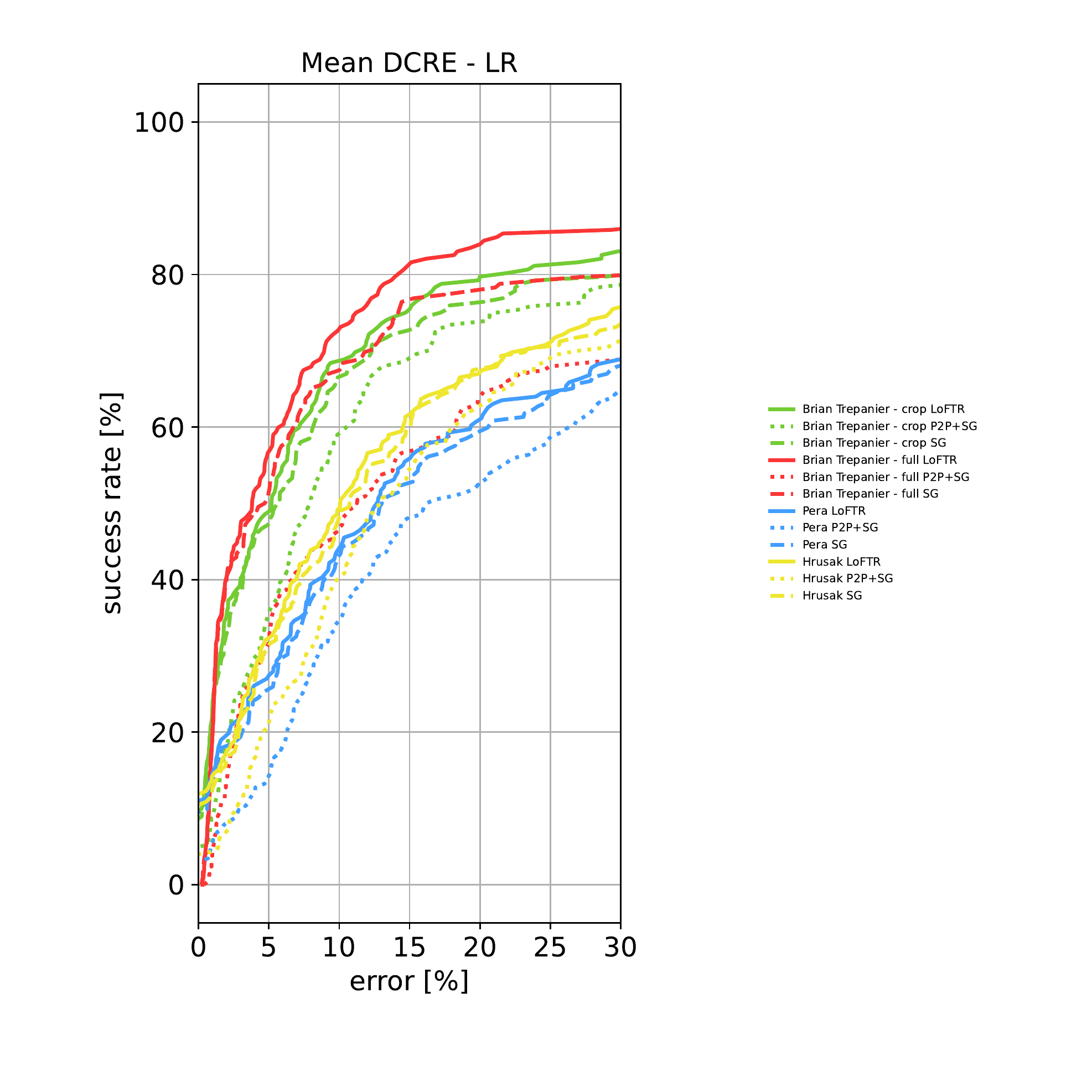}
        \caption{St. Vitus}
    \end{subfigure}
    \caption{Isolating the impact of geometric fidelity by combining real images with geometry from the Internet models. We show cumulative histograms of the mean DCRE, as a percentage of the image diagonal, over all query images in a scene for the ground truth poses obtained via local refinement (LR).}
    \label{fig:eval_dcre_orig_mean_lr}
\end{figure*}

\begin{figure*}[t!]
    \centering
    \begin{subfigure}[b]{0.404\textwidth}
        \centering
        \includegraphics[width=\textwidth,trim={0 2.9cm 8cm 0},clip]{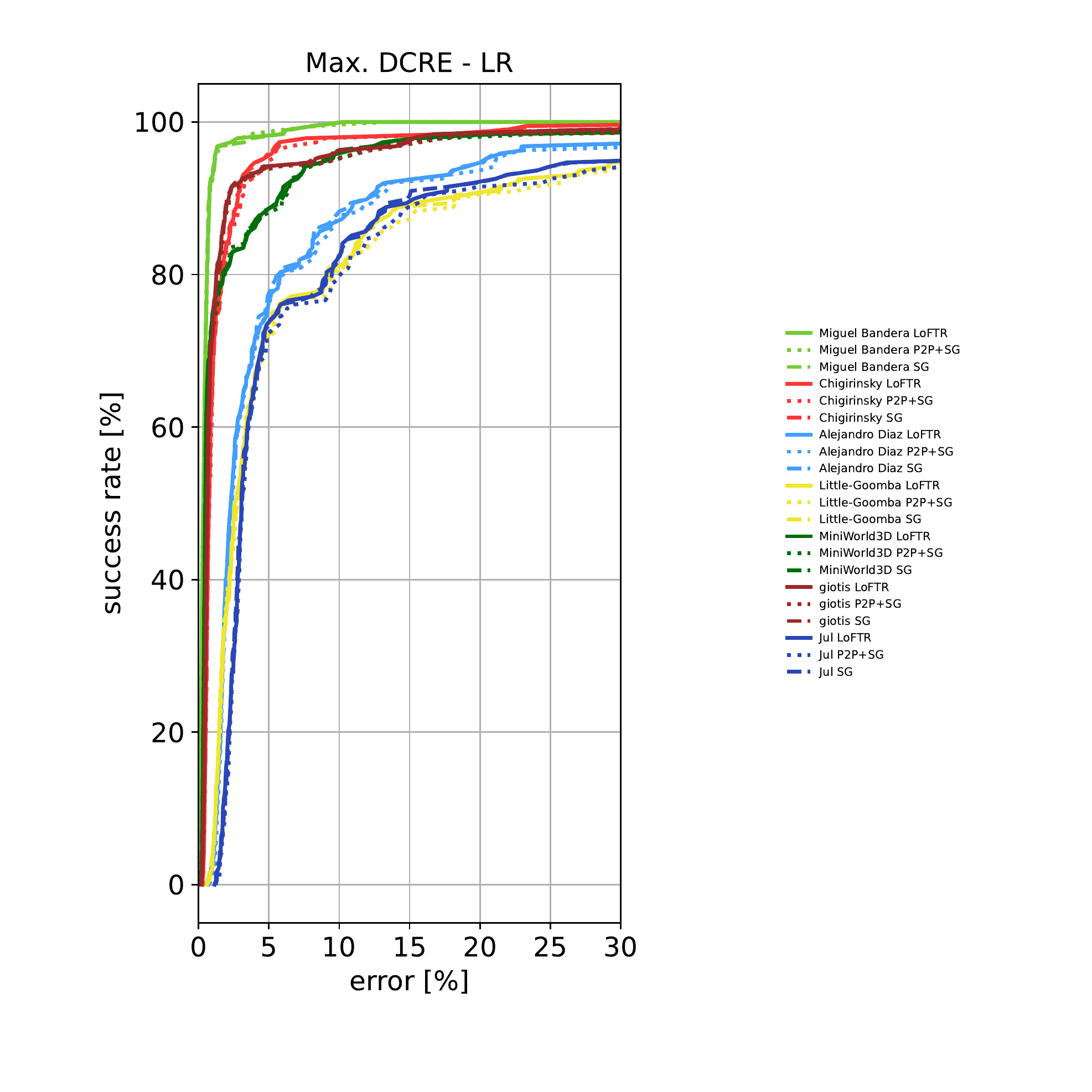}
        \caption{Notre Dame}
    \end{subfigure}
    \begin{subfigure}[b]{0.29\textwidth}
        \centering
        \includegraphics[width=\textwidth,trim={3.4cm 2.9cm 8cm 0},clip]{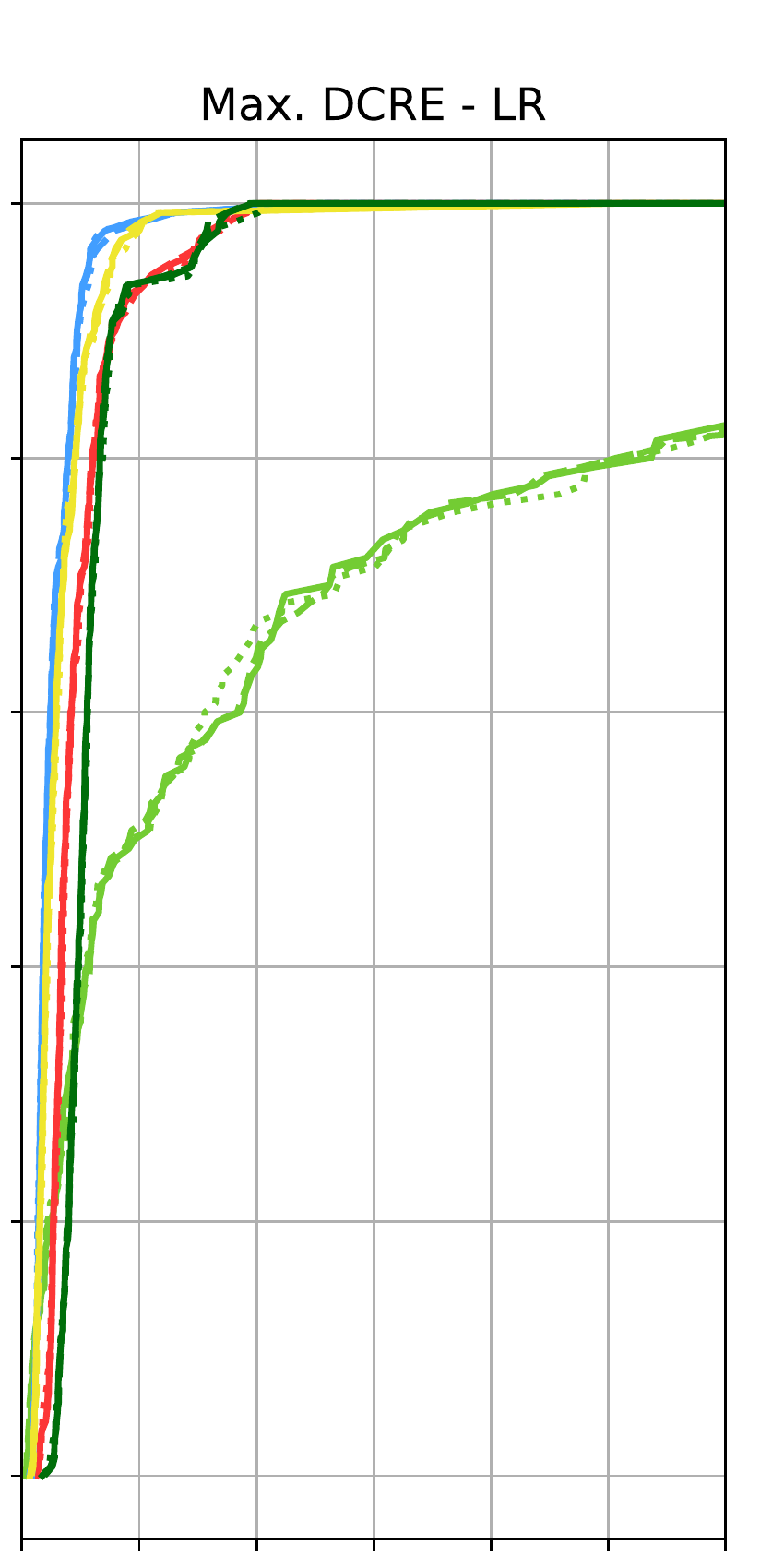}
        \caption{Pantheon}
    \end{subfigure}
    \begin{subfigure}[b]{0.29\textwidth}
        \centering
        \includegraphics[width=\textwidth,trim={-2cm -4cm -2cm 0cm},clip]{figures/example_legend_2x1.pdf}
    \end{subfigure}
    \\
    \begin{subfigure}[b]{0.404\textwidth}
        \centering
        \includegraphics[width=\textwidth,trim={0 1.7cm 8cm 1.4cm},clip]{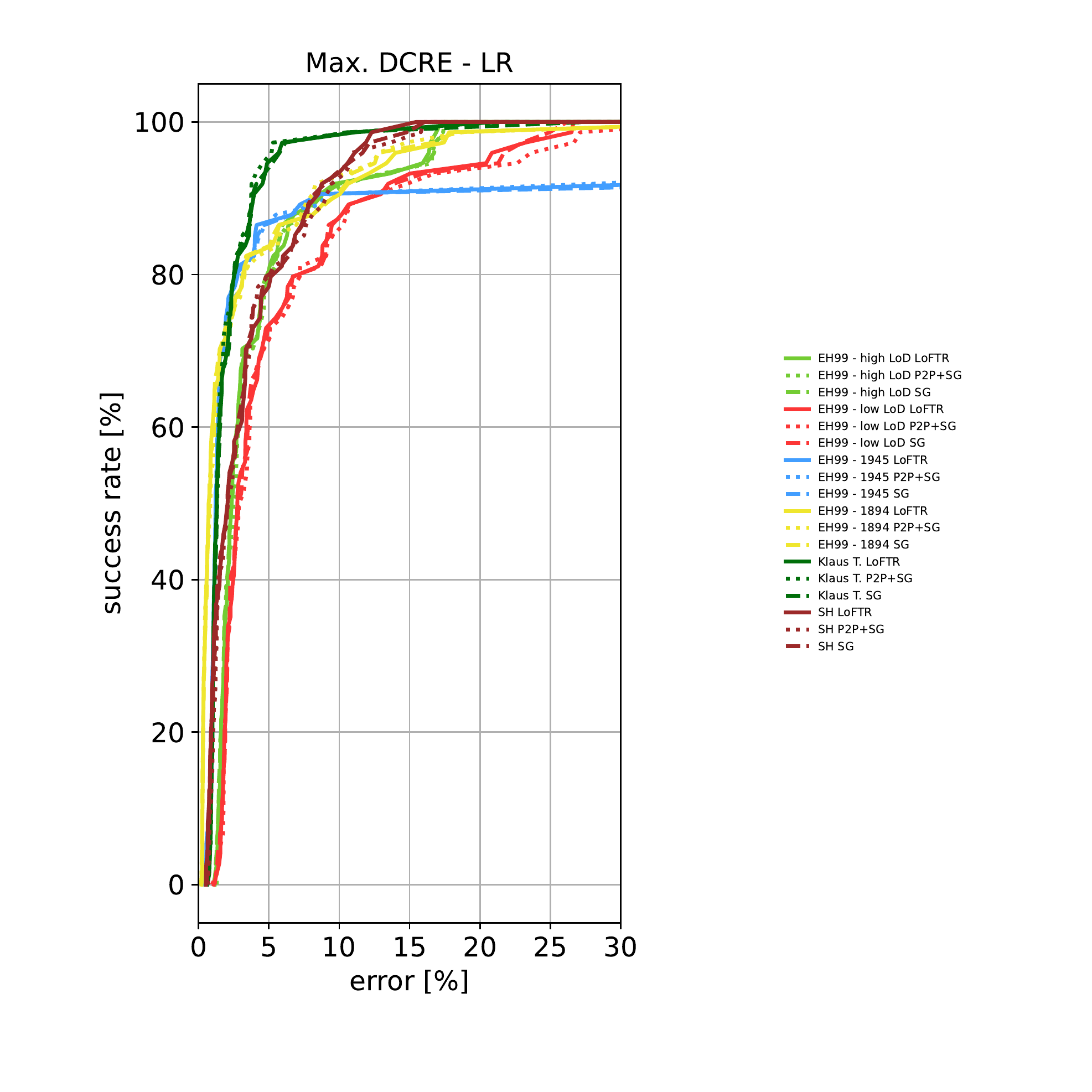}
        \caption{Reichstag}
    \end{subfigure}
    \begin{subfigure}[b]{0.29\textwidth}
        \centering
        \includegraphics[width=\textwidth,trim={3.4cm 1.7cm 8cm 1.4cm},clip]{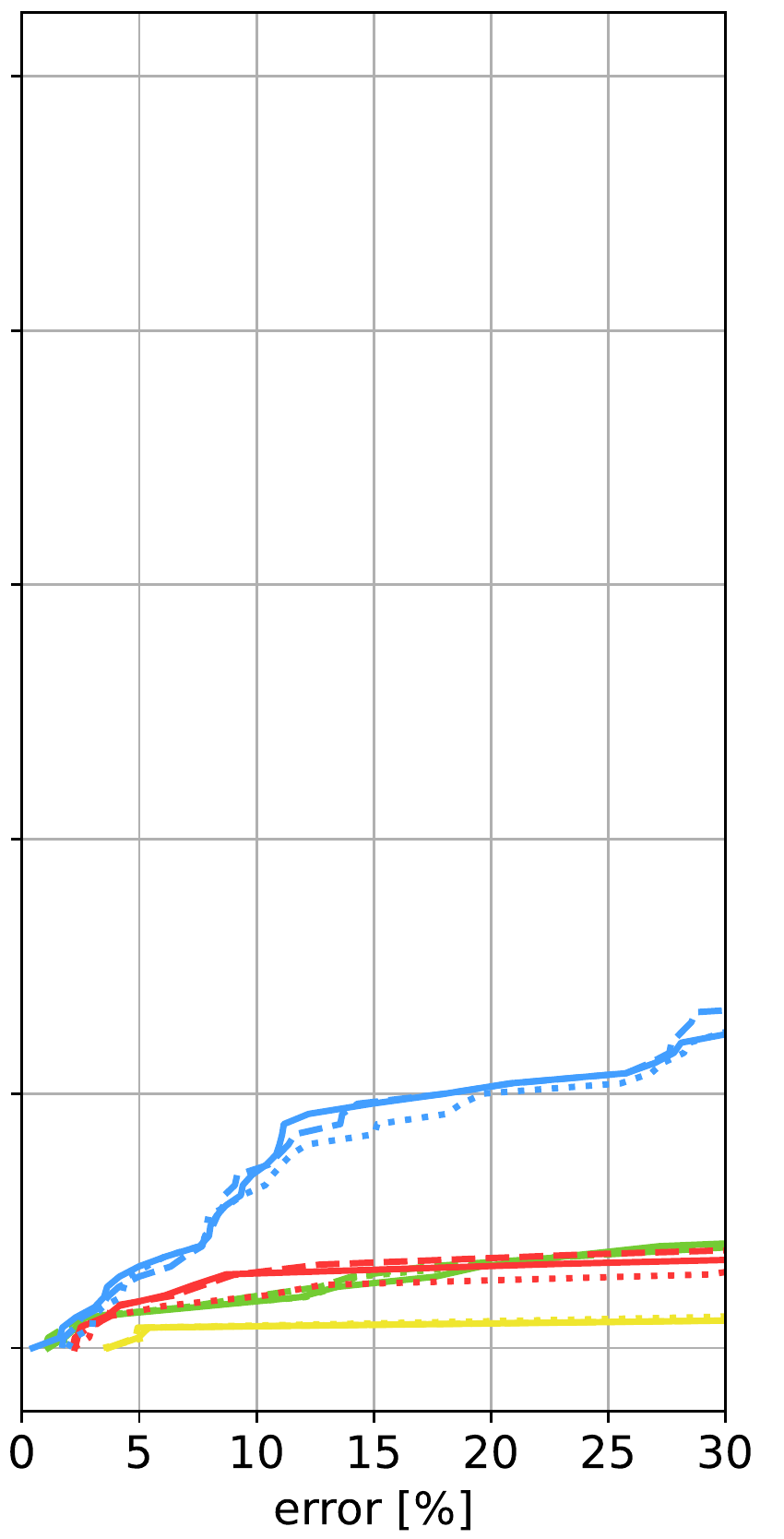}
        \caption{St. Peter's square}
    \end{subfigure}
    \begin{subfigure}[b]{0.29\textwidth}
        \centering
        \includegraphics[width=\textwidth,trim={3.4cm 1.7cm 8cm 0.8cm},clip]{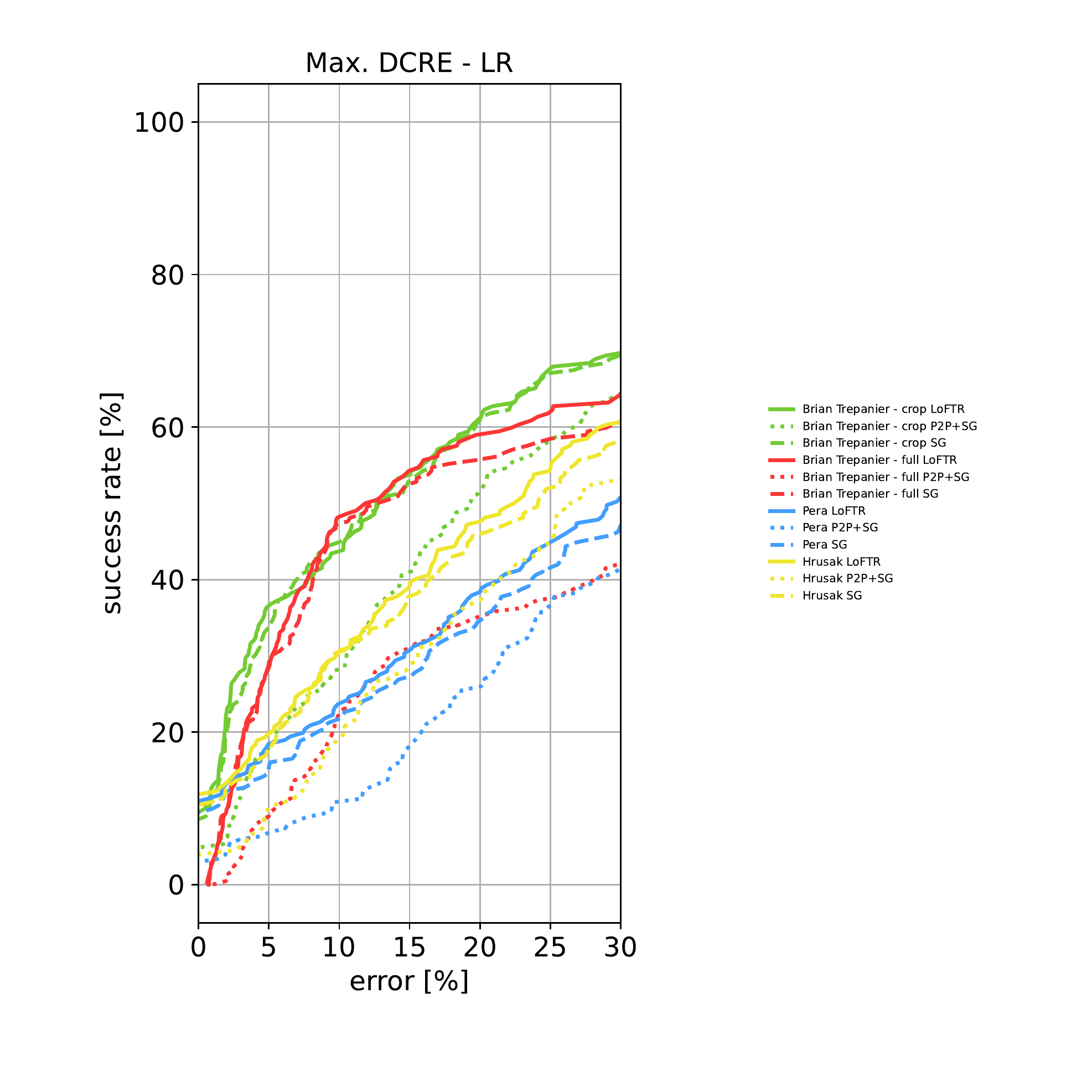}
        \caption{St. Vitus}
    \end{subfigure}
    \caption{Isolating the impact of geometric fidelity by combining real images with geometry from the Internet models. We show cumulative histograms of the maximum DCRE, as a percentage of the image diagonal, over all query images in a scene for the ground truth poses obtained via local refinement (LR).}
    \label{fig:eval_dcre_orig_max_lr}
\end{figure*}

We show the results with mean DCRE and global alignment (GA) ground truth in \cref{fig:eval_dcre_orig_mean_ga} (which is a reproduction of Fig.~\ref{fig:eval_dcre_orig_imgs_cad_depth}) and with local refinement (LR) in \cref{fig:eval_dcre_orig_mean_lr}. We also show maximum DCREs with GA in \cref{fig:eval_dcre_orig_max_ga} and with LR in \cref{fig:eval_dcre_orig_max_lr}. 
\cref{tab:datasets} associates the model IDs to the individual models. 

Compared to the results presented in Sec.~\ref{sec:supp_exps}, the gap between the mean and maximum DCRE curves is significantly smaller when matching against real images and only using the 3D geometry of the Internet models (in the form of rendered depth maps). As already mentioned in the main paper, the results show that finding sufficiently many matches to facilitate accurate pose estimation seems to be the main bottleneck, even if the underlying geometry is rather coarse.
Thus, we can attribute the majority of the outliers skewing the maximum DCRE curves shown in \cref{sec:supp_exps} to the feature matching stage.

\begin{figure*}[t!]
    \centering
    \begin{subfigure}[b]{0.295\textwidth}
        \centering
        \includegraphics[width=\textwidth,trim={1.5cm 1.7cm 8cm 0},clip]{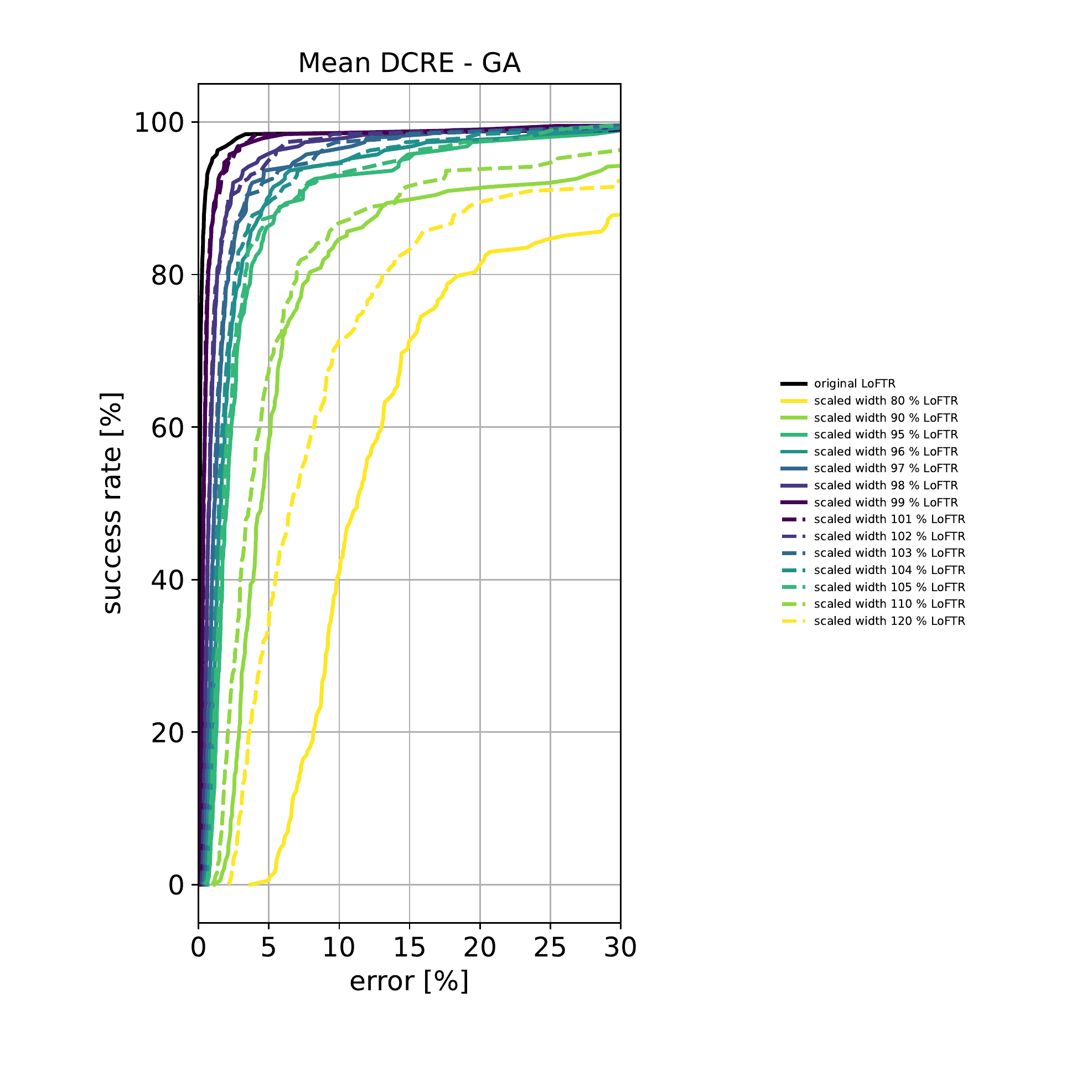}
        \caption{width scaling}
    \end{subfigure}
    \hfill
    \begin{subfigure}[b]{0.239\textwidth}
        \centering
        \includegraphics[width=\textwidth,trim={3.5cm 1.7cm 8cm 0},clip]{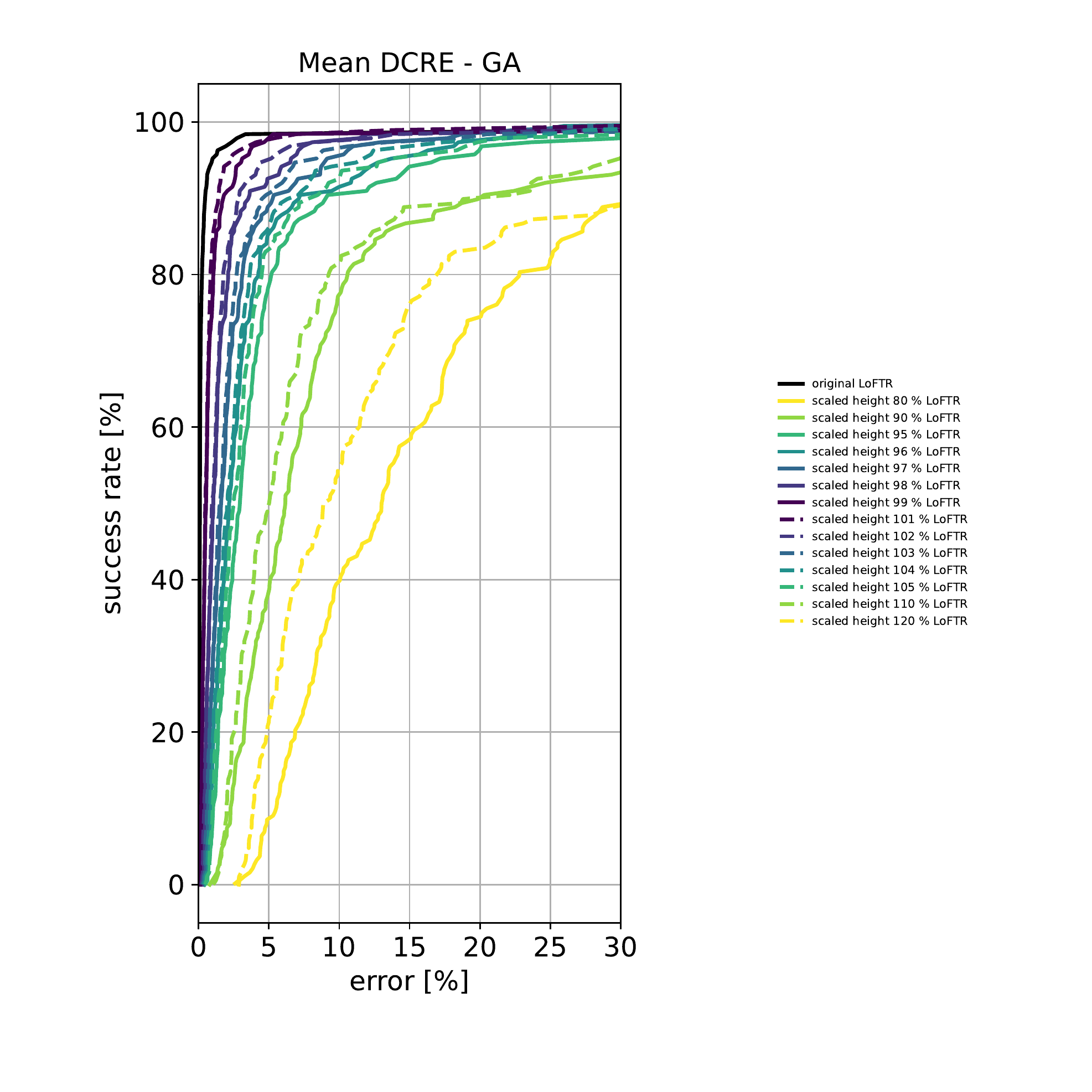}
        \caption{height scaling}
    \end{subfigure}
    \hfill
    \begin{subfigure}[b]{0.239\textwidth}
        \centering
        \includegraphics[width=\textwidth,trim={3.5cm 1.7cm 8cm 0},clip]{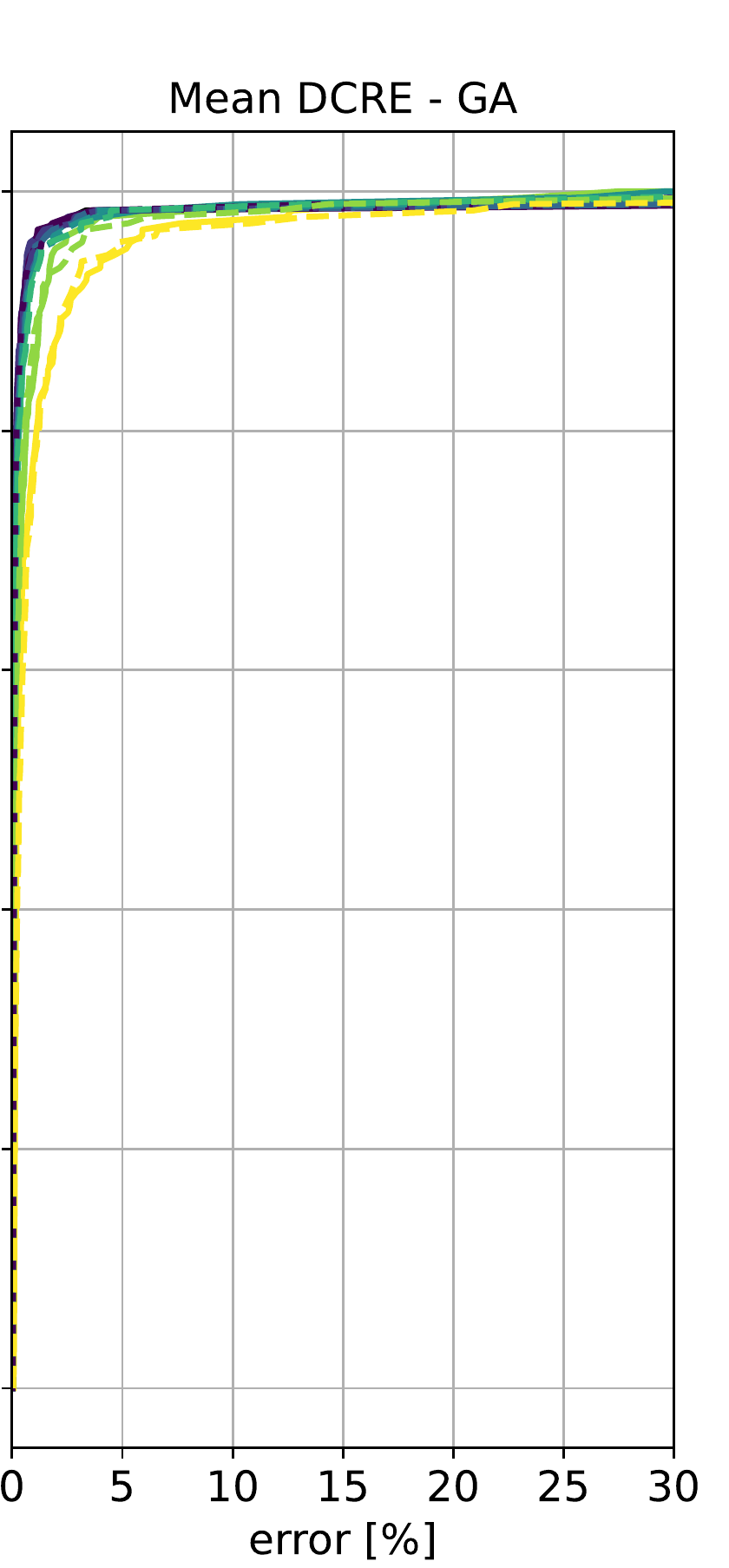}
        \caption{depth scaling}
    \end{subfigure}
    \hfill
    \begin{subfigure}[b]{0.21\textwidth}
        \centering
        \includegraphics[width=\textwidth,trim={14.5cm 6.3cm 2.5cm 6.5cm},clip]{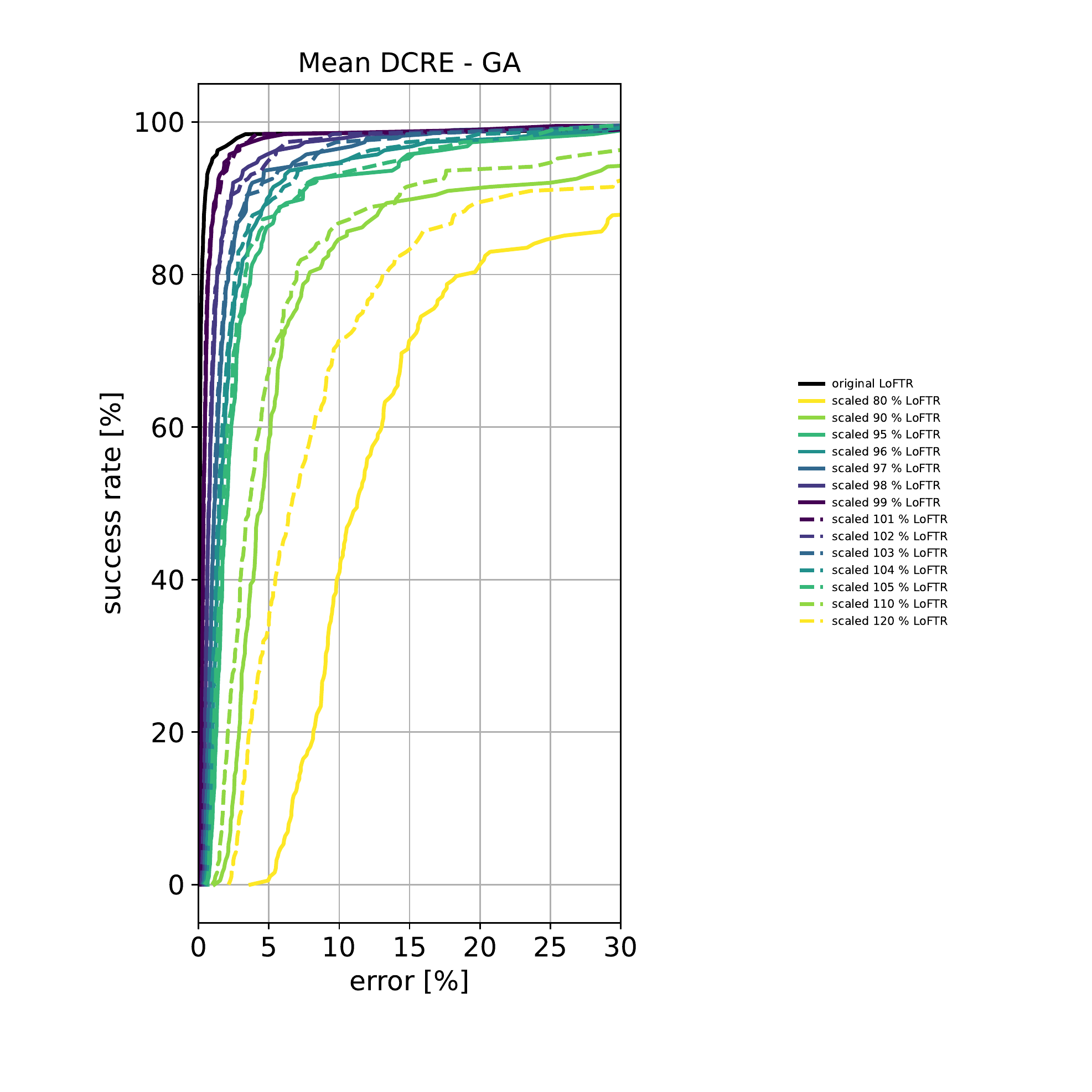}
    \end{subfigure}
    
    \caption{Results from Fig.~\ref{fig:scaling_exp}, using more intermediate scaling steps. Isolating the impact of geometric fidelity by applying non-uniform scaling on the 3D model. We show cumulative histograms of the mean DCRE, as a percentage of the image diagonal, over all query images in a scene for the ground truth poses obtained via global alignment (GA).}
    \label{fig:stretch_fine}
\end{figure*}

To directly observe the influence of the geometric fidelity, we further experimented with non-uniformly scaling the most precise model (Notre Dame A) to measure the impact of a changing aspect ratio on localization accuracy. The non-uniform scaling in width and height was done relative to the center of the model bounding box. The scaling in depth direction had a center in the main plane of the building's front facade (see \cref{fig:stretch_comparison} for visualization). 
\cref{fig:stretch_fine} extends the results from Fig.~\ref{fig:scaling_exp} by using intermediate scaling factors. The main conclusion drawn in the paper, that changing the aspect ratio significantly reduces localization accuracy, remains valid. 

\section{Ablation Study: Simplifying the Representation}
\label{sec:simplification}
To better understand the influence of the level of geometric and visual fidelity on localization accuracy, we experiment with reducing the geometric resolution (number of faces) and texture resolution of the Notre Dame A model already used in the main paper for ablation studies.

We used Quadric Mesh Collapse Decimation algorithm~\cite{garland1997surface, garland1998simplifying}, implemented in MeshLab~\cite{Cignoni2008MeshLabAO}, for geometry simplification. Note that even when we use the version of the algorithm that is supposed to be more suitable for meshes with textures~\cite{garland1998simplifying}, the textures are significantly distorted during the simplification. Therefore, the appearance fidelity is not completely isolated from the geometric simplification. The other way around, the reduction of texture resolution does not influence the geometry of the model. The texture simplification was done by downsampling all the texture files in the model. We also combined the models with the simplified geometry with the downsampled textures at the same simplification ratio, \eg, the model with half the number of the original faces is combined with the textures with half the original width and height.

The renderings of the simplified models are shown in Fig.~\ref{fig:simplification_comparison}. Note the distortion of the texture present in the models with higher levels of geometry simplification. We did not observe such artifacts in models available on the Internet. 
As such, the results obtained for these severe distortions are not indicative of real-world performance. 

\begin{figure*}[t!]
    \centering
    \begin{subfigure}[b]{0.295\textwidth}
        \centering
        \includegraphics[width=\textwidth,trim={1.5cm 1.7cm 8cm 0},clip]{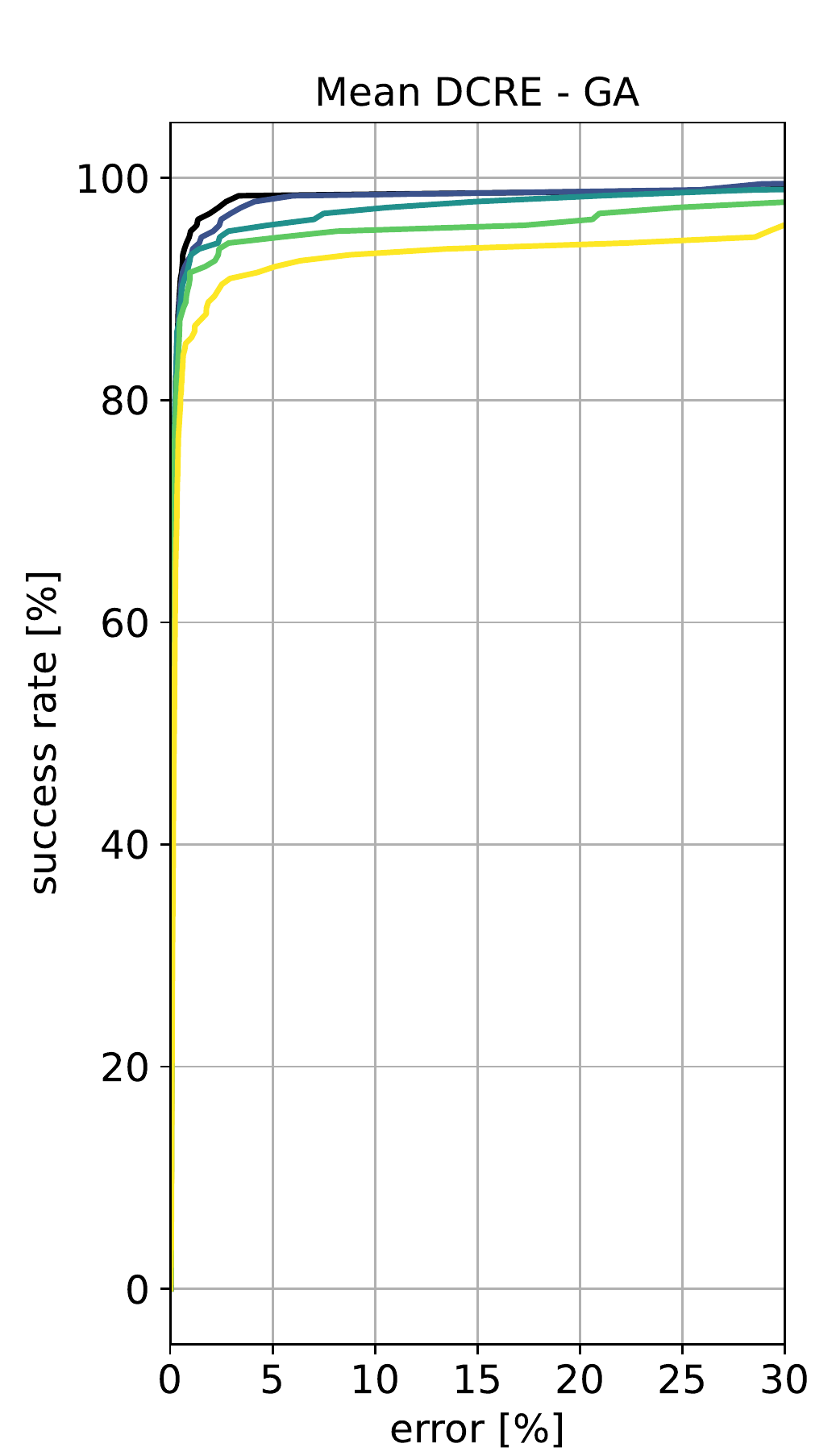}
        \caption{geometry simplification}
    \end{subfigure}
    \hfill
    \begin{subfigure}[b]{0.239\textwidth}
        \centering
        \includegraphics[width=\textwidth,trim={3.5cm 1.7cm 8cm 0},clip]{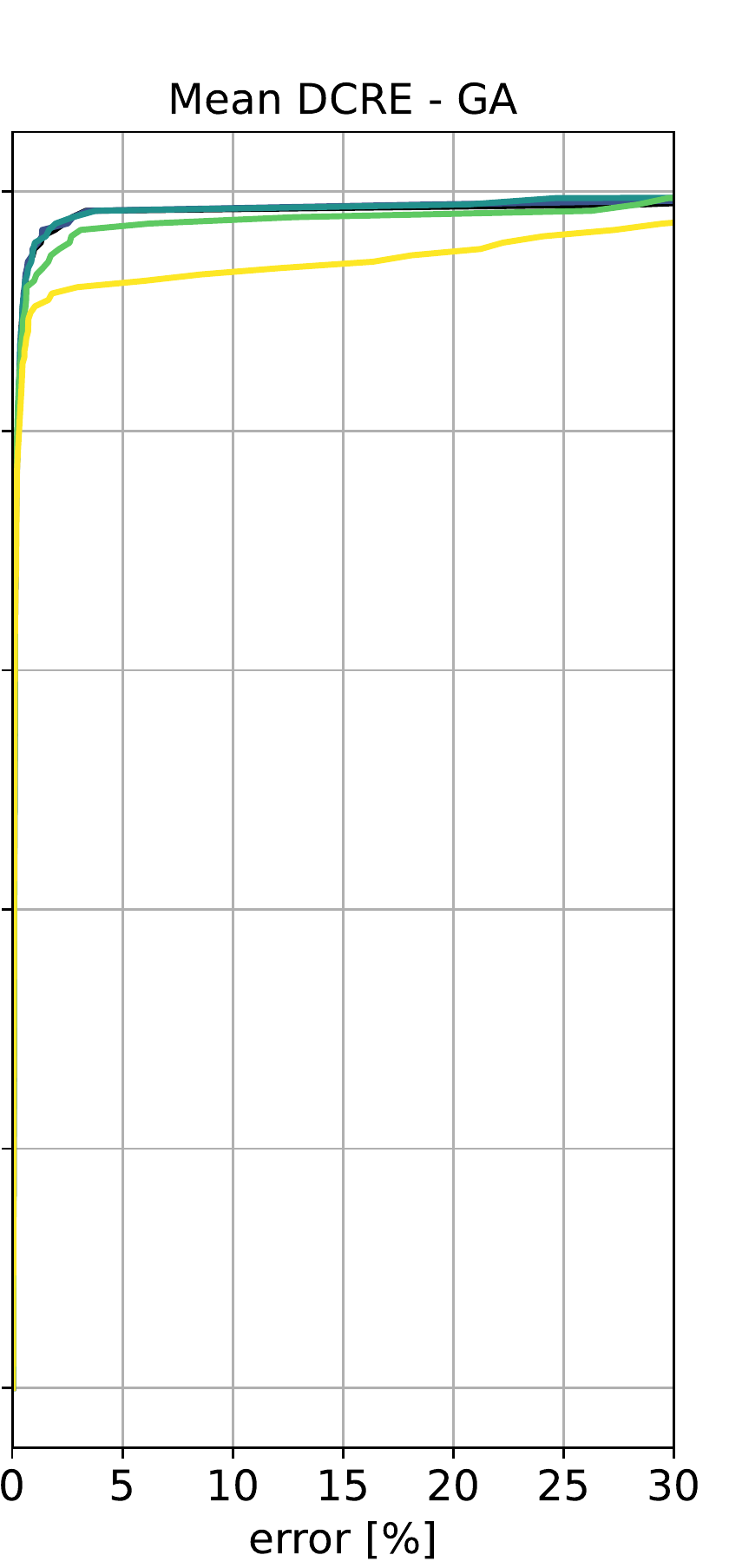}
        \caption{texture downsampling}
    \end{subfigure}
    \hfill
    \begin{subfigure}[b]{0.239\textwidth}
        \centering
        \includegraphics[width=\textwidth,trim={3.5cm 1.7cm 8cm 0},clip]{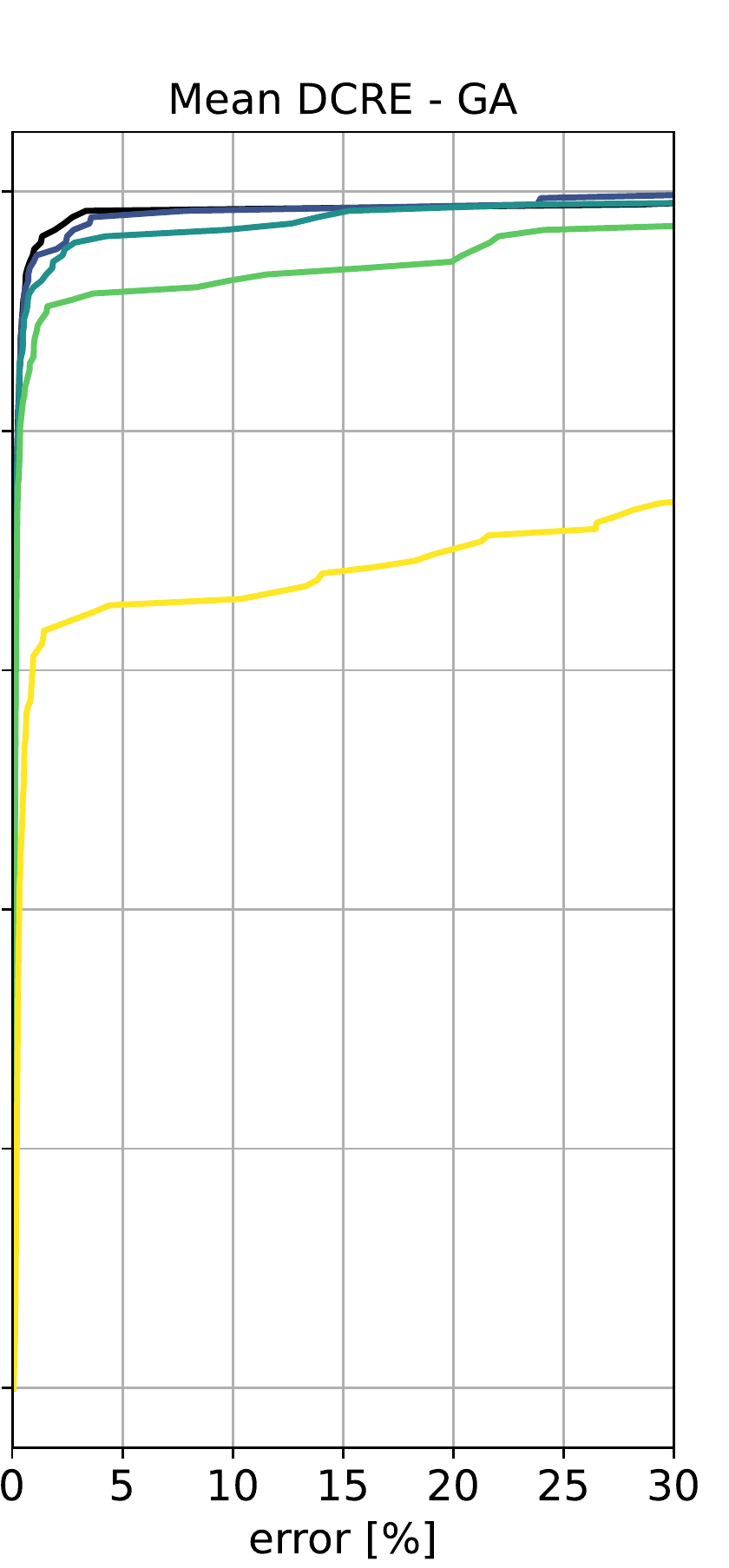}
        \caption{simplification of both}
    \end{subfigure}
    \hfill
    \begin{subfigure}[b]{0.21\textwidth}
        \centering
        \includegraphics[width=\textwidth,trim={14.1cm 6.3cm 2.2cm 6.5cm},clip]{figures/simplification_notre_dame/simplified_all.pdf}
    \end{subfigure}
    
    \caption{Localization performance of MeshLoc~\cite{Panek2022ECCV} using LoFTR when reducing the geometric and / or texture resolution of the Notre Dame A model.}
    \label{fig:simplify_fine}
\end{figure*}

The localization results of the MeshLoc~\cite{Panek2022ECCV} pipeline with LoFTR~\cite{Sun2021CVPR} are show in Fig.~\ref{fig:simplify_fine}. The localization method uses both the rendered images and depth maps. The localization pipeline copes surprisingly well, even with very high levels of geometric and appearance simplification. We can see a major drop in accuracy only after the combination of simplified geometry and downsampled texture at a very high simplification ratio of 1/16. Note that the simplified model is significantly more compact than the original one (18.4 MB original vs. 0.7 MB at 1/16 ratio), which suggests a potential use of the simplified meshes as very compact scene representations.

Note that the experiment was done using a MVS mesh reconstructed from images. Automatically simplifying the geometry of manually created CAD models can result in a complete collapse of the model geometry even for very low simplification ratios, as the CAD models are often composed of a small number of planar walls.

{\small
\bibliographystyle{ieee_fullname}
\bibliography{egbib}
}

\clearpage
\end{appendix}

\end{document}